\documentclass{article} %
\usepackage{iclr2025_conference,times}
\usepackage{caption}
\usepackage{subcaption}
\usepackage[utf8]{inputenc} %
\usepackage[T1]{fontenc}    %
\usepackage{url}            %
\usepackage{booktabs}       %
\usepackage{amsfonts}       %
\usepackage{nicefrac}       %
\usepackage{microtype}      %
\usepackage{xcolor}         %
\usepackage{makecell}
\usepackage{booktabs}
\usepackage{collcell}
\usepackage{amsmath}
\usepackage{amssymb}
\usepackage{mathtools}
\usepackage{slashed}
\usepackage{braket}
\usepackage{enumitem}
\usepackage{nicematrix}
\usepackage{hhline}
\usepackage[colorlinks,citecolor=darkgreen,linkcolor=firebrick,urlcolor=firebrick,linktocpage,unicode]{hyperref}
\usepackage{array} %
\allowdisplaybreaks

\usepackage{graphicx}
\usepackage{tikz}
\usepackage{tikz-cd}
\usepackage{times}

\usepackage{bm}
\usepackage{physics}
\usepackage{xcolor}
\usepackage{natbib}
\usepackage{mdframed}
\usepackage{nicefrac}
\usepackage{booktabs}
\usepackage{lipsum}
\usepackage{titlesec}
\usepackage{wrapfig,lipsum,booktabs}
\usepackage{blindtext}
\usepackage{algorithm}
\usepackage{algorithmicx}
\usepackage{algpseudocode}
\usepackage{titletoc}
\usepackage{todonotes}
\usepackage{authblk}
\usepackage{titletoc}
\usepackage{setspace}
\usepackage{dsfont} %
\newcommand{\one}{\mathds{1}}

\usepackage[nameinlink,capitalize,noabbrev]{cleveref}

\usepackage{CJK}

\newenvironment{claim}{  \begin{mdframed}[linecolor=black!0,backgroundcolor=black!10]\noindent%
		\ignorespaces}{\end{mdframed}}

\usepackage{array}
\usepackage{makecell}

\def\half{{\frac{1}{2}}}

\newcommand{\bea}{\begin{eqnarray}}
\newcommand{\eea}{\end{eqnarray}}

\usepackage{slashed}

\def\({\left(}
\def\){\right)}
\def\[{\left[}
\def\]{\right]}

\usepackage{dashbox}
\usepackage{xcolor}
\usepackage{colortbl}
\usepackage{arydshln}
\definecolor{lightyellow}{rgb}{1.0, 0.95, 0.7}
\definecolor{Blue}{rgb}{0, 0, 0.8}
\definecolor{blue}{rgb}{0,0,1}
\definecolor{darkgreen}{rgb}{0,0.40,0}
\definecolor{firebrick}{rgb}{0.698,0.133,0.133}

\newcommand*{\blue}[1]{\textcolor{blue}{#1}}

\definecolor{colorA}{rgb}{1,0,0}
\definecolor{colorB}{rgb}{0,0.3,1}
\definecolor{colorC}{rgb}{0.9,0.8,0.2}
\definecolor{colorD}{rgb}{0,0.65,0}
\definecolor{lesslightgray}{rgb}{0.5,0.5,0.5}
\definecolor{light-gray}{gray}{0.95}

\let\tilde\widetilde

\newcommand{\calD}{\mathcal{D}}

\newcommand{\calL}{\mathcal{L}}

\newcommand{\calO}{\mathcal{O}}

\newcommand{\bA}{A}
\newcommand{\bB}{B}
\newcommand{\bC}{C}
\newcommand{\bD}{D}
\newcommand{\bE}{E}

\newcommand{\bG}{G}

\newcommand{\bI}{I}
\newcommand{\bJ}{J}
\newcommand{\bK}{K}

\newcommand{\bQ}{Q}

\newcommand{\bT}{T}
\newcommand{\bU}{U}
\newcommand{\bV}{V}
\newcommand{\bW}{W}
\newcommand{\bX}{X}
\newcommand{\bY}{Y}
\newcommand{\bZ}{Z}
\newcommand{\ba}{a}
\newcommand{\bb}{b}

\newcommand{\be}{e}

\newcommand{\bu}{u}

\newcommand{\bv}{v}

\newcommand{\bz}{z}
\newcommand{\diag}{\mathop{\rm{diag}}}

\newcommand{\Softmax}{\mathop{\rm{Softmax}}}

\newcommand{\C}{{\mathsf{C}}}
\newcommand{\poly}{\mathrm{poly}}

\newcommand{\wt}{\widetilde}

\renewcommand{\d}{\mathrm{d}}

\renewcommand{\tilde}{\wt}

\DeclareMathOperator{\vect}{vec}
\DeclareMathOperator{\A}{\mathsf{A}}

\newcommand{\sT}{ \mathsf{T} }

\def\R{\mathbb{R}}

\let\cite\citep 

\usepackage{amsthm}
\makeatletter
\def\th@remark{%
  \thm@headfont{\bfseries}%
  \normalfont %
  \thm@preskip\parskip %
  \thm@postskip\parskip %
}
\makeatother
\usepackage[many]{tcolorbox}
\theoremstyle{definition}
\newtheorem{theorem}{Theorem}[section]
\tcolorboxenvironment{theorem}{
  breakable,
  colback=black!10,
  colframe=white,%
  width=\dimexpr\linewidth+10pt\relax,%
  enlarge left by=-5pt,%
  enlarge right by=-5pt,%
  boxsep=5pt,%
  boxrule=0pt,
  left=0pt,right=0pt,top=0pt,bottom=0pt,
  sharp corners,
  before skip=\topsep,
  after skip=\topsep
}
\newtheorem{lemma}{Lemma}[section]
\tcolorboxenvironment{lemma}{
  breakable,
  colback=black!10,
  colframe=white,%
  width=\dimexpr\linewidth+10pt\relax,%
  enlarge left by=-5pt,%
  enlarge right by=-5pt,%
  boxsep=5pt,%
  boxrule=0pt,
  left=0pt,right=0pt,top=0pt,bottom=0pt,
  sharp corners,
  before skip=\topsep,
  after skip=\topsep
}

\tcolorboxenvironment{corollary}{
  breakable,
  colback=black!10,
  colframe=white,%
  width=\dimexpr\linewidth+10pt\relax,%
  enlarge left by=-5pt,%
  enlarge right by=-5pt,%
  boxsep=5pt,%
  boxrule=0pt,
  left=0pt,right=0pt,top=0pt,bottom=0pt,
  sharp corners,
  before skip=\topsep,
  after skip=\topsep
}

\theoremstyle{definition}
\newtheorem{definition}{Definition}[section]
\tcolorboxenvironment{definition}{
  breakable,
  colback=black!10,
  colframe=white,%
  width=\dimexpr\linewidth+10pt\relax,%
  enlarge left by=-5pt,%
  enlarge right by=-5pt,%
  boxsep=5pt,%
  boxrule=0pt,
  left=0pt,right=0pt,top=0pt,bottom=0pt,
  sharp corners,
  before skip=\topsep,
  after skip=\topsep
}
\theoremstyle{remark}
\newtheorem{remark}{Remark}[section]

\tcolorboxenvironment{assumption}{
  breakable,
  colback=black!10,
  colframe=white,%
  width=\dimexpr\linewidth+10pt\relax,%
  enlarge left by=-5pt,%
  enlarge right by=-5pt,%
  boxsep=5pt,%
  boxrule=0pt,
  left=0pt,right=0pt,top=0pt,bottom=0pt,
  sharp corners,
  before skip=\topsep,
  after skip=\topsep
}
\newtheorem{problem}{Problem}
\tcolorboxenvironment{problem}{
  breakable,
  colback=black!10,
  colframe=white,%
  width=\dimexpr\linewidth+10pt\relax,%
  enlarge left by=-5pt,%
  enlarge right by=-5pt,%
  boxsep=5pt,%
  boxrule=0pt,
  left=0pt,right=0pt,top=0pt,bottom=0pt,
  sharp corners,
  before skip=\topsep,
  after skip=\topsep
}
\newtheorem{question}{Question}
\tcolorboxenvironment{question}{
  breakable,
  colback=black!10,
  colframe=white,%
  width=\dimexpr\linewidth+10pt\relax,%
  enlarge left by=-5pt,%
  enlarge right by=-5pt,%
  boxsep=5pt,%
  boxrule=0pt,
  left=0pt,right=0pt,top=0pt,bottom=0pt,
  sharp corners,
  before skip=\topsep,
  after skip=\topsep
}
\newtheorem{hypothesis}{Hypothesis}
\tcolorboxenvironment{hypothesis}{
  breakable,
  colback=black!10,
  colframe=white,%
  width=\dimexpr\linewidth+10pt\relax,%
  enlarge left by=-5pt,%
  enlarge right by=-5pt,%
  boxsep=5pt,%
  boxrule=0pt,
  left=0pt,right=0pt,top=0pt,bottom=0pt,
  sharp corners,
  before skip=\topsep,
  after skip=\topsep
}
\crefname{hypothesis}{Hypothesis}{Hypothesises}

\crefname{theorem}{Theorem}{Theorems}
\crefname{proposition}{Proposition}{Propositions}
\crefname{lemma}{Lemma}{Lemmas}
\crefname{corollary}{Corollary}{Corollaries}
\crefname{definition}{Definition}{Definitions}
\crefname{assumption}{Assumption}{Assumptions}
\crefname{remark}{Remark}{Remarks}
\crefname{problem}{Problem}{Problems}
\crefname{property}{Property}{property}
\crefname{question}{Question}{Questions}

\numberwithin{equation}{section}
\numberwithin{theorem}{section}
\numberwithin{proposition}{section}
\numberwithin{definition}{section}
\numberwithin{lemma}{section}
\numberwithin{assumption}{section}
\numberwithin{remark}{section}

\usepackage{lipsum}

\newcommand*{\annot}[1]{\tag*{\footnotesize{\textcolor{black!50}{\big(#1\big)}}}}

\makeatletter
\let\save@mathaccent\mathaccent
\newcommand*\if@single[3]{%
    \setbox0\hbox{${\mathaccent"0362{#1}}^H$}%
    \setbox2\hbox{${\mathaccent"0362{\kern0pt#1}}^H$}%
    \ifdim\ht0=\ht2 #3\else #2\fi
}
\newcommand*\rel@kern[1]{\kern#1\dimexpr\macc@kerna}
\newcommand*\widebar[1]{\@ifnextchar^{{\wide@bar{#1}{0}}}{\wide@bar{#1}{1}}}
\newcommand*\wide@bar[2]{\if@single{#1}{\wide@bar@{#1}{#2}{1}}{\wide@bar@{#1}{#2}{2}}}
\newcommand*\wide@bar@[3]{%
    \begingroup
    \def\mathaccent##1##2{%
        \let\mathaccent\save@mathaccent
        \if#32 \let\macc@nucleus\first@char \fi
        \setbox\z@\hbox{$\macc@style{\macc@nucleus}_{}$}%
        \setbox\tw@\hbox{$\macc@style{\macc@nucleus}{}_{}$}%
        \dimen@\wd\tw@
        \advance\dimen@-\wd\z@
        \divide\dimen@ 3
        \@tempdima\wd\tw@
        \advance\@tempdima-\scriptspace
        \divide\@tempdima 10
        \advance\dimen@-\@tempdima
        \ifdim\dimen@>\z@ \dimen@0pt\fi
        \rel@kern{0.6}\kern-\dimen@
        \if#31
        \overline{\rel@kern{-0.6}\kern\dimen@\macc@nucleus\rel@kern{0.4}\kern\dimen@}%
        \advance\dimen@0.4\dimexpr\macc@kerna
        \let\final@kern#2%
        \ifdim\dimen@<\z@ \let\final@kern1\fi
        \if\final@kern1 \kern-\dimen@\fi
        \else
        \overline{\rel@kern{-0.6}\kern\dimen@#1}%
        \fi
    }%
    \macc@depth\@ne
    \let\math@bgroup\@empty \let\math@egroup\macc@set@skewchar
    \mathsurround\z@ \frozen@everymath{\mathgroup\macc@group\relax}%
    \macc@set@skewchar\relax
    \let\mathaccentV\macc@nested@a
    \if#31
    \macc@nested@a\relax111{#1}%
    \else
    \def\gobble@till@marker##1\endmarker{}%
    \futurelet\first@char\gobble@till@marker#1\endmarker
    \ifcat\noexpand\first@char A\else
    \def\first@char{}%
    \fi
    \macc@nested@a\relax111{\first@char}%
    \fi
    \endgroup
    }
\makeatother

\makeatletter
\newcommand*{\redefinesymbolwitharg}[1]{%
  \expandafter\let\csname ltx#1\expandafter\endcsname\csname #1\endcsname
  \@namedef{#1}{\@ifnextchar{^}{\@nameuse{#1@}}{\@nameuse{#1@}^{}}}%
  \expandafter\def\csname #1@\endcsname^##1##2{%
     \csname ltx#1\endcsname\ifx!##1!\else^{##1}\fi\mathopen{}\mathclose\bgroup\left(##2\aftergroup\egroup\right)
     }%
}
\makeatother
\redefinesymbolwitharg{sin}
\redefinesymbolwitharg{cos}

\titlespacing\section{0pt}{6pt plus 4pt minus 2pt}{2pt plus 2pt minus 1pt}
\titlespacing\subsection{0pt}{4pt plus 3pt minus 2pt}{2pt plus 2pt minus 1pt}
\titlespacing\subsubsection{0pt}{3pt plus 3pt minus 2pt}{2pt plus 2pt minus 1pt}

\definecolor{LightCyan}{rgb}{0.8, 0.9, 1}
\usepackage{titletoc}

\usepackage{multirow}

\setlist[itemize]{leftmargin=1em, before=\vspace{-0.3em}, after=\vspace{-0.3em}, itemsep=0.1em}
\setlist[enumerate]{leftmargin=1.4em, 
before=\vspace{-0.3em}, after=\vspace{-0.3em}, 
itemsep=0.1em}

\newcommand{\email}[1]{\href{mailto:#1}{\color{black}\texttt{#1}}}

\title{
Computational Limits of Low-Rank Adaptation (LoRA) Fine-Tuning for Transformer Models\thanks{Code is available on \href{https://openreview.net/forum?id=Lf5znhZmFu}{OpenReview}; full version and future updates are on \href{https://arxiv.org/abs/2406.03136}{arXiv}.}

}

\author{
{\bf 
Jerry Yao-Chieh Hu$^{\dagger}$
\quad
Maojiang Su$^{\ddag}$
\quad
En-Jui Kuo$^{\flat}$ 
\quad
Zhao Song$^{\S}$
\quad
Han Liu$^{\dagger}$}
\\
$^\dagger$Northwestern University \quad
$^\ddag$University of Science and Technology of China\\
$^\flat$National Yang Ming Chiao Tung University\quad
$^\S$Simons Institute, UC Berkeley\\
\vspace{1em}
{\footnotesize
   \email{jhu@u.northwestern.edu},
   ~~\email{sumaojiang@mail.ustc.edu.cn},
   ~~\email{kuoenjui@nycu.edu.tw},
   \email{magic.linuxkde@gmail.com}, ~~\email{hanliu@northwestern.edu}} 
   \vspace{-2em}
}

\iclrfinalcopy %

\definecolor{blue}{named}{black}

\begin{document}

\maketitle

\begin{abstract}
We study the computational limits of Low-Rank Adaptation (LoRA) for finetuning transformer-based models using fine-grained complexity theory.
Our key observation is that the existence of low-rank decompositions within the gradient computation of LoRA adaptation leads to possible algorithmic speedup.
This allows us to (i) identify a phase transition behavior of efficiency \blue{assuming the Strong Exponential Time Hypothesis (SETH)}, and (ii) prove the existence of almost linear algorithms by controlling the LoRA update computation term by term. 
For the former, we identify a sharp transition in the efficiency of all possible rank-$r$ LoRA update algorithms for transformers, based on specific norms resulting from the multiplications of the input sequence $\bX$, pretrained weights ${\bW^\star}$, and adapter matrices $\alpha\bB\bA/r$.
Specifically, we derive a shared upper bound threshold for such norms, and show that efficient (sub-quadratic) approximation algorithms of LoRA exist only below this threshold. 
For the latter, we prove the existence of almost linear approximation algorithms for LoRA adaptation by utilizing the hierarchical low-rank structures of LoRA gradients and approximating the gradients with a series of chained low-rank approximations. 
To showcase our theory, we consider two practical scenarios: partial (e.g., only $\bW_V$ and $\bW_Q$) and full adaptations (e.g., $\bW_Q$, $\bW_V$, and $\bW_K$) of weights in attention heads.

\end{abstract}

\section{Introduction}
\label{sec:intro}
We investigate the computational limits of  finetuning large transformer-based pretrained model with \textbf{Lo}w-\textbf{R}ank \textbf{A}daptation (\textbf{LoRA}).
This analysis is of practical importance in the era of Large Foundation Models \cite{bommasani2021opportunities}.
Large foundation models are gigantic transformer-based architectures, pretrained on vast datasets, are pivotal across multiple fields, including natural language processing \cite{achiam2023gpt,touvron2023llama2,touvron2023llama,brown2020language,floridi2020gpt}, finance  \cite{yang2023fingpt,wu2023bloomberggpt}, genomics \cite{nguyen2024hyenadna,zhou2025genomeocean,zhou2024dnaberts,zhou2023dnabert,ji2021dnabert}, medical science \cite{thirunavukarasu2023large,singhal2023large,moor2023foundation} and more.
They are powerful but very expensive to pretrain.
Therefore, most practitioners rely on finetuing methods to adapt these models for their specific needs \cite{zheng2024llamafactory,ding2022delta}.
LoRA \cite{mao2025survey,hu2021lora} is the most prevalent fine-tuning method due to its parameter efficiency due to the low-rank adaptation of model weights.
However, even with LoRA, updating the partial weights of pretrained transformer-based models using gradient methods remains costly. 
Notably, the naive backward pass in transformer architectures retains the same quadratic-in-sequence-length computational time complexity as its forward pass (see \cref{sec:quadratic_backprop} for discussions and a proof).
This work provides a timely theoretical analysis of LoRA's computational limits, aiming to advance efficient finetuning of large foundation models.

The hardness of LoRA finetuning transformer-based foundation model ties to both forward and backward passes.
To analyze, it suffices to focus on just transformer attention heads  due to their dominating quadratic time complexity in both passes.
We first make the following observation:
\begin{claim}
    The hardness of LoRA's forward pass is trivially characterized by \cite{alman2023fast}.
\end{claim}
To see this,
let $\bX\in \R^{L \times d}$ be input \blue{with length $L$}, and $\bW_K,\bW_Q,\bW_V\in \R^{d\times d}$ be attention weights, and
$\bQ=\bX\bW_V\in\R^{L\times d}$, $\bK=\bX\bW_K\in\R^{L\times d}$, $\bV=\bX\bV\in\R^{L\times d}$.
The Attention Mechanism is 
\begin{align}
\label{eqn:attention}
\bZ
=\Softmax\(\bQ\bK^\sT\beta\)\bV
=\bD^{-1}\exp(\bX\bW_Q \bW_K^\sT\bX^\sT\beta) \bX\bW_V,
\end{align}
with the inverse temperature $\beta>0$ and $\bD\coloneqq \diag\(\exp(\bX\bW_Q \bW_K^\sT\bX^\sT\beta)\one_L\)$.
Here, $\exp(\cdot)$ is entry-wise exponential function, 
 $\diag\(\cdot\)$ 
converts a vector into a diagonal matrix with the entries of the vector, and $\one_L$ is the length-$L$ all ones vector.
LoRA finetuning is given as
\begin{definition}[LoRA \cite{hu2021lora}]
\label{eqn:lora_def}
    Let $\bW\in\R^{b\times a}$ be any weight matrix in a pretrained model $F$, LoRA fine-tunes $F$ through updating $\bW$ with a low-rank decomposition 
    $\bW=\bW^\star + \frac{\alpha}{r} \bB\bA$.
    Here, $\bW^\star$ is the frozen pretrained weight. Only $\bB\in\R^{b\times r}$ and $\bA\in\R^{r\times a}$ are learnable (being update via gradient descent) with rank $r< \min(a,b)$ and tunable hyperparameter $\alpha\in\R$.
\end{definition}
Under the Strong Exponential Time Hypothesis (\cref{hyp:seth}), \citet{alman2023fast} state:
\begin{lemma}[Informal, \cite{alman2023fast}]
\label{result_of_alman2023}
    Fast (sub-quadratic) forward pass of transformer only exist when entries of $\bK,\bQ,\bV$ are bounded by a constant $B=\Theta(\sqrt{\log L})$.
\end{lemma}
It is easy to see that \cref{result_of_alman2023} is transferable to LoRA inference according to \cref{eqn:lora_def}.
However, we still need the hardness of backward pass to fully characterize LoRA for transformers.
The analysis of the backpropagation (backward pass) is less straightforward. 
It involves managing the computation of numerous gradients for attention scores, with the number of chain-rule terms scaling quadratically in $L$ and the numbers of LoRA weights.
While it is tempting to design algorithms to circumvent this $\Omega(L^2)$ computation time, to the best of our knowledge, there are no formal results to support and characterize such algorithms. 
To address this gap, we pose the following questions and provide a fundamental theory to fully characterize the complexity of LoRA for  transformer models:
\begin{question}\label{question1}
    Is it possible to improve the $\Omega(L^2)$ time with a bounded approximation error?
\end{question}
\begin{question}\label{question2}
    More aggressively, is it possible to do such gradient computations in almost linear time?
\end{question}
To address these questions, 
we explore approximate LoRA gradient computations with precision guarantees. 
We first layout the objective of finetuning transformer-based pretrained models.
\begin{definition}[LoRA Loss for Adapting $\bW_K$, $\bW_Q$, $\bW_V$ of an Attention Head]
\label{def:generic_attention_lora_intro}
Let $\calD=\{\bX_i,\bY_i\}_{i=1}^N$ be a dataset of size $N$ with $\bX_i\in\R^{L\times d}$ 
being the input and $\bY_i\in\R^{L\times d}$ being the label.
Fine-tuning a (self-)attention with LoRA with $\ell_2$ loss on dataset $\calD$ is formulated as
\begin{align}
&\min_{\substack{\bB_K,\bB_Q,\bB_V \in \R^{d\times r}, \\ \bA_K,\bA_Q,\bA_V \in \R^{r\times d}}}
   \calL\(\bW_K=\bW^\star_K+\frac{\alpha}{r}\bB_K\bA_K,\bW_Q=\bW^\star_Q+\frac{\alpha}{r}\bB_Q\bA_Q,\bW_V=\bW^\star_V+\frac{\alpha}{r}\bB_V\bA_V \) 
\nonumber\\
&\coloneqq  \frac{1}{2N}\sum_{i=1}^N\norm{ \bD^{-1} \exp{\bX_i \bW_Q\bW_K^\sT\bX_i^\sT\beta} \bX_i \bW_V - \bY_i }_{F}^2.
\label{eqn:full_lora}
\end{align}
Here  
    $\bD \coloneqq \diag\(  \exp{\bX\bW_Q\bW_K^\sT\bX^\sT\beta} {\one}_n \) \in \R^{L \times L}$.  
\end{definition}
We study the following approximation problem.
Let $\underline{\bZ} \coloneqq \vect(\bZ) \in \R^{ab}$ for any matrix $\bZ\in\R^{a\times b}$.
\begin{problem}[Approximate LoRA Gradient Computation ($\mathsf{ALoRAGC}(L,d,r,\epsilon)$)]\label{def:ALoRAGC}
Assume all numerical values  in $\log(L)$ bits encoding.
Let $\calL$ follow \cref{def:generic_attention_lora_intro}.
The problem of approximating gradient computation of optimizing \eqref{eqn:full_lora} is to find six surrogate gradient matrices $\{\Tilde{\bG}^{(A)}_{\mu}\in\R^{d\times r},\Tilde{\bG}^{(B)}_{\mu}\in\R^{r\times d}\}_{\mu=K,Q,V}$ such that
\begin{align*}
    \max\Big( \Big\{\norm{\Tilde{\bG}^{(B)}_\mu-\pdv{\calL}{{\bB}_{\blue{\mu}}}}_{\infty},
\norm{\Tilde{\bG}^{(A)}_\mu-\pdv{\calL}{{\bA}_{\blue{\mu}}}}_{\infty}\Big\}_{\mu=K,Q,V}\Big) \le \epsilon,
\end{align*}
for some $\epsilon>0$,  where $\norm{\bZ}_\infty\coloneqq\max_{i,j}\abs{Z_{ij}}$.
\end{problem}
\begin{remark}
    Any method or algorithm that aims to compute LoRA gradients beyond vanilla computation of \eqref{eqn:full_lora} falls within the scope of this problem. Examples include using sampling strategies to avoid full LoRA gradient computation \cite{pan2024lisa} or employing model quantization for efficiency via low-precision gradient computation \cite{li2024loftq,dettmers2024qlora}. 
    Common among these approaches is the need to compute surrogate LoRA gradients with reduced computational cost. 
    We abstract this key subroutine and consider the fundamental algorithmic \cref{def:ALoRAGC}. 
\end{remark}

In this work, we aim to investigate the computational limits of all possible efficient algorithms of $\mathsf{ALoRAGC}(L,d,r,\epsilon)$ under realistic setting $\epsilon=1/\mathrm{poly}(L)$.

\textbf{Contributions.} 
Our contributions are 2-fold:
\begin{itemize}

    \item 
    \textbf{Norm-Based Phase Transition of Efficiency (\cref{thm:main_eff}).} 
    We answer \cref{question1} by identifying a phase transition behavior on the norm of input, pretrained and adaptor weights, assuming the Strong Exponential Time Hypothesis (SETH). 
    Specifically, 
    we identify an inefficiency threshold for these norms such that, only below which, adapting transformer-based models with LoRA in $L^{2-o(1)}$ (sub-quadratic) time is possible.
    \begin{theorem}[Informal Version of \cref{thm:main_eff}]
    {\color{blue}
    Without appropriately normalized inputs $X$, pretrained attention weights $W_K^\star, W_Q^\star, W_V^\star$, and LoRA matrices $\{\alpha A_\mu B_\mu / r\}_{\mu = K, Q, V}$, there is no algorithm running in subquadratic time $O(L^{2 - \delta})$ for any constant $\delta>0$ to solve $\mathsf{ALoRAGC}$. 
    }
    \end{theorem}

    \item 
    \textbf{Existence of Almost Linear Time LoRA Algorithms.}  
    We answer \cref{question2} by proving that precision-guaranteed approximation to \cref{def:ALoRAGC} is achievable in \textit{almost linear time} via hierarchical low-rank decomposition of LoRA gradients.
    To showcase our theory, 
    we analyze two practical scenarios highlighted in \cite{hu2021lora}:
    \textit{partial} adaptations (e.g., only $\bW_V$ and $\bW_Q$ in \cref{sec:special}), and \textit{full} adaptations (e.g., $\bW_K,\bW_Q,\bW_V$ in \cref{sec:general}) of weights in attention heads.
    \begin{theorem}[Informal Version of \cref{thm:main_special,thm:main_general}]
    {\color{blue}
    Given appropriately normalized inputs $X$, pretrained attention weights $W_K^\star, W_Q^\star, W_V^\star$, and LoRA matrices $\{\alpha A_\mu B_\mu / r\}_{\mu = K, Q, V}$, there exists an algorithm that solves $\mathsf{ALoRAGC}$ in almost linear time $O(L^{1+o(1)})$. 
     }
    \end{theorem}
\end{itemize}

On the theoretical front, we characterize the computational feasibility of LoRA  by showing the existence of precision-guaranteed, efficient (subquadratic or almost linear time) LoRA methods and identifying their necessary conditions.
On the practical front, these conditions serve as valuable guidelines for implementations \blue{(please see \cref{remark:practical} for discussions and \cref{sec:exp} for numerical justifications)}.
Importantly, our theory only requires one assumption on numerical value encoding (e.g., in $\log L$ bits with $L$ being the sequence length).
Such an assumption is minimal and realistic.
No assumptions are made about the data or model, making our results widely applicable.

\textbf{Organization.}
\cref{sec:method} includes preliminaries and problem setup.
\cref{sec:special} presents analysis of  LoRA adaptation on only $\bW_Q,\bW_K$.
\cref{sec:general} presents analysis of  LoRA adaptation on all $\bW_Q,\bW_K,\bW_V$.
\cref{sec:phase_trans} characterizes the computational limits of all possible efficient algorithms for LoRA.
\cref{sec:conclusion} includes concluding remarks.
We defer discussions of related works to \cref{sec:related_works}.

\textbf{Notations.} 
We denote (column) vectors by lower case letters, and matrices by upper case  letters.
Let $\one_L$ denote the length-$L$ all ones vector. 
We write $\Braket{\ba,\bb}\coloneqq \ba^\sT \bb$ as the inner product for vectors $\ba,\bb$.
Let $\ba[i]$ denotes the $i$-th component of vector $\ba$.
Let $\bA[i,j]$ and $A_{ij}$ denotes the $(i,j)$-th entry of matrix $\bA$.
For any matrix $\bA$, let $\bA[i,\cdot]$ and $\bA[\cdot,j]$ be the $i$-th row and $j$-th column of $\bA$, respectively.
For $\bu,\bv\in\R^d$, we denote their Hadamard product as $\bu\odot\bv\coloneqq (u_1v_1,\ldots,u_dv_d)^\sT$.
The index set $\{1,\cdots,I\}$ is denoted by $[I]$, where $I\in\mathbb{N}_+$.
For any $\bz\in\R^d$, we denote $\exp(\bz)\in\R^d$ whose $i$-th entry is $\exp(z_i)$. 
Let $\norm{\bA}_{\infty}\coloneqq\max_{i,j} \abs{A_{ij}}$ for any matrix $\bA$. 
Let $\norm{ \cdot }_F$ denote the squared Frobenius norm, i.e., $\| \bA \|_F:= ( \sum_{i,j} A_{ij}^2 )^{1/2}$.

\section{Preliminaries and Problem Setup}
\label{sec:method}

This section presents the ideas we build on.

\textbf{Tensor Trick for Computing Gradients.}
The tensor trick \cite{diao2019optimal,diao2018sketching} is an instrument to compute complicated gradients in a clean and tractable fashion.
As we shall see below, the purpose of the tensor trick is to convert matrix multiplication into vector form, making the gradient w.r.t. the matrix more tractable. 
For this, we introduce vectorization and its inverse operation, matrixization.
\begin{definition}[Vectorization]
\label{def:vectorization}
    For any matrix $\bX\in\R^{L\times d}$, we define $\underline{\bX}\coloneqq \vect{(\bX)}\in \R^{Ld}$ such that $X_{i,j}=\underline{\bX}_{(i-1)d+j}$ for all $i\in[L]$ and $j\in[d]$.
\end{definition}
\begin{definition}[Matrixization]
\label{def:matrixization}
    For any vector $\underline{\bX}\in\R^{Ld}$, 
    we define $\mathrm{mat}(\underline{\bX})=\bX$ such that
    {$X_{i,j}=\mathrm{mat}(\underline{\bX})\coloneqq \underline{\bX}_{(i-1)d+j}$} for all $i\in[L]$ and $j\in[d]$, namely $\mathrm{mat}(\cdot)=\vect^{-1}(\cdot)$. 
\end{definition}

Next, we introduce necessary tensor terminologies.

\begin{definition}[Kronecker Product]
\label{def:kronecher_prod}
    Let $\bA\in\R^{L_a\times d_a}$ and $\bB\in\R^{L_b\times d_b}$.
We define the Kronecker product of $\bA$ and $\bB$ as $\bA\otimes \bB\in \R^{L_aL_b\times d_a d_b}$ such that 
$(\bA\otimes \bB)_{(i_a-1)L_b+i_b, (j_a-1)d_b+j_b}$,
is equal to $A_{i_a,j_a}B_{i_b,j_b}$ with 
$i_a\in[L_a],j_a\in[d_a],i_b\in[L_b],j_b\in[d_b]$.
\end{definition}

\begin{definition}[Sub-Block of a Tensor]
\label{def:subblock}
    For any $\bA\in\R^{L_a\times d_a}$ and $\bB\in\R^{L_b\times d_b}$, let $\A\coloneqq \bA\otimes\bB \in \R^{L_aL_b\times d_ad_b}$.
    For any $\underline{j}\in[L_a]$, we define $\A_{\underline{j}}\in\R^{L_b\times d_ad_b}$ be the $\underline{j}$-th $L_b\times d_ad_b$ sub-block of $\A$.
\end{definition}
\cref{def:kronecher_prod} creates a large matrix from two smaller matrices, preserving the structure and properties of the original matrices.
\cref{def:subblock} provides a refined identification of specific entry-wise multiplications between the two \textit{Kronecker-producted} matrices.
Together, they makes the gradient w.r.t. the matrix more tractable: for instance, the gradient of below vectorized LoRA loss  \eqref{eqn:tensor_loss}.

\begin{lemma}[Tensor Trick \cite{diao2019optimal,diao2018sketching}]
\label{lemma:tensor_trick}
    For any $\bA\in\R^{L_a\times d_a}$, $\bB\in\R^{L_b\times d_b}$ and $\bX\in\R^{d_a\times d_b}$,
    it holds $\vect\(\bA\bX\bB^\sT\)=(\bA\otimes \bB) \underline{\bX}\in \R^{L_aL_b}$.
\end{lemma}

To showcase the tensor trick for LoRA, let's consider a (single data point) simplified \eqref{eqn:full_lora}
\begin{align*}
\calL_{0}\coloneqq 
\big\| 
\underbrace{\bD^{-1}}_{\in \R^{L\times L}} 
\underbrace{\exp{\bX \bW\bX^\sT\beta}
}_{\in\R^{L\times  L}}
\underbrace{\bX}_{\in\R^{L\times d}} \underbrace{\bW_V}_{d\times d} - \underbrace{\bY}_{\in\R^{L\times d}} 
\big\|_{F}^2, \quad\text{with } \bW\coloneqq\bW_Q\bW_K^\sT\in\R^{d\times d}.
\end{align*}
By \cref{def:kronecher_prod} and \cref{def:subblock},
we identify $D_{\underline{j},\underline{j}}\coloneqq\Braket{\exp(\A_{\underline{j}}\underline{\bW}),\one_L}\in\R$ for all $\underline{j}\in[L]$, with $\A\coloneqq \bX\otimes \bX\in\R^{L^2\times d^2}$ and $\underline{\bW}\in\R^{d^2}$.
Therefore, for each $\underline{j}\in[L]$ and $\underline{i}\in [d]$, it holds 
\bea
\label{eqn:tensor_loss}
\calL_0= \sum_{\underline{j}=1}^L\sum_{\underline{i}=1}^d 
\half \(\Braket{D^{-1}_{\underline{j},\underline{j}}\exp(\A_{\underline{j}}\underline{\bW}), \bX\bW_V[\cdot, \underline{i}]}-Y_{\underline{j},\underline{i}}\)^2.
\eea
\citet{gao2023fast,gao2023context} show that \eqref{eqn:tensor_loss} provides term-by-term tractability for gradient computation of $\calL_0$.
Specifically, it allow us to convert the attention score $\bD^{-1}\exp(\bX\bW\bX^{\sT})$ into its vectorized form $(\bD\otimes \bI_L)^{-1}\exp(\A\underline{\bW})\in\R^{L^2}$ and split the vectorized form into $L$ terms of size $L$.
This provides a systematic way to manage the chain-rule terms in the gradient computation of losses like $\calL_0$, and opens the door to more general analytical feasibility for deep transformer-based models.

\textbf{Problem Setup: Which Attention Weights in Transformer Should We Apply LoRA to?}
Following \cite{hu2021lora}, we consider only adapting the attention weights for downstream tasks.
This consideration is sufficient to justify our techniques as the attention head dominates the time complexity of transformer-based foundation models. 
Namely, we consider updating (as in \cref{def:generic_attention_lora_intro})
\begin{align*}
    \bW_Q = \bW_Q^\star +\frac{\alpha}{r} \bB_Q\bA_Q,
    \quad
    \bW_K = \bW_K^\star +\frac{\alpha}{r} \bB_K\bA_K,
    \quad
    \bW_V = \bW_V^\star +\frac{\alpha}{r} \bB_V\bA_V.
\end{align*}
Furthermore, for completeness, we consider two de facto scenarios as in \cite[Sec.~7.1]{hu2021lora}:
\begin{itemize}[leftmargin=2.2em]
    \item [(C1)] \label{item:C1}
    \textbf{Special Case.} Adapting only $\bW_Q$ and $\bW_V$ for best performance under fixed parameter budge.

    \item [(C2)] \label{item:C2} \textbf{General Case.} 
    Adapting $\bW_K,\bW_Q,\bW_V$ for best performance.
\end{itemize}
We analyze \hyperref[item:C1]{(C1)} \textbf{Special Case} in \cref{sec:special} and \hyperref[item:C2]{(C2)} \textbf{General Case}
in \cref{sec:general}.

To consider the problem of adapting attention head, we first generalize \cref{def:generic_attention_lora_intro} to the following generic attention with triplet input sequences.
For reasons, this allows our results to be  applicable.
Moreover, this helps us to focus on parts dominating the efficiency of gradient computation.
\begin{definition}[Learning Generic Attention]
\label{def:generic_attention}
Let $\calD=\{(\bX^{(K)}_i,\bX^{(Q)}_i,\bX^{(V)}_i),\bY_i\}_{i=1}^N$ be a dataset of size $N$ with the triplet $\bX^{(K)}_i,\bX^{(Q)}_i,\bX^{(V)}_i\in\R^{L\times d}$ 
being the input and $\bY_i\in\R^{L\times d}$ being the label.
The problem of learning a generic attention with $\ell_2$ loss from dataset $\calD$ is formulated as
\begin{align*}
&\min_{\bW_K,\bW_Q,\bW_V\in\R^{d\times d} }   \frac{1}{N}\sum_{i=1}^N \calL\(\bW_K,\bW_Q,\bW_V\) 
\nonumber\\
&\coloneqq  \min_{\bW_K,\bW_Q,\bW_V\in\R^{d\times d} } 
\frac{1}{2N}\sum_{i=1}^N\norm{ \bD^{-1} \exp{\bX^{(Q)}_i \bW_Q\bW_K^\sT\(\bX^{(K)}_i\)^\sT\beta} \bX^{(V)}_i \bW_V - \bY_i }_{F}^2.
\end{align*}
Here  
    $\bD \coloneqq \diag\(  \exp{\bX^{(Q)}_i \bW_Q\bW_K^\sT\(\bX^{(K)}_i\)^\sT\beta} {\one}_n \) \in \R^{L \times L}$.     
\end{definition}

\begin{remark}\label{remark:generic_att}
\cref{def:generic_attention} is generic.
    If $\bX^{(K)}_i=\bX^{(V)}_i\neq \bX^{(Q)}_i\in\R^{L\times d}$, \cref{def:generic_attention} reduces to cross-attention.
    If $\bX^{(K)}_i=\bX^{(Q)}_i=\bX^{(V)}_i\in\R^{L\times d}$, \cref{def:generic_attention} reduces to self-attention.
    
\end{remark}

\section{Special Case: LoRA Adaptation on only \texorpdfstring{$\bW_Q$}{} and \texorpdfstring{$\bW_V$}{}}
\label{sec:special}

Formally, we formulate the \textit{partial} adaptation \hyperref[item:C1]{(C1)} of an attention head as the following LoRA loss.
\begin{definition}[Adapting $\bW_Q$, $\bW_V$ of Generic Attention with LoRA]
\label{prob:generic_attention_lora_special}
Let $\calD=\{\(\bX^{(K)}_i,\bX^{(Q)}_i,\bX^{(V)}_i\),\bY_i\}_{i=1}^N$ be a dataset of size $N$ with the triplet $\bX^{(K)}_i,\bX^{(Q)}_i,\bX^{(V)}_i\in\R^{L\times d}$ 
being the input and $\bY_i\in\R^{L\times d}$ being the label.
The problem of fine-tuning $\bW_Q$, $\bW_V$ a generic attention with LoRA with $\ell_2$ loss from dataset $\calD$ is formulated as
\begin{align}
\label{eqn:special_generic_original}
&\min_{\substack{\bB_Q,\bB_V \in \R^{d\times r} \\ \bA_Q,\bA_V \in \R^{r\times d}}}
\calL\(\bW^\star_K,\bW_Q=\bW^\star_Q+\frac{\alpha}{r}\bB_Q\bA_Q,\bW_V=\bW^\star_V+\frac{\alpha}{r}\bB_V\bA_V \) 
\\
&\coloneqq \min_{\substack{\bB_Q,\bB_V \in \R^{d\times r} \\ \bA_Q,\bA_V \in \R^{r\times d}}} \frac{1}{2N}\sum_{i=1}^N\Bigg\| 
\underbrace{
    \bD^{-1} \exp{\bX^{(Q)}_i \bW_Q(\bW_K^\star)^\sT\(\bX^{(K)}_i\)^\sT\beta} 
}_{(I)}
\underbrace{
    \bX^{(V)}_i \bW_V 
}_{(II)} - \bY_i\Bigg\|_{F}^2.
\nonumber
\end{align}
Here  
    $\bD \coloneqq \diag\(  \exp{\bX^{(Q)}_i \bW_Q(\bW_K^\star)^\sT\(\bX^{(K)}_i\)^\sT\beta} {\one}_n \) \in \R^{L \times L}$.  
\end{definition}

In this work, we are interested in the efficiency of optimizing \eqref{eqn:special_generic_original} with gradient descent.  
For simplicity of our analysis, we employ the following four simplifications:
\begin{itemize}[leftmargin=2.2em]

    \item [(S1)] \label{item:S1} Since (II) ($\bV$ multiplication) is linear in weight while (I) ($\bK$-$\bQ$ multiplication) is exponential in weights,
    we only need to focus on the gradient of $\bK$-$\bQ$ multiplication.
    Therefore, for efficiency analysis of gradient, it is equivalent to analyze a  reduced problem with fixed $\bW_V$.

    \item [(S2)] \label{item:S2}
    To further simplify, we introduce $\bC^{(1)}_i,\bC^{(2)}_i,\bC^{(3)}_i\in\R^{L\times d}$ via
    \begin{align}
    \label{eqn:C1C2C3}
    \underbrace{
    \bX^{(Q)}_i\frac{\alpha}{r} }_{\coloneqq\bC^{(1)}_i\in\R^{L\times d}}
    \Big(\frac{r}{\alpha}\bW^\star_Q+
    \bB_Q\bA_Q\Big)\underbrace{
    (\bW_K^\star)^\sT\(\bX^{(K)}_i\)^\sT
    }_{\coloneqq\(\bC^{(2)}_i\)^\sT\in \R^{d\times L}}
    \coloneqq
    \bC^{(1)}_i \bB_Q\bA_Q \(\bC^{(2)}_i\)^\sT,
    \quad
    \bX^{(V)}_i \bW_V^\star
    \coloneqq
    \bC^{(3)}_i.
    \end{align}
    Notably, $\bC^{(1)}_i,\bC^{(2)}_i,\bC^{(3)}_i$ are constants with respect to adapting \eqref{eqn:special_generic_original} with gradient updates.
    
    \item [(S3)] \label{item:S3}
    \textbf{Trivial Reduction.}
    To prove the hardness of \cref{def:ALoRAGC} for both full gradient descent and stochastic mini-batch gradient descent, 
    it suffices to consider adapting on a single data point.

    \item [(S4)] \label{item:S4}
    We set $\beta=1$ without loss of generality.
    Note that $\beta$ and $\alpha/r$  do not impact the running time of gradient computation since they are just rescaling factors.

\end{itemize}

Thus, we deduce \cref{prob:generic_attention_lora_special} to
\begin{align}
\label{eqn:special_generic_lora_loss_single}
\min_{\substack{\bB_Q \in \R^{d\times r} \\ \bA_Q \in \R^{r\times d}}} \calL(\bB_Q,\bA_Q)=\min_{\substack{\bB_Q \in \R^{d\times r} \\ \bA_Q \in \R^{r\times d}}} \half\norm{ \bD^{-1} \exp{\bC^{(1)} \Big(\Bar{\bW}_{Q}^\star+
    \bB_Q\bA_Q\Big)  \(\bC^{(2)}\)^\sT} \bC^{(3)} - \bY }_{F}^2,
\end{align}
where  $\Bar{\bW}_{Q}^\star\coloneqq r\bW^\star_Q/\alpha$ and $\bD=\diag\(  \exp{\bC^{(1)} \Big(\Bar{\bW}_{Q}^\star+
    \bB_Q\bA_Q\Big) \(\bC^{(2)}\)^\sT} {\one}_L \) \in \R^{L \times L}$.
    
We introduce the next problem to characterize all possible (efficient or not) gradient computation of optimizing \eqref{eqn:special_generic_lora_loss_single}.
Let $\bY[i,\cdot]$ and $\bY[\cdot,j]$ be the $i$-th row and $j$-th column of $\bY$, respectively.
\begin{problem}[Approximate LoRA Gradient Computation $\mathsf{ALoRAGC}(L,d,r,\epsilon)$]
\label{def:ALoRAGC_special}
Given  $\bC^{(1)}_i,\bC^{(2)}_i,\bC^{(3)}_i,\bY_i\in\R^{L\times d}$.
Let $\epsilon>0$.
Assume all numerical values are in $\log(L)$-bits encoding.
Let $\calL$ follows \eqref{eqn:special_generic_lora_loss_single}.
The problem of approximating gradient computation of optimizing \eqref{eqn:special_generic_lora_loss_single} is to find two matrices $\Tilde{\bG}^{(A)}_Q\in\R^{{d\times r}}$ and $\Tilde{\bG}^{(B)}_Q\in\R^{{r\times d}}$ such that
\begin{align*}
    \max\Big( \|\underline{\Tilde{\bG}}^{(B)}_Q-\pdv{\calL}{\underline{\bB}_Q}\|_{\infty},
    \|\underline{\Tilde{\bG}}^{(A)}_Q-\pdv{\calL}{\underline{\bA}_Q}\|_{\infty}\Big) \le \epsilon.
\end{align*}
\end{problem}

The explicit gradient of LoRA loss \eqref{eqn:special_generic_lora_loss_single} is too complicated to characterize \cref{def:ALoRAGC_special}.
To combat this, we employ the tensor trick.
Let $\bW\coloneqq\Bar{\bW}_Q^\star+\bB_Q\bA_Q\in\R^{d\times d}$ such that $\vect{(\bW)}=\underline{\bW}\in\R^{d^2}$.
\begin{definition}[\blue{Vectorized Attention Score}]
\label{def:special_u}
Let $\C\coloneqq \bC^{(1)}\otimes \bC^{(2)}$ such that $\C_{\underline{j}}\in \R^{L\times d^2}$ for all $\underline{j}\in[L]$.
For every $\underline{j} \in [L]$,
we define  $u(\underline{\bW})_{\underline{j}}: \R^{d^2} \rightarrow \R^L$ as:
$u(\underline{\bW})_{\underline{j}} \coloneqq  \exp( \C_{\underline{j}} \underline{\bW} ) \in\R^{L}$.
\end{definition}

\cref{def:special_u} decomposes the complicated matrix $\exp(\bC^{(1)} (\Bar{\bW}_{Q}^\star+
    \bB_Q\bA_Q) (\bC^{(2)}_i)^\sT)$ in loss \eqref{eqn:special_generic_lora_loss_single} into $L$ vectors.
Importantly, since the weight $\bW$ is vectorized into $\underline{\bW}$, 
such a vectorized representation allows more tractable gradient computation by its term-by-term identifiability.

\begin{definition}[\blue{Attention Score Normalization}]
\label{def:special_alpha}
Let $\C\coloneqq \bC^{(1)}\otimes \bC^{(2)}$ such that $\C_{\underline{j}}\in \R^{L\times d^2}$ for all $\underline{j}\in[L]$.
For every $\underline{j} \in [L]$, we define $\alpha(x)_{\underline{j}}: \R^{d^2} \rightarrow \R$ as:
$\alpha(\underline{\bW})_{\underline{j}}\coloneqq 
\Braket{  \exp( \C_{\underline{j}} \underline{\bW} )  ,  \one_L } \in \R$.
\end{definition}

Similarly, \cref{def:special_u,def:special_alpha} provide analytical tractability of the matrix $\bD$ in loss \eqref{eqn:special_generic_lora_loss_single}.

\begin{definition}[\blue{Vectorized, Normalized Attention Score}]
\label{def:special_f}
For a fixed $\underline{j} \in [L]$, we define $f(\underline{\bW} )_{\underline{j}} : \R^{d^2} \rightarrow \R^L$ as:
$f(\underline{\bW})_{\underline{j}} \coloneqq  \alpha(\underline{\bW})_{\underline{j}}^{-1}    u(\underline{\bW})_{\underline{j}}$ such that
 $f(\underline{\bW} ) \in \R^{L \times L}$ denotes the matrix whose $\underline{j}$-th row is $( f(\underline{\bW} )_{\underline{j}} )^\top$.
\end{definition}

\cref{def:special_f} decomposes the complicated matrix multiplication $\bD^{-1}\exp(\bC^{(1)} ({\bW}_{Q}^\star+
    \bB_Q\bA_Q)  (\bC^{(2)})^\sT) \bC^{(3)}$ in loss \eqref{eqn:special_generic_lora_loss_single} into $L$ terms.
Note that the gradients w.r.t. $\underline{\bW}$ are still tractable due to simple chain rule (by design of $\alpha(\cdot)$ and $u(\cdot)$).

\begin{definition}[\blue{Vectorized LoRA Loss \eqref{eqn:special_generic_lora_loss_single}}]
\label{def:special_c}
For every $i \in [d]$, let $ \bC^{(3)}[\cdot,i]$ follow \hyperref[item:S2]{(S2)}.
For every $\underline{j} \in [L]$ and  $i \in [d]$, 
we define $c(x)_{\underline{j},i}: \R^{d^2} \times \R^{d^2} \rightarrow \R$ as:
$c(\underline{\bW})_{\underline{j},i}\coloneqq \langle f(\underline{\bW})_{\underline{j}},  \bC^{(3)}[\cdot,i] \rangle - Y_{\underline{j},i}$.
Here $Y_{\underline{j},i}=\bY[\underline{j},i]$ is the $(\underline{j},i)$-th entry of $\bY \in \R^{L \times d}$ for $\underline{j} \in [L], i \in [d]$.
\end{definition}
From above definitions, we read out  $ c(\underline{\bW})  =f(\underline{\bW})  \bC^{(3)}  - \bY$ 
such that \eqref{eqn:special_generic_lora_loss_single} becomes
\bea
\label{eqn:special_element_form}
\calL(\underline{\bW})=\sum_{\underline{j}}^L\sum_{i=1}^d
\mathcal{L}(\underline{\bW})_{\underline{j},i} =
\half\sum_{\underline{j}}^L\sum_{i=1}^d 
c(\underline{\bW})_{\underline{j},i}^2.
\eea
\eqref{eqn:special_element_form} presents a decomposition of the LoRA loss \eqref{eqn:special_generic_lora_loss_single} into $L \cdot d$ terms, each simple enough for tracking gradient computation.
Now, we are ready to compute the gradient of the LoRA loss.
\begin{lemma}[Low-Rank Decomposition of LoRA Gradient]
\label{lemma:special1}
    Let matrix $\bB_Q,\bA_Q$ and loss function $\mathcal{L}$ follow \eqref{eqn:special_generic_lora_loss_single}, $\bW\coloneqq\Bar{\bW}_Q^\star+\bB_Q\bA_Q$ and $\C\coloneqq \bC^{(1)}\otimes \bC^{(2)}$.
    It holds
    \bea
    \label{eqn:grad_low_rank}
    \dv{\mathcal{L}(\underline{\bW})}{\underline{\bW}}
    =\sum_{\underline{j}=1}^L \sum_{i=1}^d c(\underline{\bW})_{\underline{j}, i} \C_{\underline{j}}^{\top}\underbrace{\Big(\overbrace{\diag\left(f(\underline{\bW})_j\right)}^{(II)}-\overbrace{f(\underline{\bW})_{\underline{j}} f(\underline{\bW})_{\underline{j}}^{\top}}^{(III)}\Big) 
    }_{(I)}\bC^{(3)}[\cdot, i].
    \eea
\end{lemma}

\begin{proof}
    See \cref{proof:lemma:special1} for a detailed proof.
\end{proof}
\begin{remark}[Benefit from Tensor Trick: Fast Approximation]
    As we shall show in subsequent sections, \cref{lemma:special1} also enables the construction of fast approximation algorithms for \eqref{eqn:grad_low_rank} with precision guarantees due to its analytical feasibility.
    Surprisingly, it is even possible to compute \eqref{eqn:grad_low_rank} in almost linear time.
    To proceed, we further decompose \eqref{eqn:grad_low_rank} into its fundamental building blocks according to the chain-rule in the next lemma, and then conduct the approximation term-by-term.
\end{remark}
\begin{remark}[LoRA Gradient Computation Takes Quadratic Time]
    \cref{lemma:special1} implies that LoRA's gradient computation takes quadratic time, 
    similar to inference hardness result \cite{alman2023fast}.
    This is non-trivial yet not the main focus of this work.
    Please see \cref{sec:quadratic_backprop} for details.
\end{remark}

\begin{lemma}[\blue{Vectorized $\pdv{\calL}{A_Q}$, $\pdv{\calL}{B_Q}$}]
\label{lemma:special3}
Let $q(\underline{\bW})
    \coloneqq \bC^{(3)} \(c(\underline{\bW})\)^\sT \in \R^{L\times L}$.
For every index $\underline{j}\in [L]$ , we define $p(\underline{\bW})_{\underline{j}}\in \R^L$ as
$p(\underline{\bW})_{\underline{j}}
\coloneqq \left(\diag\left(f\left(\underline{\bW}\right)_j\right)-f\left(\underline{\bW}\right)_{\underline{j}} f\left(\underline{\bW}\right)_{\underline{j}}^{\top}\right) q(\underline{\bW})$.
Then it holds
\begin{align}
\label{eqn:special_LoRA_gradients}
             \pdv{\calL}{\underline{\bA}_Q}
            =\vect\left(\bB_Q^{\top}\left(\bC^{(1)}\right)^{\top} p(\underline{\bW}) \bC^{(2)}\right) 
            ,\quad
            \frac{\partial \mathcal{L}}{\partial \underline{\bB}_Q}
            =\vect\left(\left(\bC^{(1)}\right)^{\top} p(\underline{\bW}) \bA_Q \bC^{(2)}\right).
\end{align}
\end{lemma}

\begin{proof}
    See \cref{proof:lemma:special3} for a detailed proof.
\end{proof}

\cref{lemma:special3} states that the chain rule terms for characterizing \cref{def:ALoRAGC_special} are tied to $p(\cdot)$.
Therefore, to characterize $\Tilde{\bG}^{(A)}_Q$, $\Tilde{\bG}^{(B)}_Q$ (i.e., the approximations of $\bG^{(A)}_Q$, $\bG^{(B)}_Q$), we need to approximate the functions $f(\cdot)$, $q(\cdot)$, $c(\cdot)$, and hence $p(\cdot)$ with precision guarantees.
To do so, it is convenient to consider the following decomposition of $p(\cdot)$.

\begin{definition}[\blue{Decomposition of $p(\cdot)$}]
\label{def:p1p2}
For every $\underline{j}\in [L]$, we define $p_1(\underline{\bW})_{\underline{j}},p_2(\underline{\bW})_{\underline{j}}\in \R^L$ as
\begin{align*}
p_1(\underline{\bW})_{\underline{j}}\coloneqq \diag\left(f\left(\underline{\bW}\right)_{\underline{j}}\right) q(\underline{\bW})_{\underline{j}}\quad
\text{and}\quad
p_2(\underline{\bW})_{\underline{j}}\coloneqq f\left(\underline{\bW}\right)_{\underline{j}} f\left(\underline{\bW}\right)_{\underline{j}}^{\top} q(\underline{\bW})_{\underline{j}},
\end{align*}
such that 
$p(\underline{\bW})=p_1(\underline{\bW})-p_2(\underline{\bW})$.
\end{definition}

\textbf{Overview of Our Proof Strategy.} 
\cref{def:p1p2} motivates the following strategy: term-by-term approximation for precision-guaranteed, almost linear time algorithms to compute \eqref{eqn:special_LoRA_gradients} (\cref{def:ALoRAGC_special}).
\begin{enumerate}[leftmargin=3.5em]
    \item [\textbf{Step 1.}] Prove the existence of almost linear approximation algorithms for $f(\cdot), q(\cdot),c(\cdot)$ via low-rank approximation:
    \cref{lemma:approx_f}, \cref{lemma:approx_q} and \cref{lemma:approx_c}.
    \item [\textbf{Step 2.}] Prove the existence of almost linear approximation algorithms for $p_1(\cdot),p_2(\cdot)$ and hence $p(\cdot)$ via the low-rank-preserving property of the multiplication between $f(\cdot)$ and $q(\cdot)$: \cref{lemma:approx_p1} and \cref{lemma:approx_p2}.
    
    \item [\textbf{Step 3.}]
    Prove existence of almost linear approximation algorithms for the LoRA adapter gradients (i.e., $\pdv{\calL}{\underline{\bA}_Q}$ and $\pdv{\calL}{\underline{\bB}_Q}$ in \eqref{eqn:special_LoRA_gradients}) with results from \textbf{Step 1 \& 2}:  \cref{thm:main_special}.
\end{enumerate}

\textbf{Step 1.} We start with low-rank approximations for $f(\cdot),q(\cdot),c(\cdot)$.
\begin{lemma}[Approximate $f(\cdot)$, Modified from  \cite{alman2023fast}]
\label{lemma:approx_f}
    Let $\Gamma = o(\sqrt{\log L})$ and $k_1=L^{o(1)}$.
    Let $\bC^{(1)}, \bC^{(2)} \in \mathbb{R}^{L \times d}$, $\bW\in\R^{d\times d}$, and
    $f(\underline{\bW})=\bD^{-1} \exp \left(\bC^{(1)} \bW\left(\bC^{(2)}\right)^{\top}\right)$ with $\bD=\diag\left(\exp \left(\bC^{(1)} \bW\left(\bC^{(2)}\right)^{\top}\right) {\one_L}\right)$ follows 
    \cref{def:special_u,def:special_alpha,def:special_c,def:special_f}. 
    If $\max\big(\norm{\bC^{(1)} \bW}_{\infty} \leq \Gamma$,$\norm{\bC^{(2)}}_{\infty} \big)\leq \Gamma$, then there exist two matrices $\bU_1, \bV_1 \in \mathbb{R}^{L \times k_1}$ such that $\norm{\bU_1 \bV_1^{\top} - f(\underline{\bW})}_{\infty} \leq \epsilon / \poly (L)$.  
    In addition, it takes $L^{1+o(1)}$ time to construct $\bU_1$ and $\bV_1$.
\end{lemma}

\begin{proof}
This lemma is an application of \cite[Theorem~3.8]{alman2023fast}.
\end{proof}

\begin{lemma}[Approximate $c(\cdot)$]
\label{lemma:approx_c}
    Assume all numerical values are in $O(\log L)$ bits. 
    Let $d=O(\log L)$ and $c(\underline{\bW})\in\R^{L\times d}$ follows \cref{def:special_c}.
    There exist two matrices $\bU_1, \bV_1 \in \mathbb{R}^{L \times k_1}$ such that 
    \begin{align*}
        \left\|\bU_1 \bV_1^{\top} \bC^{(3)}-\bY-c(\underline{\bW})\right\|_{\infty} \leq \epsilon / \poly(L).
    \end{align*}
\end{lemma}

\begin{proof}
    See \cref{proof:lemma:approx_c} for a detailed proof.
\end{proof}

\begin{lemma}[Approximate $q(\cdot)$]
\label{lemma:approx_q}
    Let $k_2=L^{o(1)}$, $c(\bW)\in\R^{L\times d}$ follows \cref{def:special_c} and 
    let $q(\underline{\bW})
    \coloneqq \bC^{(3)} \(c(\underline{\bW})\)^\sT \in \R^{L\times L}$ follows \cref{lemma:special3}.
    There exist two matrices $\bU_2, \bV_2 \in \mathbb{R}^{L \times k_2}$ such that $\left\|\bU_2 \bV_2^{\top}-q(\underline{\bW})\right\|_{\infty} \leq \epsilon /\poly (L)$.
    In addition, it takes $L^{1+o(1)}$ time to construct $\bU_2, \bV_2$.
\end{lemma}

\begin{proof}
    See \cref{proof:lemma:approx_q} for a detailed proof.
\end{proof}

\textbf{Step 2.} Now, we use above lemmas to construct low-rank approximations for $p_1(\cdot),p_2(\cdot), p(\cdot)$.
\begin{lemma}[Approximate $p_1(\cdot)$]
\label{lemma:approx_p1}
    Let $k_1,k_2,k_3=L^{o(1)}$.  
    Suppose $\bU_1, \bV_1 \in \mathbb{R}^{L \times k_1}$ approximates  $f(\underline{\bW})\in\R^{L\times L}$ such that $\left\|\bU_1 \bV_1^{\top}-f(\underline{\bW})\right\|_{\infty} \leq \epsilon /\poly(L)$, and
    $\bU_2, \bV_2 \in \mathbb{R}^{L \times k_2}$ approximates the $q(\underline{\bW}) \in \mathbb{R}^{L \times L}$ such that $\left\|\bU_2 \bV_2^{\top}-q(\underline{\bW})\right\|_{\infty} \leq \epsilon /\poly(L)$. 
    Then there exist two matrices $\bU_3, \bV_3 \in \mathbb{R}^{L \times k_3}$ such that 
    \begin{align*}
        \left\|\bU_3 \bV_3^{\top}-p_1(\underline{\bW})\right\|_{\infty} \leq \epsilon /\poly(L).
    \end{align*}
    In addition, it takes $L^{1+o(1)}$ time to construct $\bU_3, \bV_3$.
    
\end{lemma}

\begin{proof}[Proof Sketch]
    By tensor formulation, we construct $\bU_3$, $\bV_3$ as tensor products of $\bU_1,\bV_1$ and $\bU_2,\bV_2$, respectively, while preserving their low-rank structure.
    Then, we show the low-rank approximation of $p_1(\cdot)$ with bounded error by \cref{lemma:approx_f} and \cref{lemma:approx_q}. 
    See \cref{proof:lemma:approx_p1} for a detailed proof.
\end{proof}

\begin{lemma}[Approximate $p_2(\cdot)$]
\label{lemma:approx_p2}
    Let $k_1,k_2,k_4=L^{o(1)}$. 
    Let $p_2(\underline{\bW})\in\R^{L\times L}$ follow \cref{def:p1p2} such that its $\underline{j}$-th column is $p_2(\underline{\bW})_{\underline{j}}=f(\underline{\bW})_{\underline{j}}f(\underline{\bW})_{\underline{j}}^{\top} q(\underline{\bW})_{\underline{j}}$ for each $\underline{j} \in [L]$. 
    Suppose $\bU_1, \bV_1 \in \mathbb{R}^{L \times k_1}$ approximates the $\mathrm{f}(\mathrm{\bX})$ such that $\left\|\bU_1 \bV_1^{\top}-f(\underline{\bW})\right\|_{\infty} \leq \epsilon /\poly (L)$, and 
     $\bU_2, \bV_2 \in \mathbb{R}^{L \times k_2}$ approximates the $q(\underline{\bW}) \in \mathbb{R}^{L \times L}$ such that $\left\|\bU_2 \bV_2^{\top}-q(\underline{\bW})\right\|_{\infty} \leq \epsilon /\poly (L)$. 
     Then there exist matrices $\bU_4, \bV_4 \in \mathbb{R}^{L \times k_4}$ such that 
     \begin{align*}
         \left\|\bU_4 \bV_4^{\top}-p_2(\underline{\bW})\right\|_{\infty} \leq \epsilon /\poly(L)
     \end{align*}
     In addition, it takes $L^{1+o(1)}$ time to construct $\bU_4, \bV_4$.
\end{lemma}

\begin{proof}[Proof Sketch]
    By considering the following decomposition through tensor formulation
    \begin{align*}
    p_2(\underline{\bW})_{\underline{j}}\coloneqq \overbrace{f\left(\underline{\bW}\right)_{\underline{j}}  \underbrace{f\left(\underline{\bW}\right)_{\underline{j}}^{\top} q(\underline{\bW})_{\underline{j}}}_{(I)}
    }^{(II)},
    \end{align*}
    we approximate the $p_2(\cdot)$  part by part. 
    Specifically, 
    for (I),
    we show its low-rank approximation by observing the low-rank-preserving property of the multiplication between $f(\cdot)$ and  $q(\cdot)$ (from \cref{lemma:approx_f} and \cref{lemma:approx_q}). 
    For (II), 
    we show its low-rank approximation by the low-rank structure of  $f(\cdot)$ and (I).
    See \cref{proof:lemma:approx_p2} for a detailed proof.
\end{proof}

\textbf{Step 3.}
Combining above, we arrive our main result:
almost linear algorithm for \cref{def:ALoRAGC_special}.

\begin{theorem}[Main Result: Existence of Almost Linear Time $\mathsf{ALoRAGC}$]
\label{thm:main_special}
Suppose all numerical values are in $O(\log L)$-bits encoding.
Recall that $\bW = \Bar{\bW}_Q^\star+\bB_Q\bA_Q\in\R^{d\times d}$ with $\Bar{\bW}_{Q}^\star\coloneqq r\bW^\star_Q/\alpha$. 
Let $\bC^{(1)} = \bX^{(Q)}\frac{\alpha}{r}, \bC^{(2)}=\bX^{(K)} \bW_K^\star$ follows \eqref{eqn:C1C2C3}.
If $\norm{\bC^{(1)} \bW}_{\infty} \leq \Gamma$ and $\norm{\bC^{(2)}}_{\infty} \leq \Gamma$, where $\Gamma=o(\sqrt{\log L}$),
then there exists a $L^{1+o(1)}$ time algorithm to solve $\mathsf{ALoRAGC}\(L, d=O(\log L), r=L^{o(1)}, \epsilon = 1/ \poly(L)\)$  (i.e., \cref{def:ALoRAGC_special}). 
In particular, this algorithm outputs  gradient matrices $\Tilde{\bG}^{(A)}_Q \in \mathbb{R}^{d \times r}, \Tilde{\bG}^{(B)}_Q \in \mathbb{R}^{r \times d}$ such that 
\begin{align*}
    \|\pdv{\calL}{\underline{\bA}_Q}-\Tilde{\underline{\bG}}^{(A)}_Q\|_{\infty} \leq 1 / \poly(L), \quad \text{and}\quad \|\frac{\partial \mathcal{L}}{\partial \underline{\bB}_Q}-\Tilde{\underline{\bG}}^{(B)}_Q\|_{\infty} \leq 1 / \poly(L).
\end{align*}
\end{theorem}

\begin{proof}[Proof Sketch]
    By \cref{lemma:special3}, we have 
             $\nicefrac{\partial \calL}{\partial \underline{\bA}_Q}
             =
             \vect(\bB_Q^{\top}(\bC^{(1)})^{\top} p(\underline{\bW}) \bC^{(2)})$, and 
             $\nicefrac{\partial \calL}{\partial \underline{\bB}_Q}=\vect((\bC^{(1)})^{\top} p(\underline{\bW}) \bA_Q \bC^{(2)})$.
    By \cref{lemma:special3} and \cref{def:p1p2},
    we have
        $p(\underline{\bW})=p_1(\underline{\bW})-p_2(\underline{\bW})$.
    Firstly, we notice that the \textit{exact} computation of  $\bB_Q^{\top}(\bC^{(1)})$ and $\bA_Q \bC^{(2)}$ takes only $L^{1+o(1)}$ time, by $\bA_Q \in \mathbb{R}^{r \times d}$, $ \bB_Q \in \mathbb{R}^{d \times r}$, $ \bC^{(1)}$, $ \bC^{(2)} \in \mathbb{R}^{L \times d}$. 
    Thus, to show the existence of $L^{1+o(1)}$ time algorithms for \cref{def:ALoRAGC_special},  
    we prove fast low-rank approximations for 
    $\bB_Q^{\top}(\bC^{(1)})^{\top} p_1(\underline{\bW}) \bC^{(2)}$ and $(\bC^{(1)})^{\top} p_1(\underline{\bW}) \bA_Q \bC^{(2)}$ by \cref{lemma:approx_p1}.
    The fast low-rank approximations for  $-\bB_Q^{\top}(\bC^{(1)})^{\top} p_2(\underline{\bW}) \bC^{(2)}$ and $-(\bC^{(1)})^{\top} p_2(\underline{\bW}) \bA_Q \bC^{(2)}$ follow trivially.
    See \cref{proof:thm:main_special} for a detailed proof.
\end{proof}

\textbf{General Case:  Full LoRA Adaptation on \texorpdfstring{$W_K,W_Q,W_V$}{}.}
In the next section, we provide the analysis of full LoRA on transformer (\hyperref[item:C2]{(C2)} \textbf{General Case:} adapting both $W_K,W_Q,W_V$).
Importantly, we also prove the existence of an almost linear-time LoRA (\cref{thm:main_general}). 
In addition, we derive the norm bound conditions required for it to hold.

\section{General Case:  Full LoRA Adaptation on \texorpdfstring{$\bW_K$}{}, \texorpdfstring{$\bW_Q$}{} and \texorpdfstring{$\bW_V$}{}}
\label{sec:general}
Similarly, we formulate the full adaptation \hyperref[item:C2]{(C2)} of an attention head as the following LoRA loss.
\begin{definition}[Adapting $\bW_K$, $\bW_Q$, $\bW_V$ of Generic Attention with LoRA]
\label{def:generic_attention_lora_general}
Let $\calD=\{(\bX^{(K)}_i,\bX^{(Q)}_i,\bX^{(V)}_i),\bY_i\}_{i=1}^N$ be a dataset of size $N$ with the triplet $\bX^{(K)}_i,\bX^{(Q)}_i,\bX^{(V)}_i\in\R^{L\times d}$ 
being the input and $\bY_i\in\R^{L\times d}$ being the label.
The problem of fine-tuning a generic attention with LoRA with $\ell_2$ loss from dataset $\calD$ is formulated as
\begin{align*}
&\min_{\substack{\bB_K,\bB_Q,\bB_V \in \R^{d\times r}, \\ \bA_K,\bA_Q,\bA_V \in \R^{r\times d}}}
   \calL(\bW_K=\bW^\star_K+\frac{\alpha}{r}\bB_K\bA_K,\bW_Q=\bW^\star_Q+\frac{\alpha}{r}\bB_Q\bA_Q,\bW_V=\bW^\star_V+\frac{\alpha}{r}\bB_V\bA_V ) 
\nonumber\\
&\coloneqq  \frac{1}{2N}\sum_{i=1}^N\norm{ \bD^{-1} \exp{\bX^{(Q)}_i \bW_Q\bW_K^\sT\bX^{(K)}_i\beta} \bX^{(V)}_i \bW_V - \bY_i }_{F}^2.
\end{align*}
Here  
    $\bD \coloneqq \diag(  \exp{\bX^{(Q)} \bW_Q\bW_K^\sT\bX^{(K)}\beta} {\one}_n ) \in \R^{L \times L}$.  
\end{definition}
By simplifications \hyperref[item:S1]{(S1)}, \hyperref[item:S3]{(S3)} and \hyperref[item:S4]{(S4)}, we fix $\bW_V$, set $\beta=\nicefrac{\alpha}{r}=1$ and consider LoRA adaptation on a single data point.
Akin to simplification \hyperref[item:S2]{(S2)}, we introduce $\bC_K^{(1)}, \bC_K^{(2)}, \bC_Q^{(1)}, \bC_Q^{(2)} ,\bC^{(3)} \in\R^{L\times d}$:
\begin{align}
\label{eqn:general_Cs}
\bC_K^{(1)}&\coloneqq \bX^{(Q)}\left(\bW_Q^\star+\frac{\alpha}{r} \bB_Q \bA_Q\right), \quad \bC_K^{(2)}\coloneqq\bX^{(K)},
\\
\bC_Q^{(1)}&\coloneqq\bX^{(Q)}, \quad\bC_Q^{(2)}\coloneqq \bX^{(K)}\left(\bW_K^\star+\bB_K \bA_K\right) , \quad \text{and}\quad
\bC^{(3)}\coloneqq\bX^{(V)} \bW_V^\star.\nonumber
\end{align}
\begin{remark}
    $\bC_K^{(1)}, \bC_K^{(2)}, \bC^{(3)}$ are constants with respect to adapting $\bB_K, \bA_K$ with gradient updates. 
    $\bC_Q^{(1)}, \bC_Q^{(2)}, \bC^{(3)}$ are constants with respect to adapting $\bB_Q, \bA_Q$ with gradient updates.
\end{remark}
Therefore, the full LoRA adaptation loss in \cref{def:generic_attention_lora_general} becomes 
\begin{align}
\label{eqn:general_generic_lora_loss_single}
\min _{\substack{ \\\bB_K, \bB_Q \subset \mathbb{R}^{d \times r} \\ \bA_K, \bA_Q \subset \mathbb{R}^{r \times d}}}
\left\|\bD^{-1} \exp \left\{\bX^{(Q)}\left(\bW_Q^\star+\bB_Q \bA_Q\right)\left(\bW_K^\star+\bB_K \bA_K\right)^{\top}\left(\bX^{(K)}\right)^{\top}\right\} \bX^{(V)} \bW_V^\star-\bY\right\|_F^2,
\end{align}
where 
$\bD 
=  \diag\big(\exp\big(\bC_K^{(1)}(\bW_K^\star+\bB_K \bA_K)^{\top}(\bC_K^{(2)})^{\top}\big) \one_L\big)
= \diag\big(\exp \big(\bC_Q^{(1)}(\bW_Q^\star+\bB_Q \bA_Q)(\bC_Q^{(2)})^{\top}\big) \one_L\big) \in\mathbb{R}^{L \times L}   $.  

Similar to \cref{sec:special}, we introduce the following problem to characterize all possible  gradient computation of  \eqref{eqn:general_generic_lora_loss_single}, and arrive similar results as \cref{sec:special}: 
almost linear algorithm for \cref{def:ALoRAGC_general}.

\begin{problem}[Approximate LoRA Gradient Computation ($\mathsf{ALoRAGC}(L,d,r,\epsilon)$)]
\label{def:ALoRAGC_general}
Assume all numerical values be in $\log(L)$ bits encoding.
Let $\calL$ follow \eqref{eqn:general_generic_lora_loss_single},  $\epsilon>0$,  and $\norm{\bZ}_\infty\coloneqq\max_{i,j}\abs{Z_{ij}}$.
The problem of approximating gradient computation of optimizing \eqref{eqn:general_generic_lora_loss_single} is to find four surrogate gradient matrices $\{\Tilde{\bG}^{(A)}_{\mu}\in\R^{d\times r},\Tilde{\bG}^{(B)}_{\mu}\in\R^{r\times d}\}_{\mu=K,Q}$ such that
\begin{align*}
\max\big( \big\{\big\|\Tilde{\underline{\bG}}^{(B)}_\mu-\pdv{\calL}{{\underline{\bB}}_Q}\big\|_{\infty},
\big\|\Tilde{\underline{\bG}}^{(A)}_\mu-\pdv{\calL}{{\underline{\bA}}_Q}\big\|_{\infty}\big\}_{\mu=K,Q}\big) \le \epsilon.    
\end{align*}
\end{problem}

\begin{theorem}[Main Result: Existence of Almost Linear Time $\mathsf{ALoRAGC}$]
\label{thm:main_general}
Let $\Gamma=o(\sqrt{\log L})$.
Suppose all numerical values are in $\mathrm{O}(\log L)$-bits encoding. 
For $\mu = Q,K$, let $\bW_{\mu} = \bW_{\mu}^\star+\bB_{\mu} \bA_{\mu} \in \R^{d\times d}$. 
If $\norm{\bC_{\mu}^{(1)} \bW_{\mu}}_{\infty} \leq \Gamma$ and $\norm{\bC_{\mu}^{(2)}}_{\infty} \leq \Gamma$ for both $\mu=Q,K$,
then there exists a $L^{1+o(1)}$ time algorithm to solve $\operatorname{ALoRAGC} (L, d=O(\log L), r=L^{o(1)}, \epsilon = 1/ \poly(L))$ (i.e., \cref{def:ALoRAGC_general}) up to $1 /\poly(L)$  accuracy.
In particular, this algorithm outputs  gradient matrices $\{\Tilde{\bG}^{(A)}_{\mu}\in\R^{d\times r},\Tilde{\bG}^{(B)}_{\mu}\in\R^{r\times d}\}_{\mu=K,Q}$ such that 
\begin{align*}
\max\big(\big\{\big\|\pdv{\calL}{\underline{\bB}_{\mu}}-\Tilde{\underline{\bG}}^{(A)}_{\mu}\big\|_{\infty} ,\big\|\pdv{\calL}{\underline{\bA}_{\mu}}-\Tilde{\underline{\bG}}^{(A)}_{\mu}\big\|_{\infty}
\big\}_{\mu=K,Q}\big)
\leq 1 / \poly(L).
\end{align*}
\end{theorem}

\begin{proof}
    See \cref{proof:thm:main_general} for a detailed proof.
\end{proof}

\section{Norm-Based Phase Transition in Efficiency}
\label{sec:phase_trans}

In this section,
we characterize the computational limits of all possible efficient algorithms of $\mathsf{ALoRAGC}$, via fine-grained reduction under the Strong Exponential Time Hypothesis (SETH).

\textbf{Strong Exponential Time Hypothesis (SETH).}
\citet{ip01} introduce the Strong Exponential Time Hypothesis (SETH) as a stronger form of the $\mathtt{P} \neq \mathtt{NP}$ conjecture.
It suggests that our current best $\mathtt{SAT}$ algorithms are  optimal and is a popular conjecture for proving fine-grained lower bounds for a wide variety of algorithmic problems \cite{williams2018some,williams2013finding,cygan2016problems}.
\begin{hypothesis}[SETH]
\label{hyp:seth}
For every $\epsilon > 0$, there is a positive integer $k \geq 3$ such that $k$-$\mathtt{SAT}$ on formulas with $n$ variables cannot be solved in $\calO(2^{(1-\epsilon )n})$ time, even by a randomized algorithm.
\end{hypothesis}

\blue{Our primary technique involves casting the $\mathsf{ALoRAGC}$ problem (\cref{def:ALoRAGC}) as a fine-grained reduction under SETH, from the hardness result of fast attention approximation algorithm  \cite{alman2023fast}.
}
For simplicity of analysis, we consider the special case \hyperref[item:C1]{(C1)}.
\begin{theorem}[Inefficient Threshold]
\label{thm:main_eff}
    Let $\kappa : \mathbb{N} \to \mathbb{N}$ by any function with $\kappa(L) = \omega(1)$ and $\kappa(L) = o(\log L)$.
    Let $\Gamma=O(\sqrt{\log L}\cdot \kappa(L))$.
    Assuming \cref{hyp:seth},  there is no algorithm running in time $O(L^{2 - \delta})$ for any constant $\delta>0$ for  $\mathsf{ALoRAGC}(L,d=O(\log L),r<d,\epsilon)$, i.e., \cref{def:ALoRAGC_special}, subject to \eqref{eqn:special_generic_lora_loss_single},
    even in the case where the input and weight matrices satisfy $\|\bX^{(K)}\bW_K^\star\|_\infty \leq \Gamma$, $ \|\alpha\bX^{(Q)}_i\bB_{Q}\bA_{Q}/r\|_\infty \leq \Gamma$,  $ \bY=0$ and $\epsilon = O((\log L)^{-4})$.
\end{theorem}

\begin{proof}[Proof Sketch]
    Firstly, we recall the hardness of sub-quadratic \textbf{Att}ention \textbf{G}radient \textbf{C}omputation approximation, i.e., $\mathsf{AttLGC}$ from  \cite{alman2024fine} (defined in \cref{def:AAttLGC}).
    This serves as a reference point for the complexity we anticipate for $\mathsf{ALoRAGC}$ defined in \cref{def:ALoRAGC_special}. 
    We then proceed with a reduction from problem $\mathsf{AttLGC}$  to problem $\mathsf{ALoRAGC}$. 
    Essentially, 
    by showing that $\mathsf{AttLGC}$ is at least as hard as $\mathsf{ALoRAGC}$, and then showing how to solve $\mathsf{AttLGC}$ using a solution to $\mathsf{ALoRAGC}$, we establish the hardness of $\mathsf{ALoRAGC}$.
    See for \cref{sec:lowerB} for a detailed proof.
\end{proof}

\begin{remark}
\cref{thm:main_eff} suggests an efficiency threshold for $\Gamma$.
Only below this threshold are efficient algorithms for $\mathsf{ALoRAGC}$ possible.
This is a $\Gamma$-based phase transition behavior in efficiency.
\end{remark}
\begin{remark}
In \cref{thm:main_eff}, 
we show that even the simplest single-data-point case with $\bY=0$ is hard.
Hence, our result also applies to the special case \hyperref[item:C1]{(C1)} (i.e., \cref{def:ALoRAGC_special}) and general case \hyperref[item:C2]{(C2)} (i.e., \cref{def:ALoRAGC_general}).
Specifically,
it is evident that computing the gradient for multiple data points (whether the full gradient or a stochastic mini-batch gradient) is \textit{at least} as hard as for a single data point.
The hardness follows trivially.  
\end{remark}

\section{Proof-of-Concept Experiments}
\label{sec:exp}

\begin{wrapfigure}{r}{0.48\textwidth}
\vspace{-1em}
    \centering
    \captionof{table}{\small\textbf{Training Time (Per Epoch) Comparison between LoRA on ``Standard vs. Outlier-Free'' Transformers for 3 OPT Model Sizes.} We perform full LoRA fine-tuning on $W_K, W_Q, W_V$ of the attention heads in \underline{O}pen \underline{P}retrained \underline{T}ransformers (OPTs) \cite{zhang2022opt}. Our results show that, with norm-bound control, Outlier-Free Transformers \cite{hu2024outlier} are 5.5\% faster for OPT-125M, 13.1\% faster for OPT-350M, and 33.3\% faster for OPT-1.3B.}
    \vspace{-.5em}
    \resizebox{.48\textwidth}{!}{%
        \begin{tabular}{lcc}
        \toprule
        \textbf{Model} & \textbf{Standard Transformer} & \textbf{Outlier-Free Transformer} \\
        \midrule
        OPT-125M & 58 mins & \cellcolor{LightCyan} 55 mins (-5.2\%) \\
        OPT-350M & 69 min & \cellcolor{LightCyan} 61 min (-11.6\%) \\
        OPT-1.3B & 84 min & \cellcolor{LightCyan} 63 min (-25.0\%) \\
        \bottomrule
        \end{tabular}
    }
    \label{tab:training_time_comparison}
    \vspace{-1.5em}
\end{wrapfigure}

Here we provide minimally sufficient numerical results to back up our theory.
For generality, we consider the full LoRA fine-tuning on $W_K,W_Q,W_V$ as analyzed in \cref{sec:general}.

\begin{wrapfigure}{r}{0.4\textwidth}
\vspace{-2em}
    \centering
    \includegraphics[width=\linewidth]{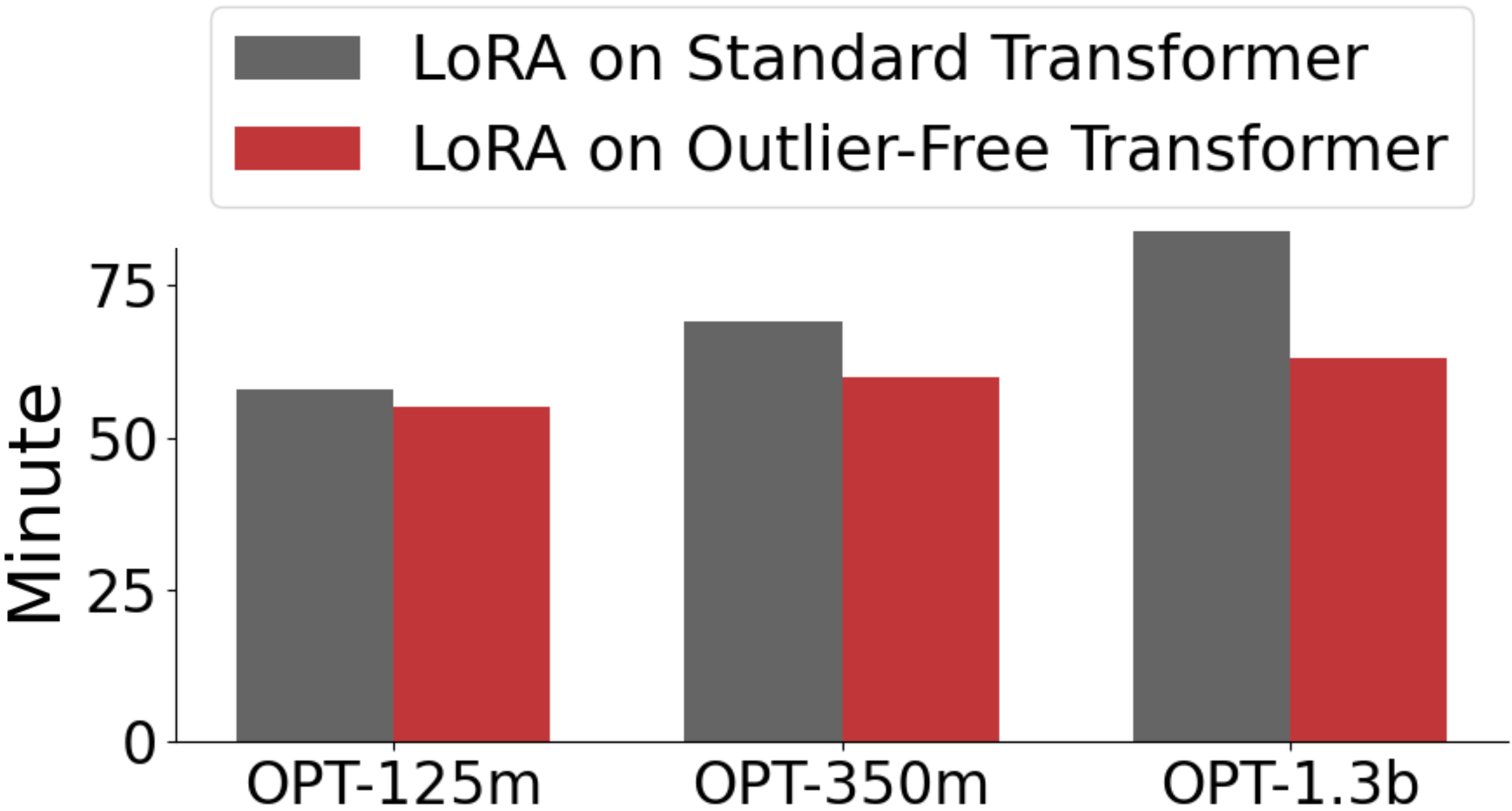}
    \caption{}
    \label{fig:lora_time}
    \vspace{-2em}
\end{wrapfigure}

\textbf{Objective: Control Norms of Attention Heads' Pretrained Weights to Achieve Speedup.}
We use the outlier-removing transformer architecture proposed by \citet{hu2024outlier} to showcase the efficiency gains from controlling the norms of $\{\|W_\mu\|, \|A_\mu\|, \|B_\mu\|\}_{\mu=K,Q,V}$.
This type of architectures bounds these norms  by preventing extreme weight values inherited from the pretraining process.

\textbf{Fine-Tuning Task.}
We perform cross-modality fine-tuning on 3 sizes of the \underline{O}pen \underline{P}retrained \underline{T}ransformer (OPT) models \cite{zhang2022opt}: OPT125M, OPT350M and OPT1.3B.
Specifically,
we adapt OPT language models to speech data, creating a SpeechLM (Speech Language Model) with both text and speech modalities, following \cite{maiti2024voxtlm,wu2024fast}.

\textbf{Pretrianed Model Setup.}
We test our theory on three OPT model sizes: OPT125M, OPT350M, and OPT1.3B. 
Each model size has two versions: one with standard transformers \cite{vaswani2017attention} and another with outlier-removing (outlier-free) transformers \cite{hu2024outlier}. 
The training process for all OPT models follows \cite{hu2024outlier}.

\textbf{LoRA Setup.}
Following the original LoRA settings \cite{hu2021lora}, we fine-tune the models using a rank of $r=128$ and an alpha value of $\alpha=256$.

\textbf{Data.}
We use the LibriLight dataset \cite{kahn2020libri} for fine-tuning. LibriLight contains 60,000 hours of audiobook recordings from 7,000 speakers, totaling 12 million utterances.

\textbf{Computational Resource.}
We conduct all experiments using 4 NVIDIA A100 GPU with 80GB of memory. 
Our code are based on standard PyTorch and the Hugging Face Transformer Library.

\textbf{Efficiency Results: Training Time Comparison.}
To demonstrate the efficiency benefits of norm control suggested by \cref{thm:main_special,thm:main_general,thm:main_eff}, we compare the training speed of the two architectures. 
In \cref{tab:training_time_comparison} and \cref{fig:lora_time}, we report the training time per epoch for both architectures across three model sizes. 
Our results indicate that the Outlier-Free Transformer 
is 5.5\% faster for OPT-125M, 13.1\% faster for OPT-350M, and 33.3\% faster for OPT-1.3B.

These numerical results align with our theory: proper normalization of weights and inputs enhances LoRA training efficiency.
Notably, we observe greater computational gains in larger models.

\section{Discussion and Concluding Remarks}
\label{sec:conclusion}
We study the computational limits of the Low-Rank Adaptation (LoRA) for transformer-based model finetuning using fine-grained complexity theory (i.e., under \cref{hyp:seth}).
Our main contribution is the proof of the existence of almost linear approximation
algorithms for LoRA adaptation on transformer-based models.
We accomplish this by utilizing the hierarchical low-rank structures of
LoRA gradients (\cref{lemma:approx_f,lemma:approx_q,lemma:approx_c}) and approximating the gradients with a series of chained low-rank approximations (\cref{lemma:approx_p1,lemma:approx_p2}).
To showcase our theory, 
we establish such almost linear approximation for both partial (\cref{thm:main_special}) and full LoRA adaptions (\cref{thm:main_general}) of attention weights.
In addition, we identify a phase transition behavior in the efficiency of all possible variants of LoRA (\cref{thm:main_eff}) by adjusting the norm upper-bound $\Gamma$ of input, pretrained, and adaptor weights. 
Specifically, we establish an ``inefficiency threshold'' for $\Gamma$, only below which adapting transformer-based models with LoRA in $L^{2-o(1)}$ (sub-quadratic) time is possible.

\begin{remark}[General Case:  Full LoRA Adaptation on $W_K,W_Q,W_V$]
We defer the analysis of full LoRA on transformer (adapting both $W_K,W_Q,W_V$ matrices) to \cref{sec:general} due to page limit.
\end{remark}

\begin{remark}[Insights for Practitionars: Necessary Conditions for Efficient and Robust LoRA]\label{remark:practical}
This work is about LoRA on transformer models. 
Therefore, the computational bottleneck is by design $\mathcal{O}(L^2)$ (see \cref{sec:quadratic_backprop} for discussions and a proof.)
In this regard, our work provides in-depth analysis to address this $\mathcal{O}(L^2)$ bottleneck and provides useful insights and guidance for designing efficient LoRA algorithms and methods  with precision guarantees:
\begin{itemize}
    \item \textbf{\cref{thm:main_eff}: Necessary Conditions for Subqudratic Time LoRA.}
    Proper normalization of the composed norms, e.g., $\|X^{(K)} W^\star_K\|\le \Gamma$ and $\|\alpha X_i^{(Q)} B_{Q} A_{Q} / r\|\le \Gamma$ with $\Gamma=\calO(\sqrt{\log L}\cdot \kappa(L))$.

    \item 
    \textbf{\cref{thm:main_special,thm:main_general}: Necessary Conditions for Almost Linear Time LoRA.}
    Proper normalization of the composed norms, e.g.,

    \begin{itemize}
        \item For partial LoRA on $W_Q,W_V$ 
        (\cref{thm:main_special}): $\norm{\frac{\alpha}{r} \bX^{(Q)}\bW}_{\infty} \leq \Gamma$ and 
        $\norm{\bX^{(K)} \bW_K^\star}_{\infty} \leq \Gamma$ with $\Gamma=o(\sqrt{\log L})$.
        \item For full LoRA on $W_K,W_Q,W_V$ 
        (\cref{thm:main_general}):
        $\norm{\bX^{(Q)}\left(\bW_Q^\star+\frac{\alpha}{r} \bB_Q \bA_Q\right) \bW_K}_{\infty} \leq \Gamma$,
        $\norm{\bX^{(K)}} \leq \Gamma$,
        $\norm{\bX^{(Q)} \bW_Q} \leq \Gamma$, and $\norm{\bX^{(K)}\left(\bW_K^\star+\frac{\alpha}{r} \bB_K \bA_K\right)}_{\infty} \leq \Gamma$ with $\Gamma=o(\sqrt{\log L})$.
    \end{itemize}
\vspace{-1em}
\end{itemize}
Suitable normalization of the composed norms can be implemented using pre-activation layer normalization \cite{xiong2020layer, wang2019learning} to control $\|X\|$, or outlier-removing attention activation functions \cite{hu2024outlier} to control $\{\|W_\mu\|, \|A_\mu\|, \|B_\mu\|\}_{\mu=K,Q}$.
On one hand, our findings provide formal justifications for these methods. 
On the other hand, these necessary conditions also motivate the design of future efficient methods with minimal model and data assumptions.

\end{remark}

\begin{remark}[Self- and Cross-Attention]
We emphasize that all these results hold for not only self-attention but also cross-attention due to our generic problem setting (\cref{def:generic_attention} and \cref{remark:generic_att}).
\end{remark}

\textbf{Proof-of-Concept Experiments.}
We provide numerical results to justify our theory in \cref{sec:exp}.

\textbf{Limitations.}
We identify necessary conditions for fast LoRA methods, not sufficient conditions.
Therefore, our results do not lead to direct implementations.
This limitation is inherent to hardness results \cite{toolkit2013lecture}.
However, as discussed above, we expect our findings to provide valuable insights for future efficient LoRA implementations in both forward and backward computations.

\textbf{Impact Statement.}
This theoretical work
aims to elucidate the foundations of large transformer-based foundation models and is not expected to have negative social impacts.

\textbf{Related Works.}
We defer the discussion of related works to \cref{sec:related_works} due to page limit.

\clearpage

\section*{Acknowledgments}
JH thanks Mimi Gallagher, Sara Sanchez, Dino Feng, and Andrew Chen for enlightening discussions; Yen-Ju Lu, Shang Wu, Robin Luo, and Jiahao Yu for collaborations on related topics; Shang Wu and authors of \cite{wu2024fast} for assistance with numerical experiments; and the Red Maple Family for their support. 
The authors also thank the anonymous reviewers and program chairs for their constructive comments.

JH is partially supported by the Walter P. Murphy Fellowship.
HL is partially supported by NIH R01LM1372201, AbbVie and Dolby.
EJK thanks the National Center for Theoretical Sciences of Taiwan for funding (112-2124-M-002-003).
This research was supported in part through the computational resources and staff contributions provided for the Quest high performance computing facility at Northwestern University which is jointly supported by the Office of the Provost, the Office for Research, and Northwestern University Information Technology.
The content is solely the responsibility of the authors and does not necessarily represent the official
views of the funding agencies.

\def\arxivfont{\rm}
\bibliographystyle{plainnat}

\bibliography{refs}

\newpage
\appendix
\label{sec:append}
\part*{Appendix}
{
\setlength{\parskip}{-0em}
\startcontents[sections]
\printcontents[sections]{ }{1}{}
}

\section{Related Works}
\label{sec:related_works}

\paragraph{Fine-Grained Complexity.} 
The Strong Exponential Time Hypothesis (SETH) is a conjecture in computational complexity theory that posits solving the Boolean satisfiability problem (SAT) for $n$ variables requires time $2^n$ in the worst case, up to sub-exponential factors \cite{ip01}. 
It extends the Exponential Time Hypothesis (ETH) by suggesting that no algorithm can solve $k$-SAT in $O(2^{(1-\epsilon)n})$ time for any $\epsilon > 0$ \cite{calabro2009complexity}. 
SETH has significant implications for the hardness of various computational problems, as proving or disproving it would greatly enhance our understanding of computational limits \cite{williams2018some,williams2013finding}.

In essence, SETH is a stronger form of the $\mathtt{P} \neq \mathtt{NP}$ conjecture, suggesting that our current best $\mathtt{SAT}$ algorithms are optimal. 
It states as follows:
\begin{hypothesis}[SETH]
For every $\epsilon > 0$, there is a positive integer $k \geq 3$ such that $k$-$\mathtt{SAT}$ on formulas with $n$ variables cannot be solved in $\calO(2^{(1-\epsilon )n})$ time, even by a randomized algorithm.
\end{hypothesis}
SETH is widely used for establishing fine-grained lower bounds for various algorithmic challenges, including $k$-Hitting Set and $k$-NAE-SAT \cite{williams2018some,cygan2016problems}. 
This conjecture is crucial in deriving conditional lower bounds for many significant problems that otherwise have polynomial-time solutions in diverse fields such as pattern matching \cite{cw19, bk18, bgl17, bm16, bi15,  b14, aww14}, graph theory \cite{dlw22, cwx22, abh+18, gikw18, kt18, rvw13}, and computational geometry \cite{km20, w18a, r18, c18, bbk+16}.

Based on this conjecture, our study employs fine-grained reductions under SETH to explore the computational limits of Low-Rank Adaptation (LoRA).
Previous research in fine-grained reductions includes the work by \citet{bis17}, who examine the computational complexity of various Empirical Risk Minimization problems, such as kernel SVMs and kernel ridge. \citet{alman2020algorithms} investigate the effectiveness of spectral graph theory on geometric graphs within the constraints of SETH. 
\citet{aggarwal2022optimal} address the computational limitations of Batch Gaussian Kernel Density Estimation. 
Expanding on these studies, \citet{gu2024conv,gu2024tensor,as23_tensor,alman2023fast} explore transformer attention and introduced a tensor generalization.
\citet{alman2024fundamental} establish the fundamental limitations on subquadratic alternatives to softmax transformers. 
\citet{hu2024computational} show that efficient dense associative memory a.k.a. modern Hopfield models and corresponding networks also need bounded query and key patterns for sub-quadratic time complexity.
Compared to existing works,
this work is, to the best of our knowledge, the first analysis of computational limits for parameter-efficient fine-tuning of large foundation models \cite{hu2021lora}.

\paragraph{Low-Rank Adaptation (LoRA).}
In this paper,
we focus on LoRA \cite{hu2021lora}, a method that leverages low-rank matrices to approximate updates to the weights of neural models. 
Various extensions of LoRA have been proposed to address different challenges in model training and deployment. 
For instance, DoRA \cite{liu2024dora} focus on enhanced parameter efficiency.
QLoRA \cite{dettmers2024qlora}, LoftQ \cite{li2024loftq}, QA-LoRA \cite{xu2024qalora}, and LQ-LoRA \cite{guo2024lqlora} focus on both memory and parameter efficiency in model compression and quantization. 
Additionally, DyLoRA \cite{li2020dylora}, AdaLoRA \cite{zhang2023adaptive}, and SoRA \cite{ding2023sparse} focus on dynamically determining the optimal rank $r$ for LoRA implementations. 
LoRAHub \cite{huang2023lorahub} focus on multi-task finetuning.
LoRA+ \cite{hayou2024lora+} focus on efficient feature learning. 
Despite the methodological and empirical successes, the theoretical side is relatively underdeveloped. 
While \citet{zeng2024the} explore the expressiveness of LoRA from a universal-approximation perspective, and \citet{hayou2024lora+} investigate the optimal adapter learning rate with respect to large model width, to the best of our knowledge, no existing analysis focuses on the computational limits of LoRA. 
Therefore, this work provides a timely theoretical analysis of LoRA's computational limits, aiming to advance efficient finetuning of large foundation models in terms of both parameter usage and computational time.

\paragraph{Outliers in Attention Heads.}
Our results indicate that outliers (e.g., large $\|\bX\bW^\star\|$ and $\|X\bW^\star + \alpha \bX\bB\bA/r\|$) in attention heads hamper LoRA efficiency and performance. 
This outlier effect is well-known in pretraining large foundation models for its negative impact on models' quantization performance \cite{sun2024massive}.
For pretraining, prior works identify the existence of no-op tokens as the main source: tokens with small value vectors tend to receive significantly large attention weights \cite{hu2024outlier, bondarenko2024quantizable}. 
Specifically, \citet{hu2024outlier} interpret this outlier effect as inefficient \textit{rare} memory retrieval from the associative memory/modern Hopfield model perspective \cite{wu2024uniform, wu2023stanhop,xu2024bishop,hu2025provably,hu2024nonparametric,hu2024computational,hu2023sparse} and propose the outlier-efficient Hopfield layer for transformer-based large models, demonstrating strong empirical performance and theoretical guarantees. 
The advantages of controlling outliers in the attention heads of transformer-based large foundation models are also emphasized in various theoretical studies \cite{gu2024conv, gu2024tensor, alman2024fine, as23_tensor, alman2023fast, gao2023fast}.
Yet, to the best of our knowledge, there is no existing work on outliers in LoRA fine-tuning. 
This is the first work establishing that the LoRA adaptor weights might lead to performance and efficiency degradation due to their additive nature: $\|X\bW^\star + \alpha \bX\bB\bA/r\|$.

\clearpage
\section{Proofs of \texorpdfstring{\cref{sec:special}}{}}
\label{sec:proofs}

\subsection{Proof of \texorpdfstring{\cref{lemma:special1}}{}}
\label{proof:lemma:special1}

\begin{proof}[Proof of \cref{lemma:special1}]
    With LoRA loss   \eqref{eqn:special_generic_lora_loss_single},
    we have
   \begin{align*}
    \frac{\d \mathcal{L}(\underline{\bW})}
    {\d \underline{\bW}}
   =\sum_{\underline{j}=1}^L \sum_{i=1}^d \frac{\d }
   {\d \underline{\bW}_i}
   \(\half c(\underline{\bW})_{\underline{j}, i}^2 \).  
   \end{align*}

    Note that for each $\underline{j}\in[L]$ and $i\in[d]$,
\begin{align*}
    &~
    \frac{\d }    {\d \underline{\bW}_i}    \(\half c(\underline{\bW})_{\underline{j}, i}^2 \)
    \annot{By \eqref{eqn:special_generic_lora_loss_single}} \\
    = &~
    c(\underline{\bW})_{\underline{j}, i}  \dv{\left\langle f(\underline{\bW})_{\underline{j}}, \bC^{(3)}[\cdot, i]\right\rangle}{\underline{\bW}_i} 
    \annot{By \cref{def:special_c}} \\
    = &~
    c(\underline{\bW})_{\underline{j}, i}  \left\langle \dv{f(\underline{\bW})_{\underline{j}}}{\underline{\bW}_i}, \bC^{(3)}[\cdot, i] \right\rangle 
    \\
    = &~ 
    c(\underline{\bW})_{\underline{j}, i}  \left\langle \dv{\left(\alpha^{-1}(\underline{\bW})_{\underline{j}} u(\underline{\bW})_{\underline{j}}\right)}{\underline{\bW}_i}, \bC^{(3)}[\cdot, i] \right\rangle 
    \annot{By \cref{def:special_f}} \\
    = &~
    c(\underline{\bW})_{\underline{j}, i}  \left\langle \alpha(\underline{\bW})_{\underline{j}}^{-1}  \underbrace{\dv{u(\underline{\bW})_{\underline{j}}}{\underline{\bW}_i}}_{(I)} - \alpha(\underline{\bW})_{\underline{j}}^{-2} \underbrace{\dv{\alpha(\underline{\bW})_{\underline{j}}}{\underline{\bW}_i}}_{(II)}  u(\underline{\bW})_{\underline{j}}, \bC^{(3)}[\cdot, i] \right\rangle 
    \annot{By product rule and then chain rule}.
\end{align*}

   \begin{itemize}
    \item \textbf{Part (I).}
    We have
    \begin{align*}
   \dv{u(\underline{\bW})_{\underline{j}}}
   {\underline{\bW}_i} 
   = &~
   \dv{\exp \left(\C_{\underline{j}} \underline{\bW}\right)}
   {\underline{\bW}_i} 
   \annot{By \cref{def:special_u}}\\ 
   = &~ \exp \left(\C_{\underline{j}} \underline{\bW}\right) \odot \dv{ \C_{\underline{j}} \underline{\bW}}
   {\underline{\bW}_i} \\
   = &~ \C_{\underline{j}}[\cdot, i] \odot u(\underline{\bW})_{\underline{j}} .
   \annot{By $\dv{\(\C_{\underline{j}} \underline{\bW}\)} {\underline{\bW}_i}
   =
   \frac{\d \C_{\underline{j}} \underline{\bW}}{\d \underline{\bW}_i}
   = \C_{\underline{j}} \cdot \dv
   {\underline{\bW}}
   {\underline{\bW}_i} = \C_{\underline{j}} \cdot \be_i
   = \left(\C_{\underline{j}}\right) 
   [\cdot, i]$}
   \end{align*}

    \item \textbf{Part (II).}
    We have
    \begin{align*}   
    \dv{\alpha(\underline{\bW})_{\underline{j}}}{\underline{\bW}_i}
    &= \dv{\left\langle u(\underline{\bW})_{\underline{j}}, \one_L \right\rangle}{\underline{\bW}_i} 
    \annot{By \cref{def:special_alpha}} \\
    &= \left\langle \C_{\underline{j}}[\cdot, i] \odot u(\underline{\bW})_{\underline{j}}, \one_L \right\rangle 
    \annot{By \cref{def:special_u}} \\
    &= \left\langle \C_{\underline{j}}[\cdot, i], u(\underline{\bW})_{\underline{j}} \right\rangle 
    \annot{By element-wise product identity}.
    \end{align*}
   
   \end{itemize}

   Combining (I) and (II), we get
   \begin{align*}
   & ~ \frac{\d }{\d \underline{\bW}_i}    \(\half c(\underline{\bW})_{\underline{j}, i}^2 \) \\
   = & ~ c(\underline{\bW})_{\underline{j}, i} \[\left\langle \bC^{(3)} [\cdot, i], \C_{\underline{j}}[\cdot, i] \odot f(\underline{\bW})_{\underline{j}} \right \rangle - \left \langle \bC^{(3)} [\cdot, i], f(\underline{\bW})_{\underline{j}} \right \rangle \cdot \left \langle \C_{\underline{j}} [\cdot, i], f(\underline{\bW})_{\underline{j}} \right \rangle\] \\
   = &~ c(\underline{\bW})_{\underline{j}, i}\C_{\underline{j}}^{\top} \left (\operatorname{\diag} \left (f({\underline{\bW}})_{\underline{j}}\right)-f({\underline{\bW}})_{\underline{j}} f({\underline{\bW}})_{\underline{j}}^{\top}\right) \bC^{(3)}[\cdot, i]
   .
   \end{align*}
   This completes the proof.
   \end{proof}

\clearpage
\subsection{Proof of \texorpdfstring{\cref{lemma:special3}}{}}
\label{proof:lemma:special3}
First, we present a helper lemma.
\begin{lemma}
\label{lemma:speical2}
For any $a \in \R$, let $\diag_d(a) \in \R^{d\times d}$ be a $d \times d$ diagonal matrix with all entries equal to $a$.
Let $\bJ_B,\bJ_A \in \R^{{d^2 \times rd}}$ be two matrices such that $\underline{\bW}
    = \underline{\Bar{\bW}}_Q^\star + \bJ_B \underline{\bA}_Q$, and $\underline{\bW}
    = \underline{\Bar{\bW}}_Q^\star + \bJ_A \underline{\bB}_Q$ via
    \begin{align*}
        \bJ_B = \begin{pmatrix}
                \bB_Q &     &        &     \\
                    & \bB_Q &        &     \\
                    &     & \ddots &     \\
                    &     &        & \bB_Q
            \end{pmatrix}, 
        \bJ_A = \begin{pmatrix}
            \diag_d\left(\bA_Q[1,1]\right)   
            & \cdots & 
            \diag_d\left(\bA_Q[r, 1]\right) \\
            \diag_d\left(\bA_Q[1,2]\right)  
            & \cdots & 
            \diag_d\left(\bA_Q[r, 2]\right) \\
            \vdots  &  & \vdots \\
            \diag_d\left(\bA_Q[1, d]\right)
            & \cdots & 
            \diag_d\left(\bA_Q[r, d]\right)
            \end{pmatrix}
    \end{align*}
     The derivatives of loss function \eqref{eqn:special_generic_lora_loss_single} w.r.t. $\bA_Q, \bB_Q$ are therefore
    \begin{align*}
        \pdv{\calL}{\underline{\bA}_Q}
         & = \sum_{\underline{j}=1}^L \sum_{i=1}^d 
         \bJ_B^{\top} c(\underline{\bW})_{\underline{j}, i} \C_{\underline{j}}^{\top}
         \left(\diag\left(f(\underline{\bW})_{\underline{j}}\right)-f(\underline{\bW})_{\underline{j}} f(\underline{\bW})_{\underline{j}}^{\top}\right) \bC^{(3)}[\cdot, i] , \\
        \pdv{\mathcal{L}} {\underline{\bB}_Q}
         & = \sum_{\underline{j}=1}^L \sum_{i=1}^d \bJ_A^{\top} c(\underline{\bW})_{\underline{j}, i} \C_{\underline{j}}^{\top}
         \left(\diag\left(f(\underline{\bW})_{\underline{j}}\right) -f(\underline{\bW})_{\underline{j}} f(\underline{\bW})_{\underline{j}}^{\top}\right) \bC^{(3)} [\cdot, i].
    \end{align*}
\end{lemma}
\begin{proof}
    The proof follows standard chain-rule and \cref{lemma:special1}.
\end{proof}

Then, we prove \cref{lemma:special3}.
\begin{proof}[Proof of \cref{lemma:special3}]
    From \cref{lemma:speical2}, we have
    \begin{align*}
            \pdv{\mathcal{L}} {\underline{\bA}_Q}
            & = \sum_{\underline{j}=1}^L \sum_{i=1}^d \bJ_B^{\top} c(\underline{\bW})_{\underline{j}, i} \C_{\underline{j}}^{\top}
            \left(\diag\left(f(\underline{\bW})_
            {\underline{j}}\right) -f(\underline{\bW})_{\underline{j}} f(\underline{\bW})_{\underline{j}}^{\top}\right) \bC^{(3)} [\cdot, i] \\
            & = \sum_{\underline{j}=1}^L \bJ_B^{\top} \C_{\underline{j}}^{\top}\left(\diag\left(f(\underline{\bW})_{\underline{j}}\right) -f(\underline{\bW})_{\underline{j}} f(\underline{\bW})_{\underline{j}}^{\top}\right) q(\underline{\bW})_{\underline{j}} 
            \annot{By $q(\underline{\bW})
    \coloneqq \bC^{(3)} \(c(\underline{\bW})\)^\sT \in \R^{L\times L}$}
            \\
            & = \sum_{\underline{j}=1}^L \bJ_B^{\top} \C_{\underline{j}}^{\top} p(\underline{\bW})_{\underline{j}} 
            \annot{By \cref{def:p1p2}}
            \\
            & = \vect 
            \left(\bB_Q^{\top}\left(\bC^{(1)}
            \right)^{\top} p(\underline{\bW}) \bC^{(2)}\right).
    \end{align*}
    Similarly,
    \begin{align*}
        \pdv{\mathcal{L}} {\underline{\bB}_Q}
        & = \sum_{j=1}^L \sum_{i=1}^d \bJ_A^{\top} c(\underline{\bW})_{\underline{j}, i} \C_{\underline{j}}^{\top}
        \left(\diag\left(f(\underline{\bW})_
        {\underline{j}}
        \right) -f(\underline{\bW})_{\underline{j}} f(\underline{\bW})_{\underline{j}}^{\top}\right) \bC^{(3)}[\cdot, i]
        \\
        & = \sum_{\underline{j}=1}^{L} \bJ_A^{\top} \C_{\underline{j}}^{\top}
        \left(\diag\left(f(\underline{\bW})_{\underline{j}}\right) -f(\underline{\bW})_{\underline{j}} f(\underline{\bW})_{\underline{j}}^{\top}\right) q(\underline{\bW})_{\underline{j}} 
        \annot{By $q(\underline{\bW})
        \coloneqq \bC^{(3)} \(c(\underline{\bW})\)^\sT \in \R^{L\times L}$}
        \\
        & =\sum_{\underline{j}=1}^L \bJ_A^{\top} \C_{\underline{j}}^{\top} p(\underline{\bW})_{\underline{j}} 
        \annot{By \cref{def:p1p2}}
        \\
        & = \vect\left(\left(\bC^{(1)}\right)^{\top} p(\underline{\bW}) \bA_Q \bC^{(2)}\right).
        \annot{By $\bJ_B^{\top} \bC_{\underline{j}}^{\top}
            =
            \left(\bC^{(1)} \bB_Q \otimes \bC^{(2)}\right)^{\top} $, and $
            \bJ_A^{\top} \bC_{\underline{j}}^{\top}
            =
            \left(\bC^{(1)} \otimes \bA_Q \bC^{(2)}\right)^{\top}$}
    \end{align*}
This completes the proof.
\end{proof}

\subsection{Proof of \texorpdfstring{\cref{lemma:approx_c}}{}}
\label{proof:lemma:approx_c}
\begin{proof}[Proof of \cref{lemma:approx_c}]
Our proof is built on \cite[Lemma~D.2]{alman2023fast}.
    By definitions,
    \begin{align*}
            &~ \left\|\bU_1 \bV_1^{\top} \bC^{(3)}-\bY-c(\underline{\bW})\right\|_{\infty} \\
            = &~  \left\|\bU_1 \bV_1^{\top} \bC^{(3)} -\bY-f(\underline{\bW}) \bC^{(3)} + \bY\right\|_{\infty} 
            \annot{By $c(\underline{\bW}) = f(\underline{\bW}) \bC^{(3)} - \bY$}
            \\
            = &~ \left\|\left(\bU_1 \bV_1^{\top}-f(\underline{\bW})\right) \bC^{(3)}\right\|_{\infty} \\
            \leq &~ 
            \epsilon / \poly(L).
            \annot{By \cite[Lemma~D.2]{alman2023fast}}
    \end{align*}
    This completes the proof.
\end{proof}

\subsection{Proof of \texorpdfstring{\cref{lemma:approx_q}}{}}
\label{proof:lemma:approx_q}
\begin{proof}[Proof of \cref{lemma:approx_q}]
Our proof is built on \cite[Lemma~D.3]{alman2023fast}.

    Let $\tilde{q}(\underline{\bW})$ denote an approximation to $q(\underline{\bW})$.  
    By \cref{lemma:approx_c},  $\bU_1 \bV_1^{\top} \bC^{(3)}-\bY$ approximates $c(\underline{\bW})$ with a controllable error.

    Then, by setting 
    \begin{align*}
    \tilde{q}(\underline{\bW}) = \bC^{(3)}
    \left(\bU_1 \bV_1^{\top} \bC^{(3)}-\bY\right)^{\top},    
    \end{align*} 
    we turn $\tilde{q}(\underline{\bW})$ into some low-rank representation
    \begin{align*}
        \tilde{q}(\underline{\bW})
        = \bC^{(3)}
        \left(\bC^{(3)}\right)^{\top} \bV_1 \bU_1^{\top}-\bC^{(3)} \bY^{\top}.
    \end{align*}
    By $k_1,d=L^{o(1)}$, it is obvious that computing $\underbrace{\left(\bC^{(3)}\right)^{\top}}_{{d\times L}} \underbrace{\bV_1}_{L\times k_1} \underbrace{\bU_1^{\top}}_{k_1\times L} $ only takes $L^{1+o(1)}$ time.

     Then we can explicitly construct 
     $\bU_2, \bV_2 \in \mathbb{R}
     ^{L \times k_2}$ in $L^{1+o(1)}$ time as follows:
     \begin{align*}
     \bU_2 \coloneqq 
     \underbrace{\begin{pmatrix}
         \bC^{(3)}  & -\bC^{(3)}
     \end{pmatrix}}_{L\times (d+d)}\in\R^{L\times k_2},
     \quad
     \bV_2 \coloneqq 
     \underbrace{\begin{pmatrix}
         \bU_1 \bV_1^{\top} \bC^{(3)} &
         \bY 
     \end{pmatrix}}_{L\times (d+d)}\in\R^{L\times k_2},
     \end{align*}
     with
     {$k_2 = 2d = L^{o(1)}$}  
     by $d=O(\log L)$.
    This leads to
    \begin{align*}
        \tilde{q}(\underline{\bW})=
        \begin{pmatrix}
            \bC^{(3)} & -\bC^{(3)}
        \end{pmatrix}
        \begin{pmatrix}
        \left(\bC^{(3)}\right)^{\top} \bV_1 \bU_1^{\top} \\
                \bY^{\top}
        \end{pmatrix}
        =\bU_2 \bV_2^{\top}.
    \end{align*}

    Therefore, 
    for controlling the approximation error, it holds
    \begin{align*}
        \|\tilde{q}(\underline{\bW}) -q(\underline{\bW})\|_{\infty}
        & = \left\|\bC^{(3)}\left(\bU_1 \bV_1^{\top} \bC^{(3)}-\bY\right)^{\top} - \bC^{(3)} \bY^{\top}\right\|_{\infty} 
        \\
        & \leq d\left\|\bC^{(3)}\right\|_{\infty}
        \left\|\bU_1 \bV_1^{\top} \bC^{(3)} - \bY - c(\underline{\bW})\right\|_{\infty} \\
        & \leq \epsilon / \poly (L).
        \annot{By \cref{lemma:approx_c}}
    \end{align*}
    Thus, we complete the proof.
\end{proof}

\clearpage
\subsection{Proof of \texorpdfstring{\cref{lemma:approx_p1}}{}}
\label{proof:lemma:approx_p1}
\begin{proof}[Proof of \cref{lemma:approx_p1}]
    We proceed the proof by 
    constructing low-rank approximation of $p_1(\cdot)$ with decomposing $p_1(\cdot)$ into $f(\cdot)$ and $q(\cdot)$ through tensor formulation, and then approximating $p_1$ part by part.
    
    We denote $\oslash$ for {\textit{column-wise} Kronecker product} such that $\bA \oslash \bB \coloneqq [\bA[\cdot,1] \otimes \bB [\cdot,1] \mid \ldots \mid \bA [\cdot,k_1] \otimes \bB [\cdot,k_1]] \in \R^{L \times k_1k_2}$ for $\bA \in \R^{L \times k_1},\bB \in \R^{L \times k_2}$. 
    
    Let  $\tilde{f}(\underline{\bW}) \coloneqq \bU_1 \bV_1^\sT$ and $\tilde{q}(\underline{\bW}) \coloneqq \bU_2\bV_2^\sT$  denote matrix-multiplication approximations to  $f(\underline{\bW})$ and $q(\underline{\bW})$, respectively.

    For the case of presentation, let $\bU_3 = \overbrace{\bU_1}^{L\times k_1} \oslash \overbrace{\bU_2}^{L\times k_2}$ and 
    $\bV_3 = \overbrace{\bV_1}^{L \times k_1} \oslash \overbrace{\bV_2}^{L \times k_2}$. 
    It holds
    \begin{align*}
        & ~ \left\|\bU_3 \bV_3^{\top}-p_1(\underline{\bW})\right\|_{\infty}
        \nonumber\\
        = & ~\left\|\bU_3 \bV_3^{\top}-f(\underline{\bW}) \odot q(\underline{\bW})\right\|_{\infty}
        \annot{
        By $p_1(\underline{\bW})= f(\underline{\bW}) \odot q(\underline{\bW})$}
        \\
        = & ~\left\|\left(\bU_1 \oslash \bU_2\right)\left(\bV_1 \oslash \bV_2\right)^{\top} - f(\underline{\bW}) \odot q(\underline{\bW}) \right\|_{\infty} \\
        = & ~
        \left\|\left(\bU_1 \bV_1^{\top}\right) \odot \left(\bU_2 \bV_2^{\top}\right)-f(\underline{\bW}) \odot q(\underline{\bW})\right\|_{\infty}
        \\
        = & ~
        \|\tilde{f}(\underline{\bW}) \odot \tilde{q} (\underline{\bW})-f(\underline{\bW}) \odot q(\underline{\bW})\|_{\infty} 
        \\
        \leq & ~
        \|\tilde{f}(\underline{\bW}) \odot \tilde{q}(\underline{\bW})-\tilde{f}(\underline{\bW}) \odot q(\underline{\bW})\|_{\infty}
        +
        \|\tilde{f}(\underline{\bW}) \odot q(\underline{\bW})-f(\underline{\bW}) \odot q(\underline{\bW})\|_{\infty} 
        \annot{By triangle inequality}
        \\
        \leq &~ \epsilon / \poly (L).
        \annot{By \cref{lemma:approx_f} and \cref{lemma:approx_q}}
    \end{align*}
    Computationally, by $k_1,k_2=L^{o(1)}$, computing $\bU_3$ and $\bV_3$ takes $L^{1+o(1)}$ time.
    
    This completes the proof.
\end{proof}

\subsection{Proof of \texorpdfstring{\cref{lemma:approx_p2}}{}}
\label{proof:lemma:approx_p2}
\begin{proof}[Proof of \cref{lemma:approx_p2}]
    By considering the following decomposition through tensor formulation
    \begin{align*}
    p_2(\underline{\bW})_{\underline{j}}
    \coloneqq \overbrace{f\left(\underline{\bW}\right)_{\underline{j}}  \underbrace{f\left(\underline{\bW}\right)_{\underline{j}}^{\top} q(\underline{\bW})_{\underline{j}}}_{(I)}
    }^{(II)},
    \end{align*}
    we approximate the $p_2(\cdot)$  part by part. 
    Specifically, 
    for (I),
    we show its low-rank approximation by observing the low-rank-preserving property of the multiplication between $f(\cdot)$ and  $q(\cdot)$ (from \cref{lemma:approx_f} and \cref{lemma:approx_q}). 
    For (II), 
    we show its low-rank approximation by the low-rank structure of  $f(\cdot)$ and (I).

    \paragraph{Part (I).}
    We define a function $r(\underline{\bW}): \R^{d^2} \to \R^L$ such that the $\underline{j}$-th component $r(\underline{\bW})_{\underline{j}} \coloneqq
    \left(f(\underline{\bW})_{\underline{j}}\right)^{\top} q(\underline{\bW})_{\underline{j}}$ for all $\underline{j}\in[L]$. 
    Let $\tilde{r}(\underline{\bW})$ denote the approximation of $r(\underline{\bW})$ via decomposing  into $f(\cdot)$ and $q(\cdot)$:
\begin{align}
            \tilde{r}(\underline{\bW})_{\underline{j}}
             & \coloneqq \left \langle \tilde{f}(\underline{\bW})_{\underline{j}}, \tilde{q}(\underline{\bW})_{\underline{j}}\right\rangle
             = \left(\bU_1 \bV_1^{\top}\right)[\underline{j}, \cdot] \cdot\left[\left(\bU_2 \bV_2^{\top}\right)[\underline{j}, \cdot]\right]^{\top} 
             \nonumber
             \\
             & = \bU_1[\underline{j}, \cdot] \underbrace{\bV_1^{\top} }_{{k_1\times L}} \underbrace{\bV_2}_{{L\times k_2}}\left(\bU_2[\underline{j}, \cdot]\right)^{\top},
             \label{eqn:U1V1V2U1}
\end{align}
for all $\underline{j}\in[L]$. 
This allows us to write ${p}_2(\underline{\bW}) ={f}(\underline{\bW}) \diag({r}(\underline{\bW}))$ with
    $\diag(\tilde{r}(\underline{\bW}))$ denoting a diagonal matrix with diagonal entries being components of $\tilde{r}(\underline{\bW})$.

    \paragraph{Part (II).}
    With $r(\cdot)$, we approximate $p_2(\cdot)$ with $\tilde{p}_2(\underline{\bW}) = \tilde{f}(\underline{\bW}) \diag (\tilde{r}(\underline{\bW}))$ as follows.

    Since $\tilde{f}(\underline{\bW})$ has low rank representation, and $\diag(\tilde{r}(\underline{\bW}))$ is a diagonal matrix, 
    $\tilde{p}_2(\cdot)$ has low-rank representation by definition.
    Thus, we set $\tilde{p}_2(\underline{\bW}) = \bU_4\bV_4^\sT$ with $\bU_4 = \bU_1$ and $\bV_4 = \diag(\tilde{r}(\underline{\bW})) \bV_1$.
    Then, we bound the approximation error
\begin{align*}
            & ~ \left\|\bU_4 \bV_4^{\top}-p_2(\underline{\bW})\right\|_{\infty} \\
            = & ~  \left\|\tilde{p}_2(\underline{\bW})-p_2(\underline{\bW})\right\|_{\infty}
            \\
            = & ~
            \max _{\underline{j} \in[L]}\left\|
            {
            \tilde{f}(\underline{\bW})_{\underline{j}} \tilde{r}(\underline{\bW})_{\underline{j}}-f(\underline{\bW})_{\underline{j}} r(\underline{\bW})_{\underline{j}}
            }
            \right\|_{\infty} 
            \\
            \leq & ~ \max _{\underline{j} \in[L]}\[\left\|\tilde{f}(\underline{\bW})_{\underline{j}} \tilde{r}(\underline{\bW})_{\underline{j}}-f(\underline{\bW})_{\underline{j}} {r}(\underline{\bW})_{\underline{j}}\right\|_{\infty}+\left\|\tilde{f}(\underline{\bW})_{\underline{j}} \tilde{r}(\underline{\bW})_{\underline{j}}-f(\underline{\bW})_{\underline{j}} r(\underline{\bW})_{\underline{j}}\right\|_{\infty} \]
            \annot{By triangle inequality}
            \\
            \leq & ~ \epsilon / \poly(L).  
\end{align*}
Computationally, computing  $\bV_1^{\top} \bV_2$ takes $L^{1+o(1)}$ time by $k_1,k_2=L^{o(1)}$.

    Once we have $\bV_1^{\top} \bV_2$ precomputed, \eqref{eqn:U1V1V2U1}
    only takes $O(k_1 k_2)$ time for each $\underline{j}\in[L]$.
    Thus, the total time is 
    $O\left(L k_1 k_2\right)=L^{1+o(1)} $.
    Since $\bU_1$ and $\bV_1$  takes $L^{1+o(1)}$ time to construct and $\bV_4 = \underbrace{\diag(\tilde{r}(\underline{\bW}))}_{L\times L} \underbrace{\bV_1}_{L\times k_1}$ also takes $L^{1+o(1)}$ time, $\bU_4$ and $\bV_4$ takes $L^{1+o(1)}$ time to construct.
    
This completes the proof.
\end{proof}

\subsection{Proof of \texorpdfstring{\cref{thm:main_special}}{}}
\label{proof:thm:main_special}
\begin{proof}[Proof of \cref{thm:main_special}]
    By the definitions of matrices $p(\underline{\bW})$ (\cref{lemma:special3}), $p_1(\underline{\bW})$  and $p_2(\underline{\bW})$ (\cref{def:p1p2}),
    we have
        $p(\underline{\bW})=p_1(\underline{\bW})-p_2(\underline{\bW})$.

    By \cref{lemma:special3}, we have 
    \bea\label{eqn:lora_grad_proof}
             \pdv{\calL}{\underline{\bA}_Q}
             =
             \vect\left(\bB_Q^{\top}\left(\bC^{(1)}\right)^{\top} p(\underline{\bW}) \bC^{(2)}\right),
             \quad
             \pdv{\calL}{\underline{\bB}_Q}=\vect\left(\left(\bC^{(1)}\right)^{\top} p(\underline{\bW}) \bA_Q \bC^{(2)}\right).
    \eea
    Firstly, we note that the \textit{exact} computation of  $\bB_Q^{\top}\left(\bC^{(1)}\right)$ and $\bA_Q \bC^{(2)}$ takes $L^{1+o(1)}$ time, by $\bA_Q \in \mathbb{R}^{r \times d}, \bB_Q \in \mathbb{R}^{d \times r}, \bC^{(1)}, \bC^{(2)} \in \mathbb{R}^{L \times d}$. 
    Therefore, to show the existence of $L^{1+o(1)}$ algorithms for \cref{def:ALoRAGC_special},  
    we prove fast low-rank approximations for 
    $\bB_Q^{\top}\left(\bC^{(1)}\right)^{\top} p_1(\underline{\bW}) \bC^{(2)}$ and $\left(\bC^{(1)}\right)^{\top} p_1(\underline{\bW}) \bA_Q \bC^{(2)}$ as follows.
    The fast low-rank approximations for  $-\bB_Q^{\top}\left(\bC^{(1)}\right)^{\top} p_2(\underline{\bW}) \bC^{(2)}$ and $-\left(\bC^{(1)}\right)^{\top} p_2(\underline{\bW}) \bA_Q \bC^{(2)}$ trivially follow.
    
    \textbf{Fast Approximation for $\bB_Q^{\top}\left(\bC^{(1)}\right)^{\top} p_1(\underline{\bW}) \bC^{(2)}$.}
    Using $\tilde{p}_1(\underline{\bW}), \tilde{p_2}(\underline{\bW})$ as the approximations to $p_1(\underline{\bW}), p_2(\underline{\bW})$, by \cref{lemma:approx_p1}, it takes $L^{1+o(1)}$ time to construct $\bU_3, \bV_3 \in \R^{L \times k_3}$ subject to
    \begin{align*}
    \bB_Q^{\top}\left(\bC^{(1)}\right)^{\top} \tilde{p}_1(\underline{\bW}) \bC^{(2)}=\bB_Q^{\top}\left(\bC^{(1)}\right)^{\top} \bU_3 \bV_3^{\top} \bC^{(2)}. 
    \end{align*}

    Then we compute $\overbrace{\bB_Q^{\top}}^{r\times d}\overbrace{\left(\bC^{(1)}\right)^{\top}}^{d\times L} \overbrace{\bU_3}^{L\times k_3}, \overbrace{\bV_3^{\top}}^{k_3\times L} \overbrace{\bC^{(2)}}^{L\times d}$.
    By $r,d,k_1,k_3=L^{o(1)}$, this takes $L^{1+o(1)}$ time. 
    
    Finally we compute $\overbrace{\left(\bB_Q^{\top}\left(\bC^{(1)}\right)^{\top} \bU_3\right)}^{r\times k_3}\overbrace{\left(\bV_3^{\top} \bC^{(2)}\right)}^{k_3\times d}$.
    By $r,d,k_1,k_3=L^{o(1)}$, this takes $L^{1+o(1)}$ time. 
    So, overall running time is still $L^{1+o(1)}$.

    \paragraph{Fast Approximation for $\left(\bC^{(1)}\right)^{\top} p_1(\underline{\bW}) \bA_Q \bC^{(2)}$.}
    Similarly, computing $\left(\bC^{(1)}\right)^{\top} p_1(\underline{\bW}) \bA_Q \bC^{(2)}$ takes $L^{1+o(1)}$ time. 
    
    \paragraph{Fast Approximation for \eqref{eqn:lora_grad_proof}.}
    Notably, above results hold for both $p_2(x)$ and $p_1(x)$. 
    Therefore, computing $\bB_Q^{\top}\left(\bC^{(1)}\right)^{\top} p(\underline{\bW}) \bC^{(2)},\left(\bC^{(1)}\right)^{\top} p(\underline{\bW}) \bA_Q \bC^{(2)}$ also takes $L^{1+o(1)}$ time.

    \paragraph{Approximation Error.}
    We have
\begin{align*}
        &~ \left\|\pdv{\calL}{\underline{\bA}_Q} -\tilde{\bG}^{(A)}_Q\right\|_{\infty}
        \\
        = &~ \left\|\vect\left(\bB_Q^{\top}\left(\bC^{(1)}\right)^{\top} p(\underline{\bW}) \bC^{(2)}\right) -\vect\left(\bB_Q^{\top}\left(\bC^{(1)}\right)^{\top} \tilde{p}(\underline{\bW}) \bC^{(2)}\right)\right\|_{\infty}
        \annot{By \cref{lemma:special3}}
    \\
    = &~ \left\|\left(\bB_Q^{\top}\left(\bC^{(1)}\right)^{\top} p(\underline{\bW}) \bC^{(2)}\right) -\left(\bB_Q^{\top}\left(\bC^{(1)}\right)^{\top} \tilde{p}(\underline{\bW}) \bC^{(2)}\right)\right\|_{\infty} 
    \annot{By definition, $\norm{\bA}_{\infty} \coloneqq \max_{i,j} \abs{A_{ij}}$ for any matrix $\bA$} 
    \\
    \leq &~ \left\|\left(\bB_Q^{\top}\left(\bC^{(1)}\right)^{\top} \left(p_1(\underline{\bW}) -\tilde{p}_1(\underline{\bW})\right) \bC^{(2)}\right)\right\|_{\infty}
    + \left\|\left(\bB_Q^{\top}\left(\bC^{(1)}\right)^{\top} \left(p_2(\underline{\bW}) -\tilde{p}_2(\underline{\bW})\right) \bC^{(2)}\right)\right\|_{\infty} 
    \annot{By \cref{def:p1p2} and triangle inequality}
    \\
    \leq &~ \left\|\bB_Q\right\|_{\infty}
    \left\|\bC^{(1)}\right\|_{\infty}\left\|\bC^{(2)}\right\|_{\infty}
    \left(\left\|\left(p_1(\underline{\bW})-\tilde{p}_1(\underline{\bW})\right)\right\|_{\infty} + 
    \left\|\left(p_2(\underline{\bW})-\tilde{p}_2(\underline{\bW})\right)\right\|_{\infty}\right)
    \annot{By the sub-multiplicative property of $\infty$-norm}
    \\
    \leq &~  \epsilon / \poly(L).
    \annot{By \cref{lemma:approx_p1} and \cref{lemma:approx_p2}}
\end{align*}
Similarly, it holds
\begin{align*}
    &~ \left\|\pdv{\calL}{\underline{\bB}_Q} -\tilde{\bG}^{(B)}_Q\right\|_{\infty}
        \\
    = &~  \left\|
    \vect\left(\left(\bC^{(1)}\right)^{\top} p(\underline{\bW}) \bA_Q\bC^{(2)}\right) -\vect\left(\bB_Q^{\top}\left(\bC^{(1)}\right)^{\top} \tilde{p}(\underline{\bW}) \bA_Q\bC^{(2)}\right)\right\|_{\infty}
    \\
    = &~  \left\|
    \left(\left(\bC^{(1)}\right)^{\top} p(\underline{\bW}) \bA_Q\bC^{(2)}\right) -\left(\left(\bC^{(1)}\right)^{\top} \tilde{p}(\underline{\bW}) \bA_Q\bC^{(2)}\right)\right\|_{\infty} 
    \\
    \leq &~  \left\|
    \left(\left(\bC^{(1)}\right)^{\top}\left(p_1(\underline{\bW}) -\tilde{p}_1(\underline{\bW})\right) \bA_Q\bC^{(2)}\right)\right\|_{\infty}+\left\|\left(\left(\bC^{(1)}\right)^{\top}\left(p_2(\underline{\bW}) -\tilde{p}_2(\underline{\bW})\right) \bA_Q\bC^{(2)}\right)\right\|_{\infty} 
    \\ \leq &~ \left\|\bA_Q\right\|_{\infty}\left\|\bC^{(1)}\right\|_{\infty}\left\|\bC^{(2)}\right\|_{\infty}\left(\left\|\left(p_1(\underline{\bW}) -\tilde{p}_1(\underline{\bW})\right)\right\|_{\infty}+\left\|\left(p_2(\underline{\bW}) -\tilde{p}_2(\underline{\bW})\right)\right\|_{\infty}\right)
    \\
    \leq &~  \epsilon / \poly(L).
\end{align*}
Setting $\epsilon = 1/ \poly(L)$ , we complete the proof.
\end{proof}

\clearpage
\section{Proof of \texorpdfstring{\cref{thm:main_general}}{}}
\label{proof:thm:main_general}
We prepare the proof with the following definitions and lemmas.

Similar to \cref{sec:special},
we introduce the $u(\cdot),\alpha(\cdot),f(\cdot),c(\cdot)$ notations.
Notably, we introduce them for both $K$ and $Q$ because there are two sets of adaptors: $\bB_K, \bA_K$ and $\bB_Q, \bA_Q$.

\begin{definition}[$u(\cdot)$]
\label{def:general_u}
Let $\C
^K\coloneqq \bC_K^{(1)}\otimes \bC_K^{(2)}$, and $\C
^Q\coloneqq \bC_Q^{(1)}\otimes \bC_Q^{(2)}$.
Recall that $\C^K_{\underline{j}},\C^Q_{\underline{j}}\in \R^{L\times d^2}$ are sub-block matrices of $\C^K,\C^Q$. 
For every $\underline{j} \in [L]$,
we define two functions $u_K(\underline{\bW})_{\underline{j}} , u_Q(\underline{\bW})_{\underline{j}}: \R^{d^2} \to \R^L$:
$ u_K(\underline{\bW})_{\underline{j}} 
\coloneqq  \exp( \C^K_{\underline{j}} \underline{\bW} ) \in \R^L $ and $ u_Q(\underline{\bW})_{\underline{j}} \coloneqq  \exp( \C^Q_{\underline{j}} \underline{\bW} ) \in \R^L$.
\end{definition}

\begin{definition}[$\alpha(\cdot)$]\label{def:general_alpha}
Let $\C^K \coloneqq \bC_K^{(1)} \otimes \bC_K^{(2)}$, and $\C^Q \coloneqq \bC_Q^{(1)} \otimes \bC_Q^{(2)}$.
Recall that $\C^K_{\underline{j}} , \C^Q_{\underline{j}} \in \R^{L\times d^2}$ are sub-block matrices of $\C^K , \C^Q$. 
For every index $\underline{j} \in [L]$, we define two functions $\alpha_Q(\underline{\bW})_{\underline{j}} , \alpha_K(\underline{\bW})_{\underline{j}}: \R^{d^2} \rightarrow \R$:
$\alpha_Q(\underline{\bW})_{\underline{j}}
\coloneqq \langle  \exp( \C^Q_{\underline{j}}\underline{\bW} )  ,  \one_L  \rangle \in \R$
and
$\alpha_K(\underline{\bW})_{\underline{j}}
\coloneqq \langle  \exp( \C^K_{\underline{j}}\underline{\bW} )  , \one_L  \rangle \in \R$.
\end{definition}

\begin{definition}[$f(\cdot)$]\label{def:f}

Let
$\alpha_Q(\underline{\bW})_{\underline{j}} , \alpha_K(\underline{\bW})_{\underline{j}} \in \R$ follow \cref{def:general_alpha}, and
$u_K(\underline{\bW})_{\underline{j}},u_Q(\underline{\bW})_{\underline{j}} \in \R^L$ follow \cref{def:general_u}.
For any $\underline{j} \in [L]$, we define two functions $f_Q(\underline{\bW} )_{\underline{j}} , f_K(\underline{\bW} )_{\underline{j}} : \R^{d^2} \rightarrow \R^L$ as
\begin{align*}
    f_Q(\underline{\bW})_{\underline{j}} \coloneqq \underbrace{ \alpha_Q(\underline{\bW})_{\underline{j}}^{-1} }_{ \mathrm{scalar} } \underbrace{ u_Q(\underline{\bW})_{\underline{j}} }_{ L \times 1 } ,
    \quad
    f_K(\underline{\bW})_{\underline{j}} \coloneqq \underbrace{ \alpha_K(\underline{\bW})_{\underline{j}}^{-1} }_{ \mathrm{scalar} } \underbrace{ u_K(\underline{\bW})_{\underline{j}} }_{ L \times 1 } ,
\end{align*}
such that $f_Q(\underline{\bW} ) ,f_K(\underline{\bW} ) \in \R^{L \times L}$ denote the matrices whose $\underline{j}$-th rows are $f_Q(\underline{\bW} )_{\underline{j}} ^\top,f_K(\underline{\bW} )_{\underline{j}} ^\top$.
\end{definition}

\begin{definition}[$c(\cdot)$]\label{def:c}
For every $\underline{j} \in [L]$, let $f_Q(\underline{\bW} )_{\underline{j}},f_K(\underline{\bW} )_{\underline{j}} : \R^{d^2} \rightarrow \R^L$ follow \cref{def:f}. 
For every $i \in [d]$,  let $ \bC^{(3)}[\cdot,i]\in \R^L$ follow \eqref{eqn:general_Cs}.
For each $\underline{j} \in [L]$ and $i \in [d]$,
we define two functions $c_Q(\underline{\bW})_{\underline{j},i} , c_K(\underline{\bW})_{\underline{j},i} : \R^{d^2} \times \R^{d^2} \rightarrow \R$ as
\begin{align*}
    c_Q(\underline{\bW})_{\underline{j},i}
    \coloneqq \langle f_Q(\underline{\bW})_{\underline{j}},  \bC^{(3)}[\cdot,i] \rangle - \bY_{\underline{j},i},
    \quad
    c_K(\underline{\bW})_{\underline{j},i}
    \coloneqq \langle f_K(\underline{\bW})_{\underline{j}},  \bC^{(3)}[\cdot,i]  \rangle - \bY_{\underline{j},i}.
\end{align*}

Here $\bY_{\underline{j},i}$ is the $(\underline{j},i)$-th coordinate/location of $\bY \in \R^{L \times d}$ for $\underline{j} \in [L], i \in [d]$.
\end{definition}
These give 
\begin{align*}
\underbrace{ c_Q(\underline{\bW}) }_{L \times d} 
= 
\underbrace{ f_Q(\underline{\bW}) }_{L \times L} 
\underbrace{ \bC^{(3)} }_{L \times d} 
- \underbrace{\bY}_{L \times d},
\quad \text{and} \quad
\underbrace{c_K(\underline{\bW}) }_{L \times d} 
= \underbrace{ f_K(\underline{\bW}) }_{L \times L} \underbrace{  \bC^{(3)} }_{L \times d} - \underbrace{\bY}_{L \times d}.    
\end{align*}

\begin{definition}\label{def:general_loss}
For every $\underline{j} \in [L]$ and every $i \in [d]$,  let 
$
 \mathcal{L}_Q(\underline{\bW})_{\underline{j},i} \coloneqq  c_Q(\underline{\bW})_{\underline{j},i}^2/2$, and $\mathcal{L}_K(\underline{\bW})_{\underline{j},i} \coloneqq  c_K(\underline{\bW})_{\underline{j},i}^2/2
 $. 
\end{definition}
Let matrix $\bW_Q=$ $\bW_Q^\star+\bB_Q \bA_Q \cdot \bW_K=\bW_K^\star+\bB_K \bA_K$ and loss function $\mathcal{L}$ be \eqref{eqn:general_generic_lora_loss_single}.
From above definitions, it holds
$\mathcal{L} (\bA_K,\bB_K,\bA_Q,\bB_Q)
= \calL(\underline{\bW}_Q,\underline{\bW}_K)$ and the adaptation gradients of  $\mathcal{L}$ \eqref{eqn:general_generic_lora_loss_single} become
\begin{align}
\label{eqn:general_element_form_Q}
\pdv{\mathcal{L} \left(\underline{\bW}_Q,\underline{\bW}_K\right)}{\underline{\bW}_Q}
= \pdv{}{\underline{\bW}_Q} \sum_{\underline{j}}^L \sum_{i=1}^d
\mathcal{L}_Q(\underline{\bW}_Q)_{\underline{j},i} 
& = ~
\pdv{}{\underline{\bW}_Q}\half\sum_{\underline{j}}^L\sum_{i=1}^d 
c_Q(\underline{\bW}_Q)_{\underline{j},i}^2,
\end{align}
and
\begin{align}
\label{eqn:general_element_form_K}
\pdv{\mathcal{L}\left(\underline{\bW}_Q , \underline{\bW}_K\right)}
{\underline{\bW}_K^{\top}}
= \pdv{}{\underline{\bW}_K^{\top}}
\sum_{\underline{j}}^L\sum_{i=1}^d
\mathcal{L}_K(\underline{\bW}_{K}^{\top})_{\underline{j},i} 
& = ~
\pdv{}{\underline{\bW}_K^{\top}}\half\sum_{\underline{j}}^L\sum_{i=1}^d 
c_K(\underline{\bW}_{K}^{\top})_{\underline{j},i}^2.
\end{align}

\eqref{eqn:general_element_form_Q} and
\eqref{eqn:general_element_form_K} present a decomposition of the gradients of LoRA loss $\mathcal{L}$ \eqref{eqn:general_generic_lora_loss_single} 
aspect to $\underline{\bW}_Q$ and 
$\underline{\bW}_K^{\top}$ 
into $L \cdot d$ terms, each simple enough for tracking gradient computation.

Now, we are ready to compute the gradients of the LoRA loss aspect to $\underline{\bW}_Q$ and 
$\underline{\bW}_K^{\top}$   
as follows.

\begin{lemma}
[Low-Rank Decomposition of LoRA Gradients]  
\label{lemma:general_gradient_compute}
Let $\C_K\coloneqq\bC_K^{(1)} \otimes \bC_K^{(2)}, \C_Q\coloneqq\bC_Q^{(1)} \otimes \bC_Q^{(2)}$. 
Let fine-tuning weights be $\bW_Q=$ $\bW_Q^\star+\bB_Q \bA_Q $ and $ \bW_K = \bW_K^\star + \bB_K \bA_K$, and the loss function $\mathcal{L}$ follow \cref{def:general_loss}.
It holds

{\footnotesize
\begin{align*}
\pdv{\mathcal{L}\left(\underline{\bW}_Q,\underline{\bW}_K\right)}{\underline{\bW}_Q}
= & ~\sum_{\underline{j}=1}^L \sum_{i=1}^d c_Q\left(\underline{\bW}_Q\right)_{\underline{j},i}\(\C^Q_{\underline{j}}\)^{\top}
\left(\diag\left(f_Q\left(\underline{\bW}_Q\right)_{\underline{j}}\right) -f_Q\left(\underline{\bW}_Q\right)_{\underline{j}} f_Q\left(\underline{\bW}_Q\right)_{\underline{j}}^{\top}\right)\bC^{(3)} [\cdot,i] ,\\
\pdv{\mathcal{L}\left(\underline{\bW}_Q,\underline{\bW}_K\right)}
{\underline{\bW}_K^{\top}}
= & ~\sum_{\underline{j}=1}^L \sum_{i=1}^d c_K\left(\underline{\bW}_K^{\top}\right)_{\underline{j},i}
\left(\C^K_{\underline{j}}\right)^{\top}
\left(\diag\left(f_K\left(\underline{\bW}_K^{\top}\right)_{\underline{j}}\right) -f_K\left(\underline{\bW}_K^{\top}\right)_{\underline{j}} f_K\left(\underline{\bW}_K^{\top}\right)_{\underline{j}}^{\top}\right)\bC^{(3)}[\cdot,i] .  
\end{align*}}
 
\end{lemma}

\begin{proof}
 This lemma is a generalization of  \cref{lemma:special1}.
\end{proof}

Next, we introduce the $q(\cdot)$ and $p(\cdot)$ notations. 
Again, there are two sets corresponding to the two sets of adaptors.
\begin{definition}
\label{def:general_q}
Let 
    $q_K(\underline{\bW})
    \coloneqq \bC^{(3)} \(c_K(\underline{\bW})\)^\sT \in \R^{L\times L}
    $, $q_Q(\underline{\bW})
    \coloneqq \bC^{(3)} \(c_Q(\underline{\bW})\)^\sT \in \R^{L\times L}
    $.
\end{definition}

\begin{definition}\label{def:general_p}
For every index $\underline{j}\in [L]$ , we define $p_Q(\underline{\bW})_{\underline{j}}, p_Q(\underline{\bW})_{\underline{j}} \in \R^L$ as
\begin{align*}
    p_Q(\underline{\bW})_{\underline{j}}
    & \coloneqq \left(\diag\left(f_Q\left(\underline{\bW}
    \right)_{\underline{j}}\right) -f_Q\left(\underline{\bW}\right)_{\underline{j}} f_Q\left(\underline{\bW}\right)_{\underline{j}}^{\top}\right) q_Q(\underline{\bW})_{\underline{j}} , \\
    p_K(\underline{\bW})_{\underline{j}}
    & \coloneqq \left(\diag\left(f_K\left(\underline{\bW}\right)_{\underline{j}}\right) -f_K\left(\underline{\bW}\right)_{\underline{j}} f_K\left(\underline{\bW}\right)_{\underline{j}}^{\top}\right) q_K(\underline{\bW})_{\underline{j}}.
\end{align*}
\end{definition}

\cref{lemma:general_gradient_compute} presents the Low-Rank Decomposition of LoRA Gradients. 
Before using the chain rule to compute the gradients of the loss $\mathcal{L}$ \eqref{eqn:general_generic_lora_loss_single} with respect to $\bA_Q, \bA_K, \bB_Q, \bB_K$, we need to define a matrix $\bT$ to handle the transpose term $\underline{\bW}_K^{\top}$.

\begin{lemma}[Sparse Matrix $\bT$]\label{lemma:T}
For any matrix $\bW \in \mathbb{R}^{m \times n}$, there exists a matrix $\bT(m, n) \in \mathbb{R}^{mn \times mn}$ such that $\underline{\bW}^{\top} = \bT(m, n) (\underline{\bW})$. 
The matrix $\bT(m, n)$ is sparse. 
Namely,
for any $i \in [mn]$, there exist $1 \leq p \leq m$ and $1 \leq k \leq n$ such that $i = (p-1)n + k$. Then, for any $i, j \in [mn]$,
\begin{align*}
    \bT(m, n)[i, j] := \begin{cases}
        1, & \text{if } j = (k-1)m + p,\\
        0, & \text{otherwise}.
    \end{cases}    
\end{align*}
\end{lemma}

\begin{proof}
For any $1 \leq p \leq m$ and $1 \leq k \leq n$, consider the position of $\bW[p, k]$ in $\underline{\bW}$ and $\underline{\bW}^{\top}$.

In $\underline{\bW}$, $\bW[p, k] = \underline{\bW}[(k-1)m + p]$.

In $\underline{\bW}^{\top}$, $\bW[p, k] = \underline{\bW}^{\top}[(p-1)n + k]$. 

Thus,
\begin{align*}
    \underline{\bW}^{\top}[i] 
    &= \bT(m, n)[i, \cdot] \underline{\bW} \\
    &= \bT(m, n)[i, j] \cdot \underline{\bW}[j].
\end{align*}
This completes the proof.
\end{proof}

Now, we are  ready to compute the gradients of the LoRA loss $\mathcal{L}$ \eqref{eqn:general_generic_lora_loss_single} with respect to $\bA_Q, \bA_K, \bB_Q, \bB_K$ using the chain rule as follows.

\begin{lemma}\label{lemma:gradients_general}
For any $a\in\R$, let $\diag_d(a)\in\R^{d\times d}$ be a $d\times d$ diagonal matrix with all entries equal to $a$. Recall $\bW_Q = \bW_Q^\star+\bB_Q \bA_Q $ and $ \bW_K = \bW_K^\star+\bB_K \bA_K$.
Let $ \bJ_{\bB_K}, \bJ_{\bA_K} \in \R^{d^2 \times rd}$ be two matrices such that $\underline{\bW}_Q = \underline{\bW}_Q^\star+\bJ_{\bB_Q} \underline{\bA}_Q$ and $ \underline{\bW}_Q = \underline{\bW}_Q^\star+\bJ_{\bA_Q} \underline{\bB}_Q$ via
\begin{align*}
\bJ_{\bB_K}
&=\begin{pmatrix}
\bB_K & & & \\
& \bB_K & & \\
& & \ddots & \\
& & & \bB_K
\end{pmatrix}, 
\bJ_{\bA_Q} = \begin{pmatrix}
\diag_d\left(\bA_K[1,1]\right) & \cdots & \diag_d\left(\bA_K[r, 1]\right) \\
\diag_d\left(\bA_K[1,2]\right) & \cdots & \diag_d\left(\bA_K[r, 2]\right) \\
\vdots & & \vdots 
\\
\diag_d\left(\bA_K[1, d]\right) & \cdots & \diag_d\left(\bA_K[r, d]\right)
\end{pmatrix}. 
\end{align*}
Let $\bJ_{\bB_K}, \bJ_{\bA_K}$ be two matrices such that $\underline{\bW}_K = \underline{\bW_K^\star} + \bJ_{\bB_K} \underline{\bA}_K$ and $ \underline{\bW}_K = \bW_K^\star + \bJ_{\bA_K} \underline{\bB}_K$ via
\begin{align*}
\bJ_{\bB_Q}
&=
\begin{pmatrix}
\bB_Q & & & \\
& \bB_Q & & \\
& & \ddots & \\
& & & \bB_Q
\end{pmatrix}, 
\bJ_{\bA_Q}
=\begin{pmatrix}
\diag_d\left(\bA_Q[1,1]\right) & \cdots & \diag_d\left(\bA_Q[r, 1]\right) \\
\diag_d\left(\bA_Q[1,2]\right) & \cdots & \diag_d\left(\bA_Q[r, 2]\right) \\
\vdots & & \vdots \\
\diag_d\left(\bA_Q[1, d]\right) & \cdots & \diag_d\left(\bA_Q[r, d]\right)
\end{pmatrix}.
\end{align*}
Then the derivatives of loss function $\mathcal{L}$ \eqref{eqn:general_generic_lora_loss_single} respect to $\underline{\bA}_Q,\underline{\bB}_Q,\underline{\bA}_K,\underline{\bB}_K$ are
{\footnotesize
\begin{align*}
& \pdv{\mathcal{L}} {\underline{A}_Q}
= \sum_{\underline{j}=1}^L \sum_{i=1}^d \left(J_{B_Q}\right)^{\top} c_Q
\left(\underline{W}_Q\right)_
{\underline{j},i} \left(\C^Q_{\underline{j}}\right)^{\top} \left(\diag\left(f_Q\left(\underline{W}_Q\right)_{\underline{j}}\right) - f_Q\left(\underline{W}_Q\right)_{\underline{j}} f_Q\left(\underline{W}_Q\right)_{\underline{j}}^{\top}\right) C^{(3)}[\cdot, i] , \\
& \pdv{\mathcal{L}} {\underline{B}_Q}
= \sum_{\underline{j}=1}^L \sum_{i=1}^d \left(J_{A_Q}\right)^{\top} c_Q\left(\underline{W}_Q\right)_{\underline{j},i} \left(\C^Q_{\underline{j}}\right)^{\top} \left(\diag\left(f_Q\left(\underline{W}_Q\right)_{\underline{j}}\right) - 
f_Q\left(\underline{W}_Q\right)_{\underline{j}} f_Q\left(\underline{W}_Q\right)_{\underline{j}}^{\top}\right) C^{(3)}[\cdot, i] ,
\\
& \pdv{\mathcal{L}} {\underline{A}_K}
= \sum_{\underline{j}=1}^L \sum_{i=1}^d \left(T\left(d^2, d^2\right) J_{B_K}\right)^{\top} 
c_K\left(\underline{W}_K^{\top}\right)_{\underline{j}, i} \left(\C^K_{\underline{j}}\right)^{\top} \left(\diag\left(f_K\left(\underline{\bW}_K^{\top}\right)_{\underline{j}}\right) - f_K
\left(\underline{W}_K^{\top}\right)_{\underline{j}} f_K
\left(\underline{W}_K^{\top}\right)_{\underline{j}}^{\top}\right) C^{(3)}[\cdot, i] ,
\\
& \pdv{\mathcal{L}} {\underline{B}_K}
= \sum_{\underline{j}=1}^L \sum_{i=1}^d \left(T\left(d^2, d^2\right) J_{A_K}\right)^{\top} c_K\left(\underline{W}_K^{\top}\right)_{\underline{j},i}
\left(\C^K_{\underline{j}}\right)^{\top} \left(\diag\left(f_K\left(\underline{W}_K^{\top}\right)_{\underline{j}}\right) - f_K\left(\underline{W}_K^{\top}\right)_{\underline{j}} f_K\left(\underline{W}_K^{\top}\right)_{\underline{j}}^{\top}\right) C^{(3)}[\cdot, i] .
\end{align*}
}
\end{lemma}

\begin{proof}
$\pdv{\mathcal{L}}{\underline{\bA}_Q}$ and $\pdv{\mathcal{L}}{\underline{\bB}_Q}$  follow  \cref{lemma:speical2} directly. 

For $\pdv{\mathcal{L}}{\underline{\bA}_K}$ and $\pdv{\mathcal{L}}{\underline{\bB}_K}$, we have:
\begin{align*}
\underline{\bW}_K^{\top} 
&= \bT(d^2, d^2) \underline{\bW}_K \\
&= \bT(d^2, d^2) \left(\underline{\bW}_K^\star + \bJ_{\bB_K} \underline{\bA}_K\right) \\
&= \bT(d^2, d^2) \left(\underline{\bW}_K^\star + \bJ_{\bA_K} \underline{\bB}_K\right).
\end{align*}

Therefore,
\begin{align*}
\pdv{\mathcal{L}}{\underline{\bA}_K}
&= \pdv{\underline{\bW}_K^{\top}}{\underline{\bA}_K} \pdv{\mathcal{L}(\underline{\bW}_Q, \underline{\bW}_K)}{\underline{\bW}_K^{\top}} \\
&= \bT(d^2, d^2) \bJ_{\bB_K} \pdv{\mathcal{L}(\underline{\bW}_Q, \underline{\bW}_K)}{\underline{\bW}_K^{\top}}.
\end{align*}

Similarly,
\begin{align*}
\pdv{\mathcal{L}}{\underline{\bB}_K}
&= \pdv{\underline{\bW}_K^{\top}}{\underline{\bB}_K} \pdv{\mathcal{L}(\underline{\bW}_Q, \underline{\bW}_K)}{\underline{\bW}_K^{\top}} \\
&= \bT(d^2, d^2) \bJ_{\bA_K} \pdv{\mathcal{L}(\underline{\bW}_Q, \underline{\bW}_K)}{\underline{\bW}_K^{\top}}.
\end{align*}
Thus, we complete the proof by following the conclusions of \cref{lemma:general_gradient_compute}.
\end{proof}

Next, we simplify the derivatives with $p(\cdot)$ notation.
\begin{lemma}
\label{lemma:general_LoRA_gradients}
Let $q_Q, q_K \in \R^{L\times L}$ as defined in \cref{def:general_q}. 
Let $p_Q, p_K$ as defined in \cref{def:general_p}. 
Then it holds
\begin{align*}
        \pdv{\calL}{\underline{\bA}_Q}
        = & ~ \vect\left(\bB_Q^{\top}
        \left(\bC_{Q}^{(1)}\right)^{\top} p_Q(\underline{\bW}_Q) \bC_{Q}^{(2)}\right), \\
        \frac{\partial \mathcal{L}}{\partial \underline{\bB}_Q}
        = & ~ \vect\left(
        \left(\bC_{Q}^{(1)}\right)^{\top} p_Q(\underline{\bW}_Q) \bA_Q \bC_{Q}^{(2)}\right),\\
        \pdv{\mathcal{L}}{\underline{\bA
        }_K}  
        = & ~
        \bT\left(d^2, d^2\right)^{\top} \vect\left(\bB_K^{\top}\left(\bC_K^{(1)}\right)^{\top} p_K\left(\underline{\bW}_K^{\top}\right) \bC_K^{(2)}\right), \\
        \pdv{\mathcal{L}}{\underline{\bB
        }_K} 
        = & ~
        \bT\left(d^2, d^2\right)^{\top} \vect\left(\left(\bC_K^{(1)}\right)^{\top} p_K\left(\underline{\bW}_K^{\top}\right)\bA_K \bC_K^{(2)}\right).
\end{align*}
\end{lemma}

\begin{proof}
    For $\pdv{\mathcal{L}} {\underline{\bA}_Q}$ and
    $ \pdv{\mathcal{L}} {\underline{\bB}_Q}$,
    we follow the proof of \cref{thm:main_special}.

    For $\pdv{\mathcal{L}} {\underline{\bA}_K}$, 
    we have
    \begin{align*}
    &~ \pdv{\mathcal{L}} {\underline{\bA}_K}
    \\
    = & ~\sum_{\underline{j}=1}^L \sum_{i=1}^d
    \left(\bT\left(d^2, d^2\right) \bJ_{\bB_K}\right)^{\top} c_K\left(\underline{\bW}_K^{\top}\right)_{\underline{j},{i}}\(\C^K_{\underline{j}}\)^{\top}
    \left(\diag\left(f_K\left(\underline{\bW}_K^{\top}\right)_{\underline{j}}\right)-f_K\left(\underline{\bW}_K^{\top}\right)_{\underline{j}} f_K\left(\underline{\bW}_K^{\top}\right)_{\underline{j}}^{\top}\right) \bC^{(3)} 
    [\cdot, i] 
    \annot{By \cref{lemma:gradients_general}}
    \\
    = & ~ \sum_{\underline{j}=1}^L\left(\bT\left(d^2, d^2\right) \bJ_{\bB_K}\right)^{\top}\(\C^K_{\underline{j}}\)^{\top}
    \left(\diag\left(f_K\left(\underline{\bW}_K^{\top}\right)_{\underline{j}}\right)-f_K\left(\underline{\bW}_K^{\top}\right)_{\underline{j}} f_K\left(\underline{\bW}_K^{\top}\right)_{\underline{j}}^{\top}\right) q_K\left(\underline{\bW}_K^{\top}\right)_{\underline{j}}
    \annot{By \cref{def:general_q}}
    \\
    =& ~\bT\left(d^2, d^2\right)^{\top} \sum_{\underline{j}=1}^L \bJ_{\bB_K}^{\top}\(\C^K_{\underline{j}}\)^{\top} p_K\left(\underline{\bW}_K^{\top}\right)_{\underline{j}} 
    \annot{By \cref{def:general_p}}
    \\
    =& ~\bT\left(d^2, d^2\right)^{\top} \vect\left(\bB_K^{\top}\left(\bC_K^{(1)}\right)^{\top} p_K\left(\underline{\bW}_K^{\top}\right) \bC_K^{(2)}\right).
    \annot{By \cref{lemma:tensor_trick}}
    \end{align*}

Similarly, for $\pdv{\mathcal{L}} {\underline{\bB
}_K}$, it holds
\begin{align*}
    & ~ \pdv{\mathcal{L}}{\underline{\bB
    }_K}\\
    = & ~ \sum_{\underline{j}=1}^{L}\sum_{i=1}^{d}\left(\bT(d^2,d^2)\bJ_{\bA_K}\right)^{\top}
    c_K\left(\underline{\bW}_K^{\top}\right)_{\underline{j},{i}}\(\C^K_{\underline{j}}\)^{\top}
    \left(\diag\left(f_K\left(\underline{\bW}_K^{\top}\right)_{\underline{j}}\right)-f_K\left(\underline{\bW}_K^{\top}\right)_{\underline{j}} f_K\left(\underline{\bW}_K^{\top}\right)_{\underline{j}}^{\top}\right) \bC^{(3)}[\cdot, i]\\
    = & ~ \sum_{\underline{j}=1}^L\left(\bT\left(d^2, d^2\right) \bJ_{\bA_K}\right)^{\top}
    \(\C^K_{\underline{j}}\)^{\top}
    \left(\diag\left(f_K\left(\underline{\bW}_K^{\top}\right)_{\underline{j}}\right)-f_K\left(\underline{\bW}_K^{\top}\right)_{\underline{j}} f_K\left(\underline{\bW}_K^{\top}\right)_{\underline{j}}^{\top}\right) q_K\left(\underline{\bW}_K^{\top}\right)_{\underline{j}}
\\
    = & ~\bT\left(d^2, d^2\right)^{\top} \sum_{\underline{j}=1}^L \bJ_{\bA_K}^{\top}\(\C^K_{\underline{j}}\)^{\top} q_K\left(\underline{\bW}_K^{\top}\right)_{\underline{j}} 
\\
    = & ~\bT\left(d^2, d^2\right)^{\top} \vect\left(\left(\bC_K^{(1)}\right)^{\top} p_K\left(\underline{\bW}_K^{\top}\right)\bA_K \bC_K^{(2)}\right).
\end{align*} 
This completes the proof.
\end{proof}

Similarly, \cref{lemma:general_LoRA_gradients} states that the chain rule terms for characterizing \cref{def:ALoRAGC_general} are tied to $p_Q(\cdot)$ and $p_KQ(\cdot)$.
Therefore, to characterize $\Tilde{\bG}^{(A)}_Q$, $\Tilde{\bG}^{(B)}_Q$, $\Tilde{\bG}^{(A)}_K$, and $\Tilde{\bG}^{(B)}_K$ (i.e., the approximations of $\bG^{(A)}_Q$, $\bG^{(B)}_Q$, $\bG^{(A)}_K$, and $\bG^{(B)}_K$), for $\mu = Q, K$, we need to approximate the functions $f_{\mu}(\cdot)$, $q_{\mu}(\cdot)$, $c_{\mu}(\cdot)$, and thus $p_{\mu}(\cdot)$ with precision guarantees. 
To do so, it is convenient to consider the following decomposition of $p_{\mu}(\cdot)$ for $\mu=Q,K$.

\begin{definition}
\label{def:p1p2_general}
For every index  $\underline{j}\in [L]$, we define $p^K_{1}(\underline{\bW})_{\underline{j}},p^K_{2}(\underline{\bW})_{\underline{j}}\in \R^L$ as
\begin{align*}
p^Q_{1}(\underline{\bW})_{\underline{j}} & \coloneqq \diag\left(f_Q\left(\underline{\bW}\right)_
{\underline{j}}\right) q_Q(\underline{\bW})_{\underline{j}},
\quad
p^Q_{2}(\underline{\bW})_{\underline{j}}
\coloneqq f_Q\left(\underline{\bW}\right)_{\underline{j}} f_Q\left(\underline{\bW}\right)_{\underline{j}}^{\top} q_Q(\underline{\bW})_{\underline{j}},
\\
p^K_{1}(\underline{\bW})_{\underline{j}} & \coloneqq \diag\left(f_K\left(\underline{\bW}\right)_
{\underline{j}}\right) q_K(\underline{\bW})_{\underline{j}},
\quad
p^K_{2}(\underline{\bW})_{\underline{j}} \coloneqq f_K\left(\underline{\bW}\right)_{\underline{j}} f_K\left(\underline{\bW}\right)_{\underline{j}}^{\top} q_K(\underline{\bW})_{\underline{j}}.
\end{align*}

such that 
$ p_Q(\underline{\bW})
= p_1^{Q}(\underline{\bW}) - p_2^{Q}(\underline{\bW}) , p_Q(\underline{\bW})
= p_1^{Q}(\underline{\bW}) - p_2^{Q}(\underline{\bW}) $.

\end{definition}

\paragraph{Overview of Our Proof Strategy.}
Similar to \cref{sec:special}, we adopt the following strategy: term-by-term approximation for precision-guaranteed, almost linear time algorithms to compute 
LoRA gradients in \cref{def:ALoRAGC_general}. 
For all $\mu = Q, K$, we do the following.

\begin{enumerate}[leftmargin=3.5em]
    \item [\textbf{Step 1.}] Prove the existence of almost linear approximation algorithms for $f_{\mu}(\cdot)$, $q_{\mu}(\cdot)$, and $c_{\mu}(\cdot)$ via low-rank approximation (\cref{lemma:approx_general_f}, \cref{lemma:approx_general_q}, and \cref{lemma:approx_general_c}).
    
    \item [\textbf{Step 2.}] Prove the existence of almost linear approximation algorithms for $p_1^{\mu}(\cdot)$, $p_2^{\mu}(\cdot)$, and thus $p_{\mu}(\cdot)$ via the low-rank-preserving property of the multiplication between $f_{\mu}(\cdot)$ and $q_{\mu}(\cdot)$ (\cref{lemma:approx_general_p1} and \cref{lemma:approx_general_p2}).
    
    \item [\textbf{Step 3.}] Prove the existence of almost linear approximation algorithms for the LoRA adapter gradients (i.e., $\pdv{\mathcal{L}}{\underline{\bA}_Q}$, $\pdv{\mathcal{L}}{\underline{\bA}_K}$, $\pdv{\mathcal{L}}{\underline{\bB}_Q}$, and $\pdv{\mathcal{L}}{\underline{\bB}_K}$ in \cref{lemma:general_LoRA_gradients}) using the results from \textbf{Step 1} and \textbf{Step 2} (\cref{thm:main_general}).
\end{enumerate}

\textbf{Step 1.} We start with low-rank approximations for $f_{\mu}(\cdot),
q_{\mu}(\cdot),c_{\mu}(\cdot)$.

\begin{lemma}[Approximate $f_Q(\cdot), f_K(\cdot)$]
\label{lemma:approx_general_f}
    Let $\Gamma = o(\sqrt{\log L}) $, for $\mu = Q,K $, suppose $\bC_{\mu}^{(1)}, 
    \bC_{\mu}^{(2)} \in \mathbb{R}^{L \times d}$, $ \bW \in \mathbb{R}^{d \times d}$, and
    $f_{\mu}(\underline{\bW}) = \bD^{-1} \exp \left(\bC_{\mu}^{(1)} \bW \left(\bC_{\mu}^{(2)}\right)^{\top}\right)$ with $\bD$ following
    \eqref{eqn:general_generic_lora_loss_single}. 
    There exists a $k_1 = L^{o(1)}$ such that if $\norm{\bC_{\mu}^{(1)} \bW}_{\infty} \leq \Gamma$ and $\norm{\bC_{\mu}^{(2)}}_{\infty} \leq \Gamma$, then there exist four matrices $\bU_1^{Q}, \bV_1^{Q} , \bU_1^{K}, \bV_1^{K}\in \mathbb{R}^{L \times k_1}$ such that 
    \begin{align*}
     \left\|\bU_1^{Q} (\bV_1^{Q})^{\top}-f_Q(\underline{\bW})\right\|_{\infty} 
     \leq & ~\epsilon /\poly (L),\\
     \left\|\bU_1^{K} (\bV_1^{K})^{\top}-f_K(\underline{\bW})\right\|_{\infty} 
     \leq & ~\epsilon /\poly (L).  
    \end{align*}  
    In addition, it takes $L^{1+o(1)}$ time to construct $\bU_1^{Q}, \bV_1^{Q} , \bU_1^{K}, \bV_1^{K}$.
\end{lemma}

\begin{proof}
This follows the proof of \cref{lemma:approx_f}
\end{proof}

\begin{lemma}[Approximate $c_Q(\cdot),c_K(\cdot)$]
\label{lemma:approx_general_c}
    Assume all numerical values are in $O(\log L)$ bits. 
    Let $d=O(\log L)$ and $c_Q(\underline{\bW}),c_K(\underline{\bW})\in\R^{L\times d}$ follows \cref{def:c}.
    Then there exist four matrices $\bU_1^{Q}, \bV_1^{Q}, \bU_1^{K}, \bV_1^{K} \in \mathbb{R}^{L \times k_1}$ such that 
    \begin{align*}
     \left\|\bU_1^{Q} (\bV_1^{Q})^{\top} \bC^{(3)}-\bY-c_Q(\underline{\bW})\right\|_{\infty} 
     \leq & ~\epsilon / \poly(L),
     \\
     \left\|\bU_1^{K} (\bV_1^{K})^{\top} \bC^{(3)}-\bY-c_K(\underline{\bW})\right\|_{\infty} 
     \leq & ~\epsilon / \poly(L).   
    \end{align*}
    
\end{lemma}

\begin{proof}
    This follows the proof of \cref{lemma:approx_c}
\end{proof}

\begin{lemma}[Approximate $q_Q(\cdot),q_K(\cdot)$]
\label{lemma:approx_general_q}
    Let $k_2 = L^{o(1)}$, $c_Q(\bW) , c_K(\bW) \in \R^{L\times d}$ follows \cref{def:c} and 
    let $q_K(\underline{\bW})
    \coloneqq \bC^{(3)} \(c_K(\underline{\bW})\)^\sT \in \R^{L\times L}$ , 
    $q_Q(\underline{\bW}) \coloneqq \bC^{(3)} \(c_Q(\underline{\bW})\)^\sT \in \R^{L\times L}$. (follows \cref{def:general_q}).
    Then there exist four matrices $\bU_2^{Q}, \bV_2^{Q} , \bU_2^{K}, \bV_2^{K}\in \mathbb{R}^{L \times k_2}$ such that 
    \begin{align*}
     \left\|\bU_2^{Q} (\bV_2^{Q})^{\top}-q_Q(\underline{\bW})\right\|_{\infty} 
     \leq & ~\epsilon /\poly (L),\\
     \left\|\bU_2^{K} (\bV_2^{K})^{\top}-q_K(\underline{\bW})\right\|_{\infty} 
     \leq & ~\epsilon /\poly (L).  
    \end{align*}
    In addition, it takes $L^{1+o(1)}$ time to construct $\bU_2^{Q}, \bV_2^{Q} , \bU_2^{K}, \bV_2^{K}$.
    
\end{lemma}

\begin{proof}
    This follows the proof of \cref{lemma:approx_q}
\end{proof}

\textbf{Step 2.} Now, we use above lemmas to construct low-rank approximations for $p_1^{\mu}(\cdot) , p_2^{\mu}(\cdot), p_{\mu}(\cdot)$.

\begin{lemma}[Approximate $p_1^{Q}(\cdot) , p_1^{K}(\cdot)$]
\label{lemma:approx_general_p1}
    Let $k_1,k_2,k_3=L^{o(1)}$. For $\mu=K,Q$,  
    suppose $\bU_1^{\mu}, \bV_1^{\mu} \in \mathbb{R}^{L \times k_1}$ approximate $f_{{\mu}}(\underline{\bW}) \in \R^{L\times L}$ such that 
    $\left\|\bU_1^{\mu} (\bV_1^{\mu})^{\top} -f_{{\mu}}(\underline{\bW})\right\|_{\infty} \leq \epsilon / \poly(L)$, and
    $\bU_2^{\mu}, \bV_2^{\mu} \in \mathbb{R}^{L \times k_2}$ approximate the $q_{{\mu}}(\underline{\bW}) \in \mathbb{R}^{L \times L}$ such that
    $\left\|\bU_2^{\mu} (\bV_2^{\mu})^{\top} -q_{{\mu}}(\underline{\bW})\right\|_{\infty} \leq \epsilon / \poly(L) $.
    Then there exist two matrices $\bU_3^{\mu}, \bV_3^{\mu} \in \mathbb{R}^{L \times k_3}$ such that
    \begin{align*}
     \left\|\bU_3^{\mu} (\bV_3^{\mu})^{\top} -p_1^{\mu}(\underline{\bW})\right\|_{\infty} \leq \epsilon / \poly(L),
     \quad
     \text{for } \mu=K,Q.
    \end{align*} 
     In addition, it takes $L^{1+o(1)}$ time to construct $\bU_3^{Q}, \bV_3^{Q} , \bU_3^{K}, \bV_3^{K}$.
\end{lemma}
\begin{proof}
    This follows the proof of \cref{lemma:approx_p1}
\end{proof}

\begin{lemma}[Approximate $p_2^{Q}(\cdot),p_2^{K}(\cdot)$]
\label{lemma:approx_general_p2}
    Let $k_1,k_2,k_4=L^{o(1)}$. 
    Let $p_2^{Q}(\underline{\bW}) , p_2^{K}(\underline{\bW})\in\R^{L\times L}$  such that its $\underline{j}$-th column is $p_2(\underline{\bW})_{\underline{j}} = f(\underline{\bW})_{\underline{j}}
    f(\underline{\bW})_{\underline{j}}^{\top} q(\underline{\bW})_{\underline{j}}$ follow \cref{def:p1p2_general}, for each $\underline{j} \in [L]$. For $\mu=K,Q$,
    suppose $\bU_1^{\mu}
    , \bV_1^{\mu} \in \mathbb{R}^{L \times k_1}$ approximates the ${f_{\mu}}(\underline{\bW})$ such that $\left\|\bU_1^{\mu} (\bV_1^{\mu})^{\top} - f_{\mu}(\underline{\bW})\right\|_{\infty} 
    \leq \epsilon /\poly (L)$, and 
     $\bU_2^{\mu}, \bV_2^{\mu} \in \mathbb{R}^{L \times k_2}$ approximates the $q_{\mu}(\underline{\bW}) \in \mathbb{R}^{L \times L}$ such that $\left\|\bU_2^{\mu} (\bV_2^{\mu})^{\top} - q_{\mu}(\underline{\bW})\right\|_{\infty} 
     \leq \epsilon /\poly (L)$. 
     Then there exist matrices $\bU_4^{\mu}, \bV_4^{\mu} \in \mathbb{R}^{L \times k_4}$ such that 
     \begin{align*}
      \left\|\bU_4^{\mu} (\bV_4^{\mu})^{\top} -p_2^{\mu}(\underline{\bW})\right\|_{\infty} 
      \leq \epsilon /\poly(L),
      \quad
     \text{for } \mu=K,Q.
     \end{align*} 
     In addition, it takes $L^{1+o(1)}$ time to construct $\bU_4^{Q}, \bV_4^{Q} , \bU_4^{K}, \bV_4^{K}$.
\end{lemma}
\begin{proof}
    This follows the proof of \cref{lemma:approx_p2}
\end{proof}

\textbf{Step 3.}
Combining above, we arrive our main result:
almost linear algorithm for \cref{def:ALoRAGC_general}.

\begin{theorem}[Main Result: Existence of almost Linear Time $\mathsf{ALoRAGC}$]
Let $\Gamma=o(\sqrt{\log L})$ .
Suppose all numerical values are in $\mathrm{O}(\log L)$-bits encoding. Then there exists a $L^{1+o(1)}$ time algorithm to solve $\operatorname{ALoRAGC}\left(L, d=O(\log L), r=L^{o(1)}, \epsilon = 1/ \poly(L) \right.$ (i.e \cref{def:ALoRAGC_general}) up to $1 /\poly(L)$  accuracy.
In particular, this algorithm outputs  gradient matrices $\{\Tilde{\bG}^{(A)}_{\mu} \in \R^{d\times r},\Tilde{\bG}^{(B)}_{\mu} \in \R^{r\times d}\}_{\mu=K,Q}$ such that 
\begin{align*}
\max\(
\left\|\pdv{\calL}{\underline{\bB}_{\mu}}-\Tilde{\underline{\bG}}^{(B)}_{\mu}\right\|_{\infty},\left\|\pdv{\calL} {\underline{\bA}_{\mu}}-\Tilde{\underline{\bG}}^{(A)}_{\mu}\right\|_
{\infty}\)
\leq & ~ 1 / \poly(L),
    \quad
     \text{for } \mu=K,Q.
\end{align*}
\end{theorem}

\begin{proof}[Proof of \cref{thm:main_general}]
 By the definitions of  matrices $
 p_1^{K}(\underline{\bW}), p_1^{Q}(\underline{\bW}) , p_2^{K}(\underline{\bW}), p_2^{Q}(\underline{\bW}) $ in \cref{def:p1p2_general} and $
 p_{K}(\underline{\bW}), p_{Q}(\underline{\bW})$ in 
 \cref{def:general_p}. 
 It is straightforward that
\begin{align*}
p_K(\underline{\bW})  =p^K_{1}(\underline{\bW})-p^K_{2}(\underline{\bW}),
\quad\text{and}\quad
p_Q(\underline{\bW})  =p^Q_{1}(\underline{\bW})-p^Q_{2}(\underline{\bW}).
\end{align*}

 According to  \cref{lemma:general_LoRA_gradients}, we have 
\begin{align*}
    \frac{\partial \mathcal{L}}{{\partial \underline{\bA}_Q}}
    =&~\vect\left(\bB_Q^{\top}
    \left(\bC_Q^{(1)}\right)^{\top} p_Q\left(\underline{\bW}_Q\right) \bC_Q^{(2)}\right) \\
    \frac{\partial \mathcal{L}}{\partial \underline{\bB}_Q}
    =&~\vect\left(\left(\bC_Q^{(1)}\right)^{\top} p_Q\left(\underline{\bW}_Q\right) \bA_Q \bC_Q^{(2)}\right) \\    
   \pdv{\mathcal{L}}{\underline{\bA
    }_K}  
    = & ~
    \bT\left(d^2, d^2\right)^{\top} \vect\left(\bB_K^{\top}\left(\bC_K^{(1)}\right)^{\top} p_K\left(\underline{\bW}_K^{\top}\right) \bC_K^{(2)}\right) \\
    \pdv{\mathcal{L}}{\underline{\bB
    }_K} 
    = & ~
    \bT\left(d^2, d^2\right)^{\top} \vect\left(\left(\bC_K^{(1)}\right)^{\top} p_K\left(\underline{\bW}_K^{\top}\right)\bA_K \bC_K^{(2)}\right).
\end{align*}

    Next, we compute the time complexity of approximating these gradients to $1/\poly(L)$ precision. 
    
    For $\frac{\partial \mathcal{L}}{\partial \underline{A}_Q}$ and $\frac{\partial \mathcal{L}}{\partial \underline{B}_Q}$, we follow the proof of \cref{thm:main_special}. 
    Specifically, it takes $L^{1+o(1)}$ time to approximate these gradients to $1/\text{poly}(L)$ precision.

    For $\frac{\partial \mathcal{L}}{\partial \underline{A}_K}$ and $\frac{\partial \mathcal{L}}{\partial \underline{B}_K}$, we first note that $\left(T\left(d^2, d^2\right)\right)^{\top}$ is a constant matrix. In addition, due to \cref{thm:main_special}, $\vect\left(B_K^{\top} \left(C_K^{(1)}\right)^{\top} p_K\left(\underline{W}_K^{\top}\right) C_K^{(2)}\right)$ and $\vect\left(\left(C_K^{(1)}\right)^{\top} p_K\left(\underline{W}_K^{\top}\right) A_K C_K^{(2)}\right)$, which are similar to $\frac{\partial \mathcal{L}}{\partial \underline{A}_Q}$ and $\frac{\partial \mathcal{L}}{\partial \underline{B}_Q}$, take $L^{1+o(1)}$ time to approximate to $1/\text{poly}(L)$ precision.

    Therefore, to show the existence of $L^{1+o(1)}$ algorithms for \cref{def:ALoRAGC_general},  
    we prove exact computation for 
    $\bT\left(d^2, d^2\right)^{\top}
    \operatorname  {vec} \left(\bB_K^{\top}\left(\bC_K^{(1)}\right)
    ^{\top} p_K\left(\underline{\bW}_K^{\top}\right) \bC_K^{(2)}\right)$ 
    and 
    $\bT\left(d^2, d^2\right)^{\top}
    \operatorname {vec} \left(\left(\bC_K^{(1)}\right)^{\top} p_K\left(\underline{\bW}_K^{\top}\right)
    \bA_K \bC_K^{(2)}\right)$ 
    takes $o(L^{1+o(1)})$ time as follows.
    
    \textbf{Exact Computation for $\bT\left(d^2,d^2\right)^{\top} \vect\left(\bB_K^{\top}\left(\bC_K^{(1)}\right)^{\top} p_K\left(\underline{\bW}_K^{\top}\right) \bC_K^{(2)}\right)$.}
    Recall from \cref{lemma:T} that $T\left(d^2, d^2\right)^{\top}$ is a sparse matrix with only one non-zero entry in each row. 
    Thus, for each row, the exact computation takes $O(1)$ time. 
    Therefore, the total time is $O(d^2)$. 
    Given that $d = o(\log L)$, the overall time is still $L^{1+o(1)}$.

    \textbf{Exact Computation for $\bT\left(d^2, d^2\right)^{\top} \operatorname {vec} \left(\left(\bC_K^{(1)}\right)^{\top} p_K\left(\underline{\bW}_K^{\top}\right)\bA_K \bC_K^{(2)}\right)$.}
    Similarly, computing $T\left(d^2, d^2\right)^{\top} \operatorname{vec} \left(\left(C_K^{(1)}\right)^{\top} p_K\left(\underline{W}_K^{\top}\right) A_K C_K^{(2)}\right)$ takes $O(d^2)$ time. 
    Therefore, the total time is $O(d^2)$. 
    Given that $d = o(\log L)$, the overall time is still $L^{1+o(1)}$.

\paragraph{Approximation Error.}
For $\frac{\partial \mathcal{L}}{\partial \underline{\bA}_Q}$ and $\frac{\partial \mathcal{L}}{\partial \underline{\bB}_Q}$, we follow the proof of \cref{thm:main_special}.
For $\pdv{\mathcal{L}} {\underline{\bA}_K}$,
\begin{align*}
& ~ \left\|\pdv{\mathcal{L}} {\underline{\bA}_K} - \tilde{\bG}^{(A)}_K\right\|_{\infty}
\\
= & ~ \left\|\bT\left(d^2, d^2\right)^{\top} \vect
\left(\bB_K^{\top}\left(\bC_K^{(1)}\right)^{\top} p_K\left(\underline{\bW}_K^{\top}\right) \bC_K^{(2)}\right) - \bT\left(d^2, d^2\right)^{\top} \vect
\left(\bB_K^{\top}\left(\bC_K^{(1)}\right)^{\top}\tilde{p}_K \left(\underline{\bW}_K^{\top}\right) \bC_K^{(2)}\right)\right\|_{\infty}
\\
\leq & ~\left\|\bT\left(d^2, d^2\right)^{\top}\right\|_{\infty}
\left\|\left(\bB_K^{\top}\left(\bC_K^{(1)}\right)^{\top} p_K \left(\underline{\bW}_K^{\top}\right) \bC_K^{(2)}\right) -\left(\bB_K^{\top}\left(\bC_K^{(1)}\right)^{\top}\tilde{p}_K
\left(\underline{\bW}_K^{\top}\right) \bC_K^{(2)}\right)\right\|_{\infty} 
\\
\leq & ~ \left\|\left(\bB_K^{\top} \left(\bC_K^{(1)}\right)^{\top} \left(p^K_{1}\left(\underline{\bW}_K^{\top} \right) - 
\tilde{p}_{1}^{K}\left(\underline{\bW}_K^{\top}\right)\right) \bC_K^{(2)}\right)\right\|_{\infty}+\left\|\left(\bB_K^{\top}
\left(\bC_K^{(1)}\right)^{\top}\left(p^K_{2}
\left(\underline{\bW}_K^{\top}\right)-\tilde{p}_{2}^{K}\left(\underline{\bW}_K^{\top}\right)\right) \bC_K^{(2)}\right)\right\|_{\infty}
\\
\leq & ~ \left\|\bB_K\right\|_{\infty}
\left\|\bC_K^{(1)}\right\|_{\infty}
\left\|\bC_K^{(2)}\right\|_{\infty}\left(\left\|\left(p^K_{1}\left(\underline{\bW}_K^{\top}\right) -\tilde{p}_{1}^{K}\left(\underline{\bW}_K^{\top}\right)\right)\right\|_{\infty} + \left\|\left(p^K_{2}\left(\underline{\bW}_K^{\top}\right) -\tilde{p}_{2}^{K}\left(\underline{\bW}_K^{\top}\right)\right)\right\|_{\infty}\right)
\\
\leq & ~ \epsilon / \poly(L),
\end{align*}
where the first step follows from \cref{lemma:gradients_general}, 
the second step follows from the definition $\norm{A}_{\infty} \coloneqq \max_{i,j} \abs{A_{ij}}$ for any matrix $A$,
the third step follows from \cref{def:p1p2_general} and the triangle inequality,
the fourth step follows from the sub-multiplicative property of the $\infty$-norm,
and the last step follows from \cref{lemma:approx_general_p1} and \cref{lemma:approx_general_p2}.

Similarly, for $\pdv{\mathcal{L}}{\underline{\bB}_K}$, it holds
\begin{align*}
& ~ \left\|\frac{\partial \mathcal{L}}{\partial \underline{\bB}_K}-\tilde{\bG}^{(B)}_K\right\|_{\infty}
\\ 
= & ~ \left\|\bT\left(d^2, d^2\right)^{\top} \vect
\left(\left(\bC_K^{(1)}\right)^{\top} p_K\left(\underline{\bW}_K^{\top}\right) \bA_K\bC_K^{(2)}\right) - 
\bT\left(d^2, d^2\right)^{\top} \vect
\left(\left(\bC_K^{(1)}\right)^{\top} \tilde{p}_K\left(\underline{\bW}_K^{\top}\right) \bA_K \bC_K^{(2)}\right)\right\|_{\infty}
\\
\leq & ~ \left\|\left(\bT
\left(d^2,d^2\right)\right)^{\top}
\right\|_{\infty}\left\|\left(\left(\bC_K^{(1)}
\right)^{\top} p_K\left(\underline{\bW}_K^{\top}\right) \bA_K\bC_K^{(2)}\right) -\left(\left(\bC_K^{(1)}\right)^{\top} \tilde{p}_K\left(\underline{\bW}_K^{\top}\right) \bA_K\bC_K^{(2)}\right)\right\|_{\infty} 
\\
\leq & ~ \left\|\left(\left(\bC_K^{(1)}\right)^{\top}\left(p^K_{1}\left(\underline{\bW}_K^{\top}\right) -\tilde{p}_{1}^{K}
\left(\underline{\bW}_K^{\top}\right)\right) \bA_K \bC_K^{(2)}\right)\right\|_{\infty} + \left\|\left(\left(\bC_K^{(1)}\right)^{\top}
\left(p^K_{2}\left(\underline{\bW}_K^{\top}\right) -\tilde{p}_{2}^{K}\left(\underline{\bW}_K^{\top}\right)\right) \bA_K\bC_K^{(2)}\right)\right\|_{\infty}
\\
\leq & ~ \left\|\bA_K\right\|_{\infty}\left\|\bC_K^{(1)}
\right\|_{\infty}
\left\|\bC_K^{(2)}\right\|_{\infty}\left(\left\|\left(p^K_{1}\left(\underline{\bW}_K^{\top}\right)-\tilde{p}_{1}^{K}\left(\underline{\bW}_K^{\top}\right)\right)\right\|_{\infty} 
+ \left\|\left(p^K_{2}\left(\underline{\bW}_K^{\top}\right) -\tilde{p}_{2}^{K}\left(\underline{\bW}_K^{\top}\right)\right)\right\|_{\infty}\right)
\\
\leq & ~ \epsilon / \poly(L).
\end{align*}
Setting $\epsilon = 1/ \poly(L)$, we complete the proof.
\end{proof}

\clearpage
\section{Proof of \texorpdfstring{\cref{thm:main_eff}}{}}
\label{sec:lowerB}

We recall our definition of  $\mathsf{ALoRAGC}(L,d,r,\epsilon)$ for special case from \cref{def:ALoRAGC_special} subject to LoRA loss \eqref{eqn:special_generic_lora_loss_single}.
We aim to make the reduction from  $\mathsf{AAttLGC}(L, r, \epsilon)$ \cite[Definition~1.4]{alman2024fine} to our problem $\mathsf{ALoRAGC}(L,d,r,\epsilon)$.

\begin{definition}[Approximate Attention Loss Gradient Computation ($\mathsf{AAttLGC} (L,r,\epsilon)$), Definition~1.4 of \cite{alman2024fine}]\label{def:AAttLGC}
Given four $L \times r$ size matrices $\bA_1 \in \R^{L \times r}, \bA_2 \in \R^{L \times r}, \bA_3 \in \R^{L \times r}$, $\bE \in \R^{L \times r}$ and 
a square matrix 
$\bX \in \R^{r \times r}$ to be fixed matrices. Assume that $\| \bA_1 \bX \|_{\infty} \leq B$, $ \|\bA_2 \|_{\infty} \leq B$. 
Assume all numerical values are in $\log(L)$-bits encoding. 
Let $\calL(\bX) \coloneqq \frac{1}{2} \| \bD^{-1} \exp(\bA_1 \bX \bA_2^\top/r) \bA_3  - \bE \|_F^2.
$ which $\bD \coloneqq \diag( \exp(
\bA_1 \bX \bA_2^\top /r ) \one_L).$
Let $\frac{\d \calL(\bX)}{\d \bX}$ denote the gradient of loss function $\calL$. The goal is to output a matrix $\wt{g} \in \R^{L \times L}$ such that
\begin{align*}
    \| \wt{g} - \frac{ \d \calL(\bX) }{ \d \bX } \|_{\infty} \leq \epsilon.
\end{align*}
\end{definition}

We recall the main hardness result of \cite{alman2024fine} which shows a lower bound of $\mathsf{AAttLGC}(L, r, \epsilon)$ (\cref{def:AAttLGC}) in the following particular case by assuming SETH. 
\begin{lemma}[Theorem~5.5 of \cite{alman2024fine}]\label{lem:zho}
    Let $\kappa : \mathbb{N} \to \mathbb{N}$ by any function with $\kappa(L) = \omega(1)$ and $\kappa(L) = o(\log L)$.
    Assuming SETH,  there is no algorithm running in time $O(L^{2 - \delta})$ for any constant $\delta>0$ for Approximate Attention Loss Gradient Computation $\mathsf{AAttLGC}(L, r, \epsilon)$, even in the case where $r = O(\log L)$ and the input matrices satisfy $\|\bA_1\|_\infty, \|\bA_2\|_\infty, \|\bA_3\|_\infty \leq O(\sqrt{\log L} \cdot \kappa(L))=B$, $\bE = 0$, $\bX = \lambda I_{r}$ for some scalar $\lambda \in [0,1]$, and $\varepsilon = O(1/(\log L)^4)$.
\end{lemma}

Finally, we are ready for our main proof of \cref{thm:main_eff}.
\begin{proof}
    Considering \cref{def:ALoRAGC_special}, we start with the following $O(1)$ reduction. 
    Given the instance of $\mathsf{AAttLGC}(L,r,\epsilon)$ and $\bA_1 \in \R^{L \times r}$, $\bA_2 \in \R^{L \times r}$, $\bA_3 \in \R^{L \times r}$, $\bE=0$, $B=O(\sqrt{\log L} \cdot \kappa(L))$. We then transfer this instance to the instance of $\mathsf{ALoRAGC}(L,d,r,\epsilon)$ by making the following substitution: 
    \begin{align*}
        \bC^{(1)}\bB_Q=\bA_1,
        \bC^{(2)}=\{\underbrace{\bA_2}_{{L \times r}}, \underbrace{0}_{{L \times (d-r)}} \}/r,
        \bC^{(3)}=\{ \underbrace{\bA_3}_{{L \times r}} , \underbrace{0}_{{L \times (d-r)}} \},
        \bA_{Q}=\{\underbrace{\bX}_{{r\times r}}, \underbrace{0}_{{r \times (d-r)}} \},
        \Gamma = B.
    \end{align*}
    
    Then we have
    $\|\bC^{(2)}\|_\infty, \|\bC^{(1)}\bB_{Q}\bA_{Q}\|_\infty, \|\bY\|_\infty \leq \Gamma$ such that
    \begin{align*}
        \bA_1\ \bX \bA_{2}^{T}/r=\bC^{(1)}\bB_{Q}\bA_{Q}\(\bC^{(2)}\)^\sT,
    \end{align*}
    and hence
    \begin{align*}
        \exp(\bA_1\ \bX \bA_{2}^{T})/r=\exp(\bC^{(1)}\bB_{Q}\bA_{Q}\(\bC^{(2)}\)^\sT).
    \end{align*}
This implies that the upper $L \times r$ subblock is exactly the same. (Here we can assume $\bE=\bY=0$.)
\begin{align*}
    (\bD^{-1} \exp{\bC^{(1)} \bB_Q\bA_Q (\bC^{(2)})^\top} \bC^{(3)} - \bY)|_{L \times r}= (\bD^{-1} \exp(\bA_1 \bX \bA_2^\top/r) \bA_3  - \bE)|_{L \times r}.
\end{align*}
This follows that the derivative with respect to $\bX$ of the RHS is the same as the partial derivative with respect to $\bA_Q$ by embedding $\bX$ into a subblock of $\bA_Q$. Now, by letting $\Tilde{\bG}_{\bA}=\tilde{g}$ in the $\mathsf{AAttLGCC}(L, r, \epsilon)$, which finishes the reduction.
This completes the proof.
\end{proof}

\clearpage
\section{Quadratic Time Complexity of Exact LoRA Gradient Computation}
\label{sec:quadratic_backprop}

Here, we make more comments on tensor-trick decomposed LoRA loss from \cref{lemma:special1}:
\begin{align*}
    \dv{\mathcal{L}(\underline{\bW})}{\underline{\bW}}
    =\sum_{\underline{j}=1}^L \sum_{i=1}^d c(\underline{\bW})_{\underline{j}, i} \C_{\underline{j}}^{\top}\underbrace{\Big(\overbrace{\diag\left(f(\underline{\bW})_j\right)}^{(II)}-\overbrace{f(\underline{\bW})_{\underline{j}} f(\underline{\bW})_{\underline{j}}^{\top}}^{(III)}\Big) 
    }_{(I)}\bC^{(3)}[\cdot, i].
    \annot{i.e., \eqref{eqn:grad_low_rank}}
\end{align*}

\begin{remark}[Benefit from Tensor Trick: Speedup Seemingly Cubic Time Exact Computation]
    \cref{lemma:special1} highlights the benefits of the tensor trick and the potential for speeding up \textit{exact} LoRA adaptation on transformer-based models.
    To be more specific, for any $\underline{j}\in[L]$, \textbf{Part-(I)} is an $L\times L$ matrix, thus requiring $\Theta(L^2)$ time to compute.
    Moreover, with a total of $L$ terms, the overall computation time amounts to $\Theta(L^3)$.
\end{remark}

However, \eqref{eqn:grad_low_rank} decomposes \textbf{Part-(I)} into a \textit{diagonal} \textbf{Part-(II)} and a \textit{low-rank} \textbf{Part-(III)} (specifically, rank-1).
This decomposition allows us to reduce the computation time of \textbf{Part-(I)} to $O(L)$ for each $\underline{j}\in[L]$, and of the entire $\nicefrac{\dd \mathcal{L}(\underline{\bW})}{\dd \underline{\bW}}$ to $O(L^2)$.
Our next theorem verifies this claim and shows such seemingly cubic time exact computation is in fact quadratic.

\begin{definition}
Let $n_1, n_2, n_3$ denote any three positive integers. We use 
${{\cal T}_{\mathrm{mat}}}(n_1,n_2,n_3)$ to
denote the time of multiplying an 
$n_1 \times n_2$ matrix with another 
$n_2 \times n_3$.
\end{definition}

\begin{theorem}[Exact LoRA Gradient Computation Takes Quadratic Time]\label{straightforward}
Suppose the following objects are given and if following conditions hold,
\begin{itemize}
    \item Let $\bC^{(1)}, \bC^{(2)}, \bC^{(3)} \in \R^{L \times d}$
    be in \eqref{eqn:C1C2C3}.
    Let $\bB_Q \in \R^{d \times r}, \bA_Q
    \in \R^{r \times d}, \bW \in \R^{d \times d}$ be in \eqref{eqn:special_generic_lora_loss_single}.
    \item Let $f(\cdot), c(\cdot), p_1(\cdot), p_2(\cdot)$ follow from their definitions in \cref{sec:special}.
    \item Let $\underline{\bG}_{Q}^{(A)} := 
    \pdv{\mathcal{L}}{\underline{\bA}_Q} ~,~
    \underline{\bG}_{Q}^{(B)} := 
    \pdv{\mathcal{L}}{\underline{\bB}_Q}$
    (Where $\mathcal{L} $ is defined in 
\eqref{eqn:special_generic_lora_loss_single}
).
\end{itemize}
Then we can make \textit{exact} computation of $\underline{\bG}_{Q}^{(A)}, \underline{\bG}_{Q}^{(B)}$ in 
     $O(
    {{\cal T}_{\mathrm{mat}}}(d,L,L) + 
    {{\cal T}_{\mathrm{mat}}}(d,d,L) +
    {{\cal T}_{\mathrm{mat}}}(d,d,r)) $   
 time.
\end{theorem}

\begin{proof}  
    Due to \cref{lemma:special3}, it holds
    \begin{align*}
             \pdv{\calL}{\underline{\bA}_Q}
            =\vect\left(\bB_Q^{\top}\left(\bC^{(1)}\right)^{\top} p(\underline{\bW}) \bC^{(2)}\right) 
            ,\quad
            \frac{\partial \mathcal{L}}{\partial \underline{\bB}_Q}
            =\vect\left(\left(\bC^{(1)}\right)^{\top} p(\underline{\bW}) \bA_Q \bC^{(2)}\right).
    \end{align*}   
    Recall that the decomposition of $p(\underline{\bW}) = p_1(\underline{\bW})-p_2(\underline{\bW})$. 
    And according to \cref{def:p1p2},
    for every index $\underline{j}\in [L]$,
    \begin{align*}
    p_1(\underline{\bW})_{\underline{j}}\coloneqq \diag\left(f\left(\underline{\bW}\right)_
    {\underline{j}}\right) q(\underline{\bW})_{\underline{j}},\quad
    p_2(\underline{\bW})_{\underline{j}}\coloneqq f\left(\underline{\bW}\right)_{\underline{j}} f\left(\underline{\bW}\right)_{\underline{j}}^{\top} q(\underline{\bW})_{\underline{j}},
    \end{align*}
    In addition, due to \cref{lemma:special3}, $q(\underline{\bW})$ is defined as
    \begin{align*}
        q(\underline{\bW})
        \coloneqq \bC^{(3)} \(c(\underline{\bW})\)^\sT \in \R^{L\times L}.
    \end{align*}
    Therefore, we compute $f(\underline{\bW}), c(\underline{\bW}),
    p_1(\underline{\bW}), p_2(\underline{\bW})$ in order as follows. Then we combine them together  to get total running time.
\begin{itemize}
    \item \textbf{Step 1.} We compute $f(\underline{\bW})$. 

    Note that
    \begin{align*}
        f(\underline{\bW}) =
        \bD^{-1} \exp \Big( \overbrace{\bC^{(1)}}^{L \times d} \overbrace{\bW}^{d \times d} 
        \overbrace{(\bC^{(2)})^{\top}}^{d \times L}
         \Big) ,
    \end{align*}
    where
    \begin{align*}
        \bD^{-1} =
        \diag (\exp (\bC^{(1)} \bW (\bC^{(2)})^{\top} )  {\one}_L ) .
    \end{align*}
    We firstly compute $\exp (\bC^{(1)} \bW (\bC^{(2)})^{\top}) \bC^{(3)} $ which takes time of ${{\cal T}_{\mathrm{mat}}}(d,d,L) + {{\cal T}_{\mathrm{mat}}}(d,L,L)$.
    
    Then, we can compute $\bD$ which takes $O(L^2)$ time. 
    
    Then, we can compute $f(\underline{\bW})$ which takes $O(L^2)$ time. 
    
    Thus, the overall time is
    \begin{align*}
        {{\cal T}_{\mathrm{mat}}}(d,d,L) + {{\cal T}_{\mathrm{mat}}}(d,L,L) +
        O(L^2) 
        = O({{\cal T}_{\mathrm{mat}}}(d,d,L) + {{\cal T}_{\mathrm{mat}}}(d,L,L)).
    \end{align*}
    Therefore, the proof is completed.

    \item 
    \textbf{Step 2.} We compute $c(\underline{\bW})$. Based on the \cref{def:special_c}, which is
    \begin{align*}
        c(\underline{\bW}) =
        \overbrace{f(\underline{\bW})}^{L \times L} 
        \overbrace{\bC^{(3)}}^{L \times d} - \bY.
    \end{align*}
    Computing $f(\underline{\bW}) \bC^{(3)}$ 
    takes time of ${{\cal T}_{\mathrm{mat}}}(d,L,L)$ and computing $f(\underline{\bW}) \bC^{(3)} - \bY$ 
    takes time of $O(Ld)$. 
    Thus, the overall time is ${{\cal T}_{\mathrm{mat}}}(d,L,L) + O(Ld)
    = O({{\cal T}_{\mathrm{mat}}}(d,L,L)) $.
    
    \item 
    \textbf{Step 3.} We compute $q(\underline{\bW})$. Recall that 
    \begin{align*}
        q(\underline{\bW}) :=
        \overbrace{c(\underline{\bW})}^{L \times d} 
        \overbrace{(\bC^{(3)})^{\top}}^{d \times L}.
    \end{align*}
    Therefore, it takes time $O({{\cal T}_{\mathrm{mat}}}(d,L,L))$.

    \item 
    \textbf{Step 4.} We compute $p(\underline{\bW})$. Note that due to \cref{def:p1p2}, which is
    \begin{align*}
    p_1(\underline{\bW})_{\underline{j}}\coloneqq \diag\left(f\left(\underline{\bW}\right)_
    {\underline{j}}\right) q(\underline{\bW})_{\underline{j}},\quad
    p_2(\underline{\bW})_{\underline{j}}\coloneqq f\left(\underline{\bW}\right)_{\underline{j}} f\left(\underline{\bW}\right)_{\underline{j}}^{\top} q(\underline{\bW})_{\underline{j}},
    \end{align*}
    such that 
    $p(\underline{\bW})=p_1(\underline{\bW})-p_2(\underline{\bW})$.\\
    Since $\diag (f(\underline{\bW})_{\underline{j}})$ 
    is a diagonal matrix and $f(\underline{\bW})_{\underline{j}}
    (f(\underline{\bW})_{\underline{j}})^{\top}$
    is a rank-one matrix, we know that
    $p(\underline{\bW})_{\underline{j}} \in  \R^{L}$ can be computed in $O(L)$, 
    for each $\underline{j} \in [L]$.
    Thus we can construct matrix 
    $ p(\underline{\bW}) \in \R^{L \times L}$ 
    in $ L \times O(L) = O(L^2)$ time in total. 

    \item 
    \textbf{Step 5.} Using \cref{lemma:special3}, we know that 
    \begin{align*}
             \pdv{\calL}{\underline{\bA}_Q}
            =\vect(
            \overbrace{\bB_Q^{\top}}^{r \times d}
            \overbrace{(\bC^{(1)})^{\top}}^{d \times L} 
            \overbrace{p(\underline{\bW})}^{L \times L} 
            \overbrace{\bC^{(2)}}^{L \times d}) 
            ,
            \quad
            \frac{\partial \mathcal{L}}{\partial \underline{\bB}_Q}
            =\vect(
            \overbrace{(\bC^{(1)})^{\top}}^{d \times L}
            \overbrace{p(\underline{\bW})}^{L \times L}
            \overbrace{\bA_Q}^{L \times d} \overbrace{\bC^{(2)}}^{L \times d}).
    \end{align*}  
    Suppose $\bB_{Q} \in \R^{d \times r}, \bA_{Q} \in \R^{r \times d}, \bC^{(1)}, \bC^{(2)}, \bC^{(3)} \in \R^{L \times d} 
    $ are given, then each of the gradients can be computed in time of $O(
    {{\cal T}_{\mathrm{mat}}}(d,L,L) + 
    {{\cal T}_{\mathrm{mat}}}(d,d,L) +
    {{\cal T}_{\mathrm{mat}}}(d,d,r)) $.
    \end{itemize}
    Thus, the overall running time for gradients computation is 
    \begin{align*}
     O(
    {{\cal T}_{\mathrm{mat}}}(d,L,L) + 
    {{\cal T}_{\mathrm{mat}}}(d,d,L) +
    {{\cal T}_{\mathrm{mat}}}(d,d,r)).    
    \end{align*}
    This completes the proof.
\end{proof}

\end{document}